\documentclass{article} 
\usepackage[authoryear,round]{natbib}
\usepackage[preprint]{colm}

\usepackage{microtype}
\usepackage{fontawesome5}
\usepackage{hyperref}
\usepackage{url}
\usepackage{booktabs}
\usepackage{svg}
\usepackage{amssymb} 
\usepackage{bbding}
\usepackage{tabularx}

\usepackage{tikz}
\usetikzlibrary{shapes.geometric, positioning, shadows}

\usepackage{lineno}
\usepackage{graphicx}
\usepackage{amsmath}
\usepackage{enumitem}
\usepackage{geometry}
\usepackage{xcolor}
\usepackage{xspace}
\usepackage{graphicx}
\usepackage{array} 
\usepackage{subcaption}
\usepackage{multirow}
\usepackage{xcolor}
\usepackage{colortbl}

\newif\ifshowvd
\showvdtrue    

\usepackage[colorinlistoftodos, textsize=scriptsize]{todonotes}

\newcommand{\datbench}{%
  \textcolor[HTML]{002ECF}{\textbf{\textsc{DatBench}}}\xspace
}
\newcommand{\bbfull}{%
  \textcolor[HTML]{002ECF}{\textbf{\textsc{DatBench}}{-\textsc{Full}}}\xspace
}

\definecolor{darkblue}{rgb}{0, 0, 0.5}
\definecolor{lightblue}{RGB}{173,216,230}
\definecolor{customlightblue}{HTML}{007cfa}
\hypersetup{colorlinks=true, citecolor=darkblue, linkcolor=darkblue, urlcolor=darkblue}
\usepackage{soul} 

\title{
\vspace{-1em}
\centering
\begin{tabular}{c}
    \begin{tabular}{@{} m{1.5cm} @{\hspace{0.5em}} c @{}}
        \includegraphics[width=1.5cm]{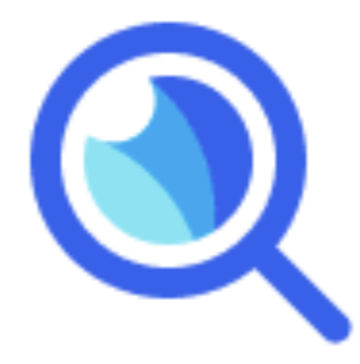} & \huge \textbf{\datbench}
    \end{tabular} \\[0.4em]
    \Large Discriminative, Faithful, and Efficient VLM Evaluations \\
\end{tabular}
}

\definecolor{LightCyan}{rgb}{0.88,1,1}
\usepackage[most]{tcolorbox} 
\tcbuselibrary{listings,breakable}

\usepackage{tabularx}   
\usepackage{array}      
\usepackage{ragged2e}   

\newcolumntype{L}{>{\RaggedRight\arraybackslash}X}

\begin{document}

\ifcolmsubmission
\linenumbers
\fi

\maketitle
\vspace{-5.75em}
\begin{center}
\textbf{DatologyAI Team}\footnotemark
\end{center}
\footnotetext{See Contributions and Acknowledgments (\S~\ref{sec:contri}) for full author list.}

\begin{abstract}
    Empirical evaluation serves as the primary compass guiding research progress in foundation models. Despite a large body of work focused on training frontier vision-language models (VLMs), approaches to their evaluation remain nascent. To guide their maturation, we propose three desiderata that evaluations should satisfy: (1) \textbf{faithfulness} to the modality and application, (2) \textbf{discriminability} between models of varying quality, and (3) \textbf{efficiency} in compute. Through this lens, we identify critical failure modes that violate faithfulness and discriminability, misrepresenting model capabilities: (i) multiple-choice formats reward guessing, do not represent downstream use-cases, and saturate early as models improve; (ii) `blindly-solvable' questions which can be answered without images, constitute up to 70\% of some evaluations; and (iii) mislabeled or ambiguous samples compromise up to 42\% of examples in certain datasets. Regarding efficiency, the computational burden of evaluating frontier models has become prohibitive: by some accounts, nearly 20\% of development compute is devoted to evaluation alone. 
    Rather than discarding existing benchmarks, we curate them via transformation and filtering to maximize their fidelity and discriminability. 
    We find that transformations such as converting MCQs to generative tasks reveal sharp capability drops of up to 35\%. In addition, filtering blindly-solvable and mislabeled samples enhances the discriminative power of these evaluations, while simultaneously reducing their computational cost. We release \bbfull, a cleaned evaluation suite of 33 datasets spanning nine VLM capabilities, and \datbench, a discriminative subset that achieves 13× average speedup (up to 50×) while closely matching the discriminative power of the original datasets. Our work provides a path towards evaluation practices that are both rigorous and sustainable as VLMs continue to scale.
    
    \vspace{0.5em}
    \noindent

    \href{https://huggingface.co/datasets/DatologyAI/DatBench}{\includegraphics[height=1em]{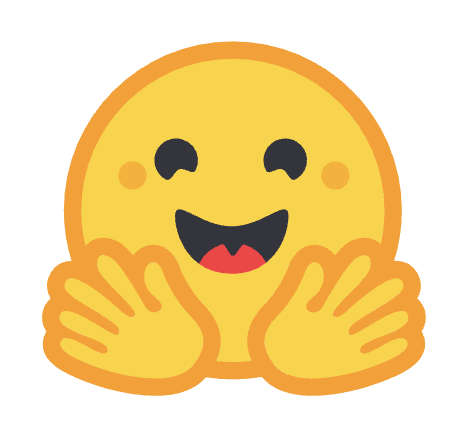} \datbench: \url{https://huggingface.co/datasets/DatologyAI/DatBench}} \\
    \href{https://huggingface.co/datasets/DatologyAI/DatBench-Full}{\includegraphics[height=1em]{figures/hf_logo.png} \bbfull: \url{https://huggingface.co/datasets/DatologyAI/DatBench-Full}} \\
    \href{https://github.com/datologyai/DatBench}{\includegraphics[height=1em]{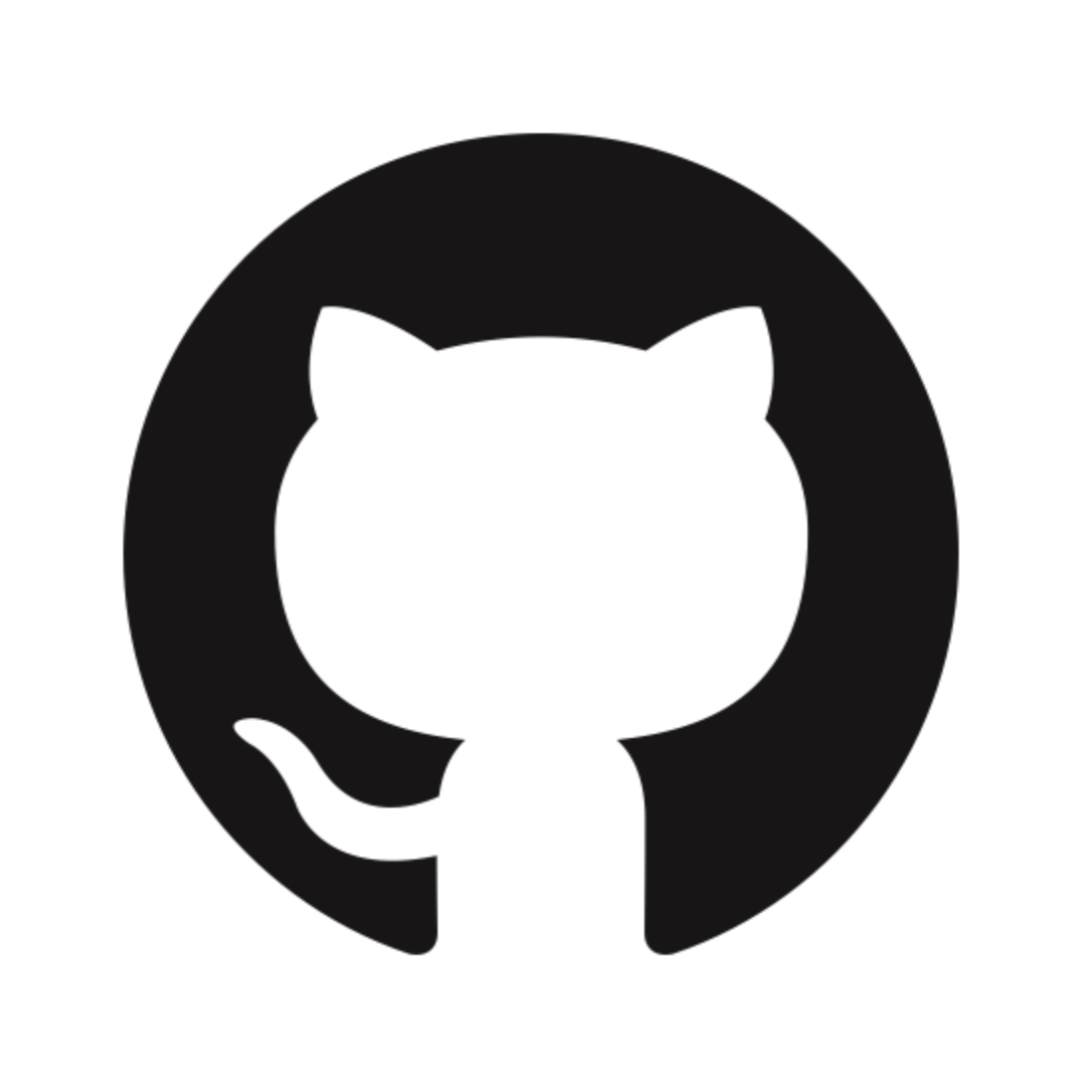} \textbf{Code}: \url{https://github.com/datologyai/DatBench}}
\end{abstract}
\section{Introduction}
\label{sec:intro}

\begin{figure*}[t]
    \centering
    \includegraphics[width=0.48\linewidth]{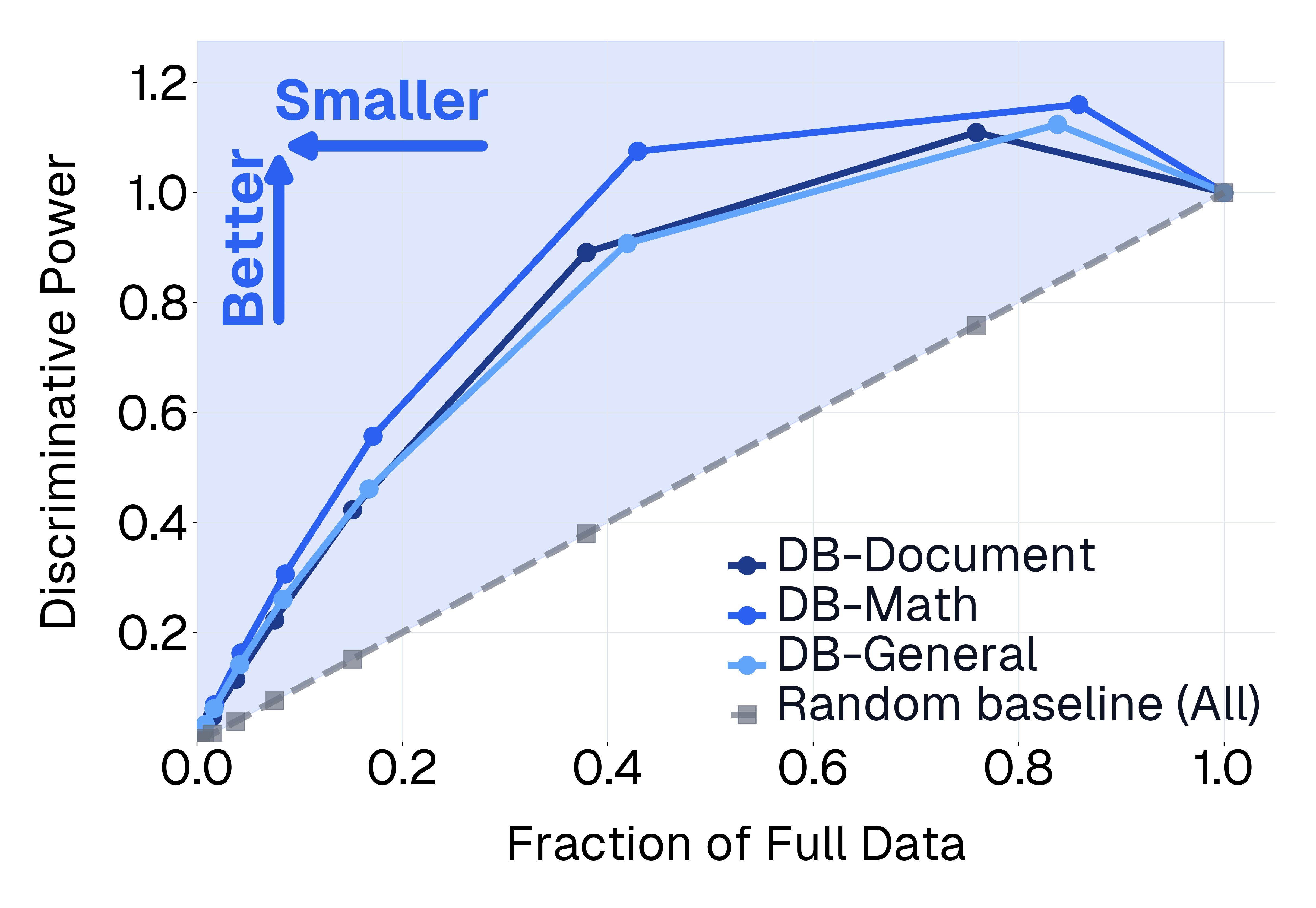}
    \hfill
    \includegraphics[width=0.5\linewidth]{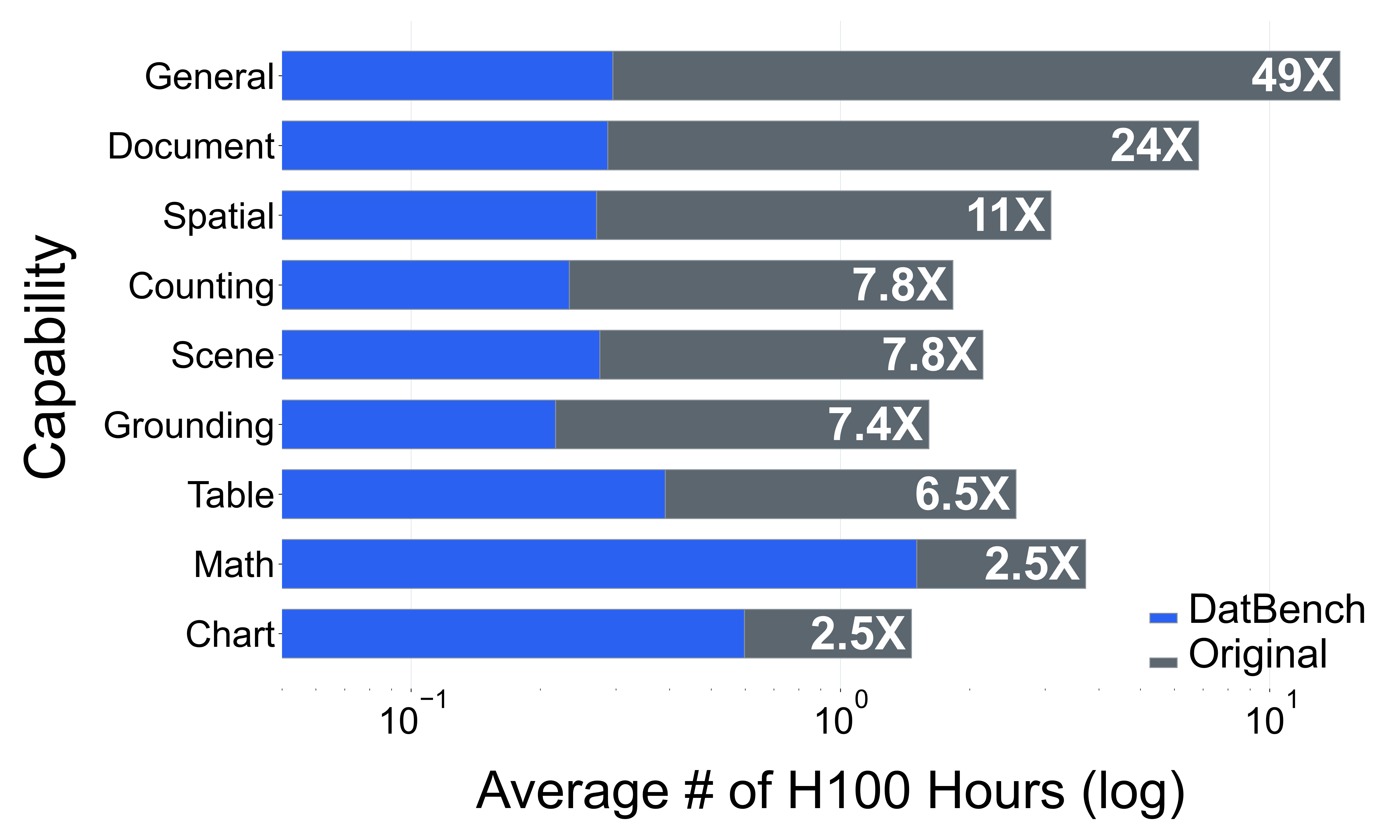}
    \caption{
    \textbf{\datbench reduces evaluation cost while increasing discriminative signal.}
    Panel (a) shows discriminative power as a function of retained data (for select capabilities), demonstrating that targeted selection reaches full-benchmark discriminative power using as little as 40\% of the samples.
    Panel (b) reports average H100 hours and relative speedup across nine capabilities.
    }
    \label{fig:hero}
\end{figure*}

Empirical evaluation is the primary mechanism through which progress in foundation models is recognized, compared, and acted upon. As machine learning has shifted from narrow, task-specific systems to general-purpose vision-language models (VLMs) with broad and compositional capabilities \citep{wei2022emergentabilitieslargelanguage}, benchmarks now play an outsized role: they define what counts as progress and directly shape how substantial computational and human resources are allocated. Evaluations are no longer a passive reporting tool but an active driver of research direction.

However, modern evaluation pipelines are increasingly misaligned with the behaviors they aim to measure. As model inputs span multiple modalities and outputs become increasingly generative and stochastic, benchmarks must better disentangle genuine capabilities from superficial heuristics and inherent variance \citep{stochastic_variation_lu2022fantasticallyorderedpromptsthem}. While the evaluation of language-only models has received sustained methodological attention \citep{llm_eval_mature_srivastava2023imitationgamequantifyingextrapolating, llm_eval_mature_2023opencompass}, VLM evaluation remains comparatively under-examined.

Recent evidence suggests that this gap has become a serious liability. Existing VLM benchmarks suffer from pervasive data quality failures, including mislabeled or ambiguous examples, questions solvable without visual input, and heavy reliance on multiple-choice formats that are not representative of downstream use-cases and are vulnerable to spurious correlations \citep{chartqapro, liu2024mmbench, realworldqa2024, zhang2024mme, yang2024ccocrcomprehensivechallengingocr}. These artifacts inflate reported accuracy, introduce a substantial noise floor, and reduce the signal-to-noise ratio of the evaluations. In such a regime, small improvements, often on the order of a few percent, are more plausibly explained by overfitting to benchmark idiosyncrasies than by real capability gains, rendering the research community vulnerable to hill-climbing on noise \citep{hillclimb_noise_recht2019imagenetclassifiersgeneralizeimagenet, hillclimb_noise_wei2023inversescalingushaped}.

At the same time, evaluation has become a major computational bottleneck. Running comprehensive VLM evaluation suites now consumes a nontrivial fraction of total development compute \citep{strubell2019energypolicyconsiderationsdeep}. For example, during the development of OLMo3, nearly 20\% of the total compute budget for the post-training phase was reportedly dedicated to evaluation alone \citep{lambert2025goodresearchers}. This burden is amplified for VLMs by the dense visual token sequences required to represent high-resolution images and the extended reasoning traces at inference time, which can collectively exceed tens of thousands of tokens per example \citep{qwen3vl}. Detailed analyses indicate that much of this cost is spent evaluating samples that are either trivial, noisy, or weakly discriminative \citep{fluid2025, polo2024tinybenchmarks}.

In this work, we argue that the design of effective evaluation should be treated as a data curation problem. Rather than repeatedly constructing new benchmarks from scratch, we propose to systematically transform and filter evaluation data to maximize faithfulness, discriminative power, and efficiency. This perspective mirrors recent successes in training data curation, in which careful data transformation and selection has produced large gains in model quality and compute efficiency \citep{fang2023datafilteringnetworks, pmlr-v202-joshi23b, abbas2023semdedupdataefficientlearningwebscale, pmlr-v238-joshi24a, joshi2025datasetdistillationknowledgedistillation, joshi2025mmgenenhancingtaskperformance, datologyai_clip_2024, datologyai_text_2024, beyondweb}. We show that the same principles apply, with similar impact, to evaluation.

Guided by this view, we define three desiderata for modern VLM evaluation datasets, (i) \textbf{faithfulness}: examples should genuinely require visual input and reflect intended downstream use cases; (ii) \textbf{discriminability}: examples should reliably separate stronger models from weaker ones; and (iii) \textbf{efficiency}: evaluation should maximize signal per unit of compute.
These criteria expose four systematic failure modes in existing benchmarks and motivate targeted interventions (Section~\ref{sec:datbench}).

First, \textbf{multiple-choice formats are both unfaithful and weakly discriminative} in generative settings. Converting MCQs to open-ended generation reveals large hidden capability gaps. On AI2D, for example, average accuracy drops from 77.56\% to 40.53\%, with the strongest MCQ model losing nearly 35 points. When generative conversion is infeasible, circular evaluation \citep{liu2024mmbench} collapses chance baselines and exposes similar inflation effects.

Second, \textbf{many VLM benchmarks can be solved without vision}. By evaluating models with the image removed, we find that over 70\% of samples in VQA-v2 \citep{balanced_vqa_v2} can be answered correctly using language priors alone. Such examples fundamentally fail to measure multimodal reasoning.

Third, \textbf{low-resolution inputs and inaccurate or ambiguous annotations introduce substantial noise.} Using a multi-stage filtering pipeline, we discard up to 42.07\% of samples in benchmarks such as MME-RealWorld (Autonomous Driving) \citep{zhang2024mme}. In these instances, evaluation is confounded by factual labeling errors and indeterminable ground truths---where poor image quality renders the target objects unrecognizable even to a human observer---effectively precluding reliable performance assessment.

Fourth, \textbf{existing evaluation suites are inefficient}. By explicitly selecting items with high discriminative power across a diverse set of 1B–10B scale models, we achieve speedups of up to 50$\times$ (13$\times$ on average) while closely matching the discriminative power of full benchmarks using a small fraction of the data (Figure~\ref{fig:hero}).

Applying these interventions, we introduce \datbench (Section~\ref{sec:datbench_artifacts}), a curated suite of VLM evaluations designed to be faithful, discriminative, and compute-efficient. To construct it, we partition the large pool of existing datasets into nine fundamental VLM capabilities and release two resulting artifacts:
\begin{itemize}
    \item \textbf{\datbench}, a high-efficiency subset for rapid iteration that provides a 13$\times$ speedup on average across all capabilities while increasing signal per sample.
    
    \item \textbf{\bbfull}, the full collection of high-quality samples remaining after excluding blind-solvable or objectively low-quality data.
\end{itemize}

Beyond efficiency, our work provides empirical insights across 27 state-of-the-art VLMs, revealing structural limitations that are invisible under conventional evaluation (Section~\ref{sec:insights}). We show that inference-time scaling can actively degrade perceptual performance through an overthinking penalty, that current VLMs exhibit a sharp tension between high-level reasoning and low-level perception, and that language priors systematically mask true multimodal capability across popular benchmarks. Together, these resources and findings improve evaluation quality while dramatically reducing its cost, offering a path toward evaluation practices that keep pace with the rapid advancement of vision-language models.
\section{Related Work}

\textbf{Faithful Evaluation.}
Recent research has identified significant issues with the validity of VLM benchmarks, prompting various mitigation strategies. 
To address inflated performance caused by high-risk baselines in multiple-choice evaluations, several studies propose reformulating tasks into generative answer-matching settings \citep{chandak2025answer} or employing circular evaluation techniques \citep{liu2024mmbench}. More broadly, prior work shows that ambiguous and hard-to-solve comparative prompts can systematically induce spurious preferences in models, meaning that the evaluation prompts themselves can become a hidden source of bias when they implicitly force a choice without sufficient grounding or context \citep{adiga-etal-2025-attention}. This further motivates interventions like circular evaluations and other option-robust MCQ protocols. Other efforts focus on statistical refinement of evaluation metrics. For instance, \citet{fluid2025} apply Item Response Theory (IRT) motivated weighting to account for item difficulty and discrimination beyond simple average accuracy.

Beyond these issues, multiple-choice formats are also misaligned with real-world VLM usage, where models are typically deployed in open-ended, generative settings rather than selecting from a small, predefined set of options. As a result, strong MCQ performance may overstate practical capability by rewarding option elimination or prompt-specific biases, as MCQ-based evaluations systematically misrepresent model abilities by constraining outputs and failing to probe the generative behaviors that dominate real-world LLM and VLM deployment \citep{li2024mcqllm}.

Additional analyses suggest that many VLMs can perform well on certain benchmarks without meaningfully leveraging visual input, calling into question whether such evaluations truly measure visual understanding or multimodal reasoning \citep{vblindlee-etal-2025-vlind, vblindli2024surveybenchmarksmultimodallarge, vblindlin2024revisitingrolelanguagepriors, vblindwang2024picture, vblindzhang2025debiasingmultimodallargelanguage}. In contrast to approaches that seek to recover signal through \textit{post hoc} statistical modeling, our method improves evaluation reliability at the source by enhancing data quality via systematic transformation and filtering of benchmark examples, building on both prior work and newly introduced techniques.

\textbf{Efficient \& Discriminative Evaluation.} 
Efforts to improve the efficiency of model evaluation largely draw from (1) psychometric modeling, and (2) exploiting semantic structure in evaluation data. IRT–based methods \citep{fluid2025, polo2024tinybenchmarks} model latent capability variables in order to estimate item difficulty and discrimination. In practice, however, these approaches typically require large, dense response matrices (many models evaluated on many items) to fit parameters stably. Without this scale, estimates can become highly sensitive to hyperparameter choices.

An alternative line of work leverages semantic structure. For example, \citet{vivek2024anchor} employ embedding-based clustering to select representative subsets, while Scales++ \citep{bean2025scales++} relies on qualitative, rubric-driven segmentation of tasks. These approaches face notable limitations. Clustering outcomes are tightly coupled to the choice of embedding model, a significant concern given the lack of unified multimodal embeddings, while rubric-based methods are inherently labor-intensive and subjective.

More broadly, approaches that optimize solely for preserving model rankings suffer from an inherent limitation. As we show in Section~\ref{sec:high_disc_limited_compute}, rank correlation saturates quickly and can often be achieved even by random subsets whose individual samples do not reliably discriminate between weak and strong models. Consequently, prioritizing rank stability risks overfitting to a fixed set of evaluated models without guaranteeing the quality of the underlying examples. Prior work \citep{ghosh2025onebenchtestallsamplelevel} has also proposed aggregating heterogeneous evaluations via Plackett-Luce models, emphasizing ordinal rankings for their robustness to metric calibration issues. While this addresses the challenge of combining diverse measurements, it operates downstream of data quality, aggregating rankings over noisy or blind-solvable samples still propagates those artifacts into the final ordering.

In contrast to these approaches, we shift the focus from preserving global rankings to the targeted curation of individual samples. First, we systematically transform and filter evaluation data to resolve quality issues such as low resolution and labeling errors. Second, we employ a discriminative subset selection strategy that, unlike rank-preservation methods, identifies high-signal samples without requiring the large-scale model response matrices necessary for stable IRT parameter fitting.
\section{The Making of \datbench}
\label{sec:datbench}

\begin{table*}[h!]
\centering
\small
\renewcommand{\arraystretch}{1.2} 
\caption{\textbf{Evaluation Suite.} We select a suite of 33 diverse datasets that balance standard academic baselines with modern ``in-the-wild'' challenges. Horizontal rules separate distinct datasets within each capability pillar.}
\label{tab:eval_suite}
\resizebox{0.99\textwidth}{!}{%
\begin{tabular}{@{}l l p{9cm}@{}}
\toprule
\textbf{Capability} & \textbf{Dataset} & \textbf{Selection Rationale \& Coverage} \\ \midrule

\multirow{4}{*}{\textbf{Chart}} 
 & ChartQA \citep{chartqa} & Standard benchmark for basic chart understanding and data extraction. \\ \cmidrule{2-3}
 & ChartQA Pro \citep{chartqapro} & Challenging counterpart requiring expert-authored reasoning on complex charts. \\ \cmidrule{2-3}
 & CharXiv (Descriptive \& Reasoning) \citep{charxiv} & Scientific charts requiring domain-specific knowledge and terminology. \\ \cmidrule{2-3}
 & InfoVQA \citep{infovqa} & Mixed-media infographics (combining dense captions with visual diagrams). \\ \midrule

\multirow{3}{*}{\textbf{Document}} 
  & CC-OCR (Document Parsing \& KIE) \citep{yang2024ccocrcomprehensivechallengingocr} & Key Information Extraction (KIE) from structured forms and receipts. \\ \cmidrule{2-3}
 & OCR-VQA \citep{mishraICDAR19} & OCR centric Q\&A on book covers. \\ \cmidrule{2-3}
 & OCRBench-V2  \citep{ocrbenchv2} & Comprehensive bilingual OCR benchmark; 31 scenarios covering text recognition, localization, extraction, and reasoning. \\ \cmidrule{2-3}
 & DocVQA \citep{mathew2021docvqadatasetvqadocument} & Standard benchmark for Q\&A on spatial document layouts. \\ \midrule

\multirow{3}{*}{\textbf{Scene OCR}} 
 & TextVQA \citep{singh2019towards} & Industry standard for recognition of text embedded in natural street scenes. \\ \cmidrule{2-3}
 & MME-RW (OCR in the wild) \citep{zhang2024mme} & ``In-the-wild'' challenges including mobile screens and digital signage. \\ \cmidrule{2-3}
 & CC-OCR (Multi-Scene OCR) \citep{yang2024ccocrcomprehensivechallengingocr} & Perspectively distorted and artistically stylized text. \\ \midrule

\multirow{2}{*}{\textbf{Math / Logic}} 
 & MathVista \citep{lu2024mathvista} & Broad coverage of algebraic reasoning and geometry problems. \\ \cmidrule{2-3}
  & Mathverse (with \& without reasoning) \citep{mathverse} & Visual math benchmark with diagram-based problems across 6 information variants; tests true visual reasoning vs. text-only deduction. \\ \cmidrule{2-3}
   & MathVision \citep{mathvision} & Real math competition problems with diagrams; 16 disciplines from algebra to topology. \\ \cmidrule{2-3}
 & LogicVista \citep{xiao2024logicvistamultimodalllmlogical} & Interleaved text-visual clues strictly separating logic from language priors. \\ \midrule

\multirow{2}{*}{\textbf{Spatial}} 
 & RealWorldQA \citep{realworldqa2024} & Physical grounding in everyday photos (depth estimation, spatial relations). \\ \cmidrule{2-3}
 & MME-RW (Video Monitoring \& Autonomous Driving) \citep{zhang2024mme} & Safety-critical spatial awareness (autonomous driving, remote sensing). \\ \midrule

\multirow{3}{*}{\textbf{Grounding}} 
 & RefCOCO \citep{kazemzadeh-etal-2014-referitgame} & General referring expressions (allows both appearance and location words). \\ \cmidrule{2-3} 
 & RefCOCO+ \citep{kazemzadeh-etal-2014-referitgame} & Strict appearance-based grounding (disallows spatial words like ``left''). \\ \cmidrule{2-3}
 & RefCOCO-g \citep{refcocog} & Long, complex syntactic descriptions (testing recursive understanding). \\ \cmidrule{2-3}
 & RefCOCO-M \citep{refcocom} & Cleaned and improved version of the RefCOCO (UNC) validation split for referring expression segmentation \\ \cmidrule{2-3}
 & Pixmo-Point \citep{deitke2024molmopixmoopenweights} & Precision test: requires coordinate-level localization vs. bounding boxes. \\ \midrule

\textbf{Counting} 

 & CountBench \citep{paiss2023teachingclipcount} & Adversarial distractors to prevent density-map estimation or guessing. \\ \cmidrule{2-3}
  & TallyQA \citep{tallyqa} & Open-ended counting VQA; distinguishes simple (detection-only) vs. complex (reasoning-required) questions. \\ \midrule

\multirow{2}{*}{\textbf{Diagrams}} 
 & AI2D \citep{kembhavi2016diagramworthdozenimages} & Standard baseline for science and engineering schematic parsing. \\ \cmidrule{2-3}
 & MME-RW (Diagram/Table) \citep{zhang2024mme} & High-resolution, complex tables found in professional reports. \\ \midrule

\multirow{3}{*}{\textbf{General}} 
 & MMMU-Pro \citep{yue2025mmmuprorobustmultidisciplinemultimodal} & Hardest exam-style questions for reasoning depth across disciplines. \\ \cmidrule{2-3}
 & MMBench \citep{liu2024mmbench} & Evaluates conversation fidelity and instruction following. \\ \cmidrule{2-3}
 & VQA-v2 \citep{balanced_vqa_v2} & Legacy baseline for open-ended visual questioning. \\
\bottomrule
\end{tabular}%
}
\end{table*}
\subsection{MCQ Evaluations: High Noise, Low Fidelity}\label{subsec:mcq_noise}

In this section, we present the methodology for \datbench, a framework designed to transform noisy, large-scale VLM evaluation suites into high-quality, discriminative benchmarks. Our approach systematically addresses four critical failures in current evaluation regimes: (1) \textit{signal dilution} in Multiple Choice Questions (MCQs), (2) examples \textit{solvable without visual context}, (3) \textit{incorrect, ambiguous, or low-resolution} samples, and (4) \textit{prohibitively high computational costs}. Collectively, the first three interventions enhance the \textbf{faithfulness} and \textbf{discrimination} of the evaluation data, while the fourth ensures the resulting benchmark is both \textbf{efficient} and \textbf{discriminative}.

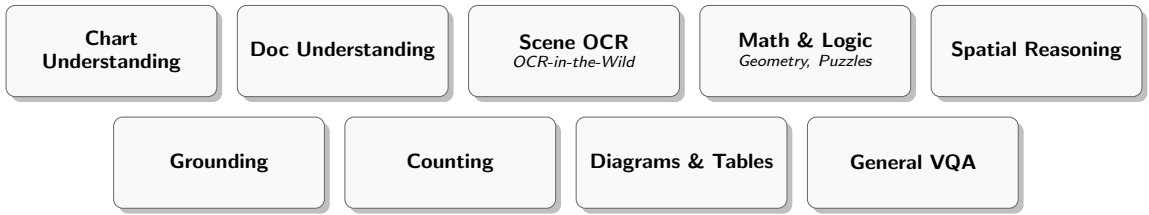
\begin{figure*}[t]
\centering
\resizebox{\textwidth}{!}{%
\begin{tikzpicture}[
    node distance=0.3cm and 0.3cm,
    capblock/.style={
        rectangle, 
        draw=black!60, 
        fill=gray!5, 
        rounded corners=4pt, 
        text width=3.0cm,    
        minimum height=1.4cm, 
        align=center, 
        font=\small\sffamily,
        drop shadow          
    }
]
\node[capblock] (c1) {\textbf{Chart\\Understanding} \\[-2pt]};
\node[capblock, right=of c1] (c2) {\textbf{Doc Understanding} \\[-2pt] };
\node[capblock, right=of c2] (c3) {\textbf{Scene OCR} \\[-2pt] \scriptsize\textit{OCR-in-the-Wild}};
\node[capblock, right=of c3] (c4) {\textbf{Math \& Logic} \\[-2pt] \scriptsize\textit{Geometry, Puzzles}};
\node[capblock, right=of c4] (c5) {\textbf{Spatial Reasoning} \\[-2pt] };
\path (c1.south west) -- (c5.south east) coordinate (row1center);
\node[capblock, below=of c1, xshift=1.65cm] (c6) {\textbf{Grounding} \\[-2pt] };
\node[capblock, right=of c6] (c7) {\textbf{Counting} \\[-2pt] };
\node[capblock, right=of c7] (c8) {\textbf{Diagrams \& Tables} \\[-2pt] };
\node[capblock, right=of c8] (c9) {\textbf{General VQA} \\[-2pt] };
\end{tikzpicture}%
}
\caption{\textbf{Capability Partition.} We evaluate the model across 9 distinct axes of multimodal performance, ranging from low-level perception (OCR, Grounding) to high-level reasoning (Math, Charts).}
\label{fig:capability_map}
\end{figure*}

\paragraph{Datasets \& Capabilities.} We define our goal as establishing a faithful, discriminative, and efficient evaluation for nine distinct VLM capabilities (c.f. Figure~\ref{fig:capability_map}): 
(1)~\textbf{Chart Understanding}: extracting quantitative data and performing trend analysis on bar charts, pie charts, line graphs, and infographics; 
(2)~\textbf{Document Understanding}: parsing structured layouts and extracting key information from digital or scanned documents, with a focus on text-heavy visual processing; 
(3)~\textbf{Scene OCR}: recognizing and interpreting textual information found in natural environments, such as storefront names, street signs, and product labels; 
(4)~\textbf{Math \& Logic}: solving multimodal mathematical problems, including geometry, physics mechanics diagrams, and complex logical puzzles; 
(5)~\textbf{Spatial Reasoning}: assessing the relative positions of objects and demonstrating a directional and physical understanding of 3D space; 
(6)~\textbf{Grounding}: identifying and localizing specific regions or objects referred to in text through bounding boxes or segmentation-style tasks; 
(7)~\textbf{Counting}: accurately enumerating specific objects across varied environments and overlapping visual contexts; 
(8)~\textbf{Diagrams \& Tables}: interpreting grade-school diagrams and structured tables to extract data points and infer underlying relationships; and 
(9)~\textbf{General}: performing high-level Visual Question Answering (VQA) based on holistic image descriptions and real-world scene comprehension. 
To achieve this, we source a diverse pool of evaluation sets for each capability and apply our methodology to address problems (1)--(4), transforming them into refined, high-quality benchmarks. Table~\ref{tab:eval_suite} details the specific dataset composition and selection rationale used to ensure broad coverage of image distributions across each capability.

\paragraph{Models.} We leverage a diverse suite of 27 state-of-the-art models to evaluate and refine our benchmarks. The model families and their corresponding parameter sizes used in this study include: 
(1)~\textbf{Qwen3-VL} (2B, 4B, and 8B Instruct variants, as well as 2B, 4B, and 8B Thinking models); 
(2)~\textbf{Qwen2.5-VL} (3B and 7B Instruct variants); 
(3)~\textbf{Qwen2.5-Omni} (3B and 7B multimodal versions); 
(4)~\textbf{InternVL3.5} (2B, 4B, and 8B Instruct variants); 
(5)~\textbf{InternVL3} (2B and 9B Instruct variants); 
(6)~\textbf{InternVL2.5} (2B, 4B, and 8B variants); 
(7)~\textbf{InternVL2} (2B, 4B, and 8B variants); and 
(8)~\textbf{Thinking \& Specialist Models}, comprising \textit{GLM-4.1V-9B} (Base and Thinking), \textit{R-4B}, \textit{SmolVLM2-2.2B}, \textit{Phi-3.5-vision}, and \textit{Gemma-3-4B-it}. 
Using these models as a broad empirical base allows us to ensure our data-centric improvements generalize beyond any single model family.

For all experiments detailed in this study, model generation was standardized with a maximum output length of 4,096 tokens and suggested sampling configs per the corresponding model card or code repository.

\begin{figure}[h]
    \centering
    \begin{subfigure}[b]{0.48\textwidth}
        \centering
        \includegraphics[width=\linewidth]{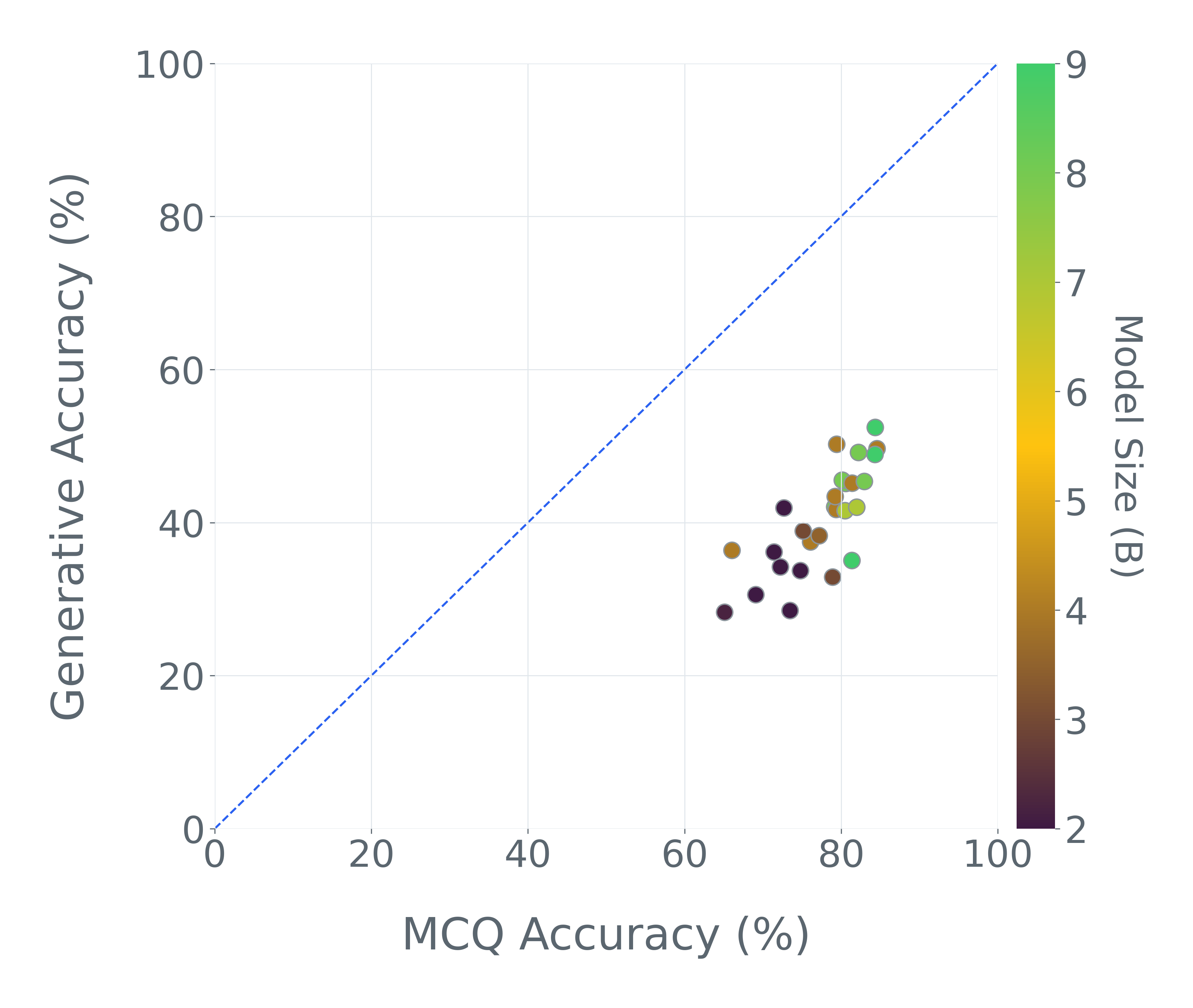} 
        \caption{Generative transformation reveals the non-linear capability gap masked by MCQ guessing.}
    \end{subfigure}
    \hfill
    \begin{subfigure}[b]{0.48\textwidth}
        \centering
        \includegraphics[width=\linewidth]{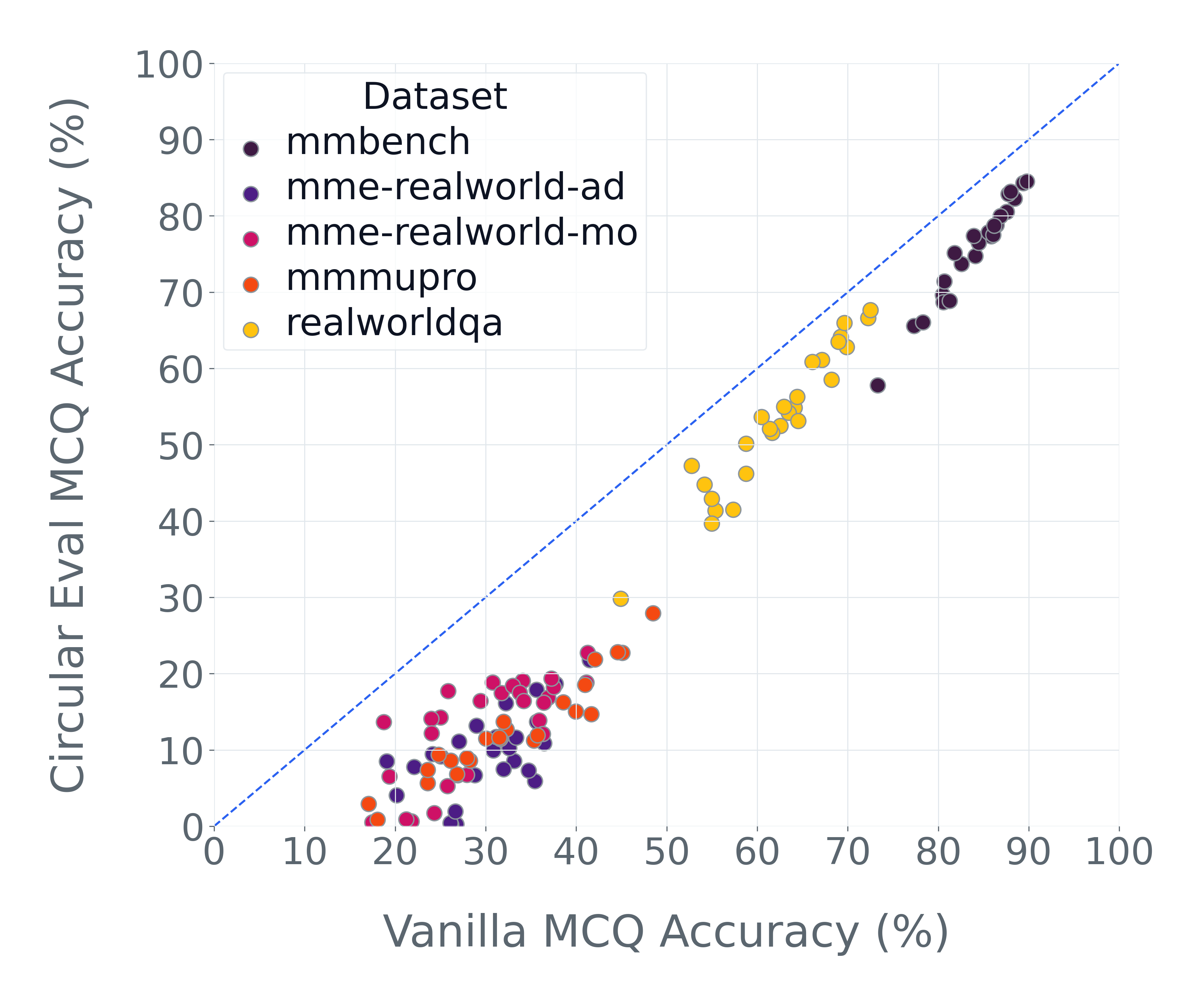} 
        \caption{Circular evaluation yields a more discriminative signal by filtering for consistent reasoning.}
    \end{subfigure}
    \caption{Mitigating performance inflation in multiple-choice formats.}
    \label{fig:mcq_strategy}
\end{figure}

\paragraph{Problem: Chance Baselines and The Evaluation-Deployment Gap}
Standard MCQ formats systematically overestimate model capability through two primary mechanisms: random guessing and task misalignment.
First, multiple-choice questions introduce a non-trivial chance baseline ($1/N$ for $N$ options), allowing models to achieve inflated scores that add significant noise to performance metrics.
This inflation is compounded when evaluating across multiple stochastic samples or models; the probability of an item appearing "solved" by at least one of $M$ uniform random guesses grows rapidly as $1 - (1 - 1/N)^{M}$. Second, there is a fundamental mismatch between evaluation and deployment: while most VLMs are used in generative contexts, MCQs merely test the ability to pick a candidate from a pre-defined list. This "closed-set" evaluation fails to capture the generative reasoning required for real-world tasks and allows models to rely on superficial shortcuts or linguistic priors within the options themselves \citep{chandak2025answer}. As shown in Figure~\ref{fig:mcq_strategy}a, this creates a "perceived capability" bubble in which models appear proficient in MCQ formats while failing to produce the same answers in a fully generative regime.

\paragraph{Solution: MCQ-to-Generative Transformation and Circular MCQ Evaluation} 
To bridge this gap, we adopt a two-pronged strategy to ensure measured performance reflects genuine visual reasoning. Wherever viable, we transform MCQs into open-ended generative tasks by removing candidate options and requiring the model to formulate a direct response. To score these free-form outputs without the brittleness of exact-string matching, we employ an LLM-as-judge (specifically Qwen3-30B \citep{yang2025qwen3technicalreport}, a cost-effective and capable judge) to perform semantic answer matching as in \cite{chandak2025answer}. 

We illustrate the impact of this transformation in Figure~\ref{fig:mcq_strategy}a, which compares standard MCQ accuracy against our generative transformation across 27 models on the AI2D dataset. We observe a distinct non-linear relationship: while high-performing models (80\%+ MCQ accuracy) show tighter convergence between generative and discriminative performance, lower-tier models exhibit a sharp drop-off in the generative setting. This confirms that for weaker models, traditional MCQ benchmarks often mask a fundamental lack of generative skill through random guessing and closed-set shortcuts. 

For tasks where options are structurally necessary, specifically inherently discriminative questions like "Which of the following..." where generative conversion would alter the question's core intent, we implement Circular Evaluation \citep{liu2024mmbench}. By rotating option permutations across $N$ passes and crediting a point only if the model identifies the correct answer across all rotations, we effectively collapse the chance baseline. As shown in Figure~\ref{fig:mcq_strategy}b across 27 models, circular evaluation yields a steeper-than-unity slope relative to vanilla MCQ. This slope captures the persistence of the "false floor" inherent in standard formats: while vanilla MCQs grant models a significant head start (often 20-30\% accuracy) through random guessing and position bias, circular evaluation reveals that genuine reasoning capability remains near zero for these same models. The steepness of the curve illustrates that vanilla MCQ continues to significantly inflate perceived performance while true accuracy remains low ($<50\%$); it is only as models achieve high-level robustness that the two metrics begin to align. By stripping away this artificial inflation, we ensure the benchmark can accurately signal the transition from zero to genuine competence, a critical signal that is otherwise obscured by the noisy MCQ baseline.

Correcting this inflation is crucial: a benchmark is most valuable when it can accurately track the emergence of a new capability. By allowing MCQ formats to provide a "false floor" of performance, we lose the ability to signal when a model truly transitions from zero to non-zero capability. Ultimately, these stricter criteria ensure that \datbench provides a more faithful representation of genuine model competence by stripping away the artificial inflation inherent in traditional formats.

\subsection{The Mirage of Visual Understanding}
\label{subsec:blind_solvability} 

\paragraph{Problem: Language Priors are often \textit{all you need}}
A significant challenge in VLM evaluation is ``blind solvability'', a phenomenon in which models correctly answer questions without visual input by exploiting the \textit{language priors} encoded in their LM backbones. This phenomenon fundamentally decouples benchmark performance from actual multimodal reasoning, inadvertently rewarding models with stronger language priors rather than superior visual understanding. This creates a ``mirage'' of progress, due to which improvements in the vision encoder or cross-modal connector are masked by the overwhelming influence of the text-based backbone. Consequently, models with more capable LMs are often deemed to be stronger VLMs simply because they are better at guessing answers from context.

\begin{figure}[h]
    \centering
    \begin{subfigure}[b]{0.49\textwidth}
        \centering
        \includegraphics[width=\linewidth]{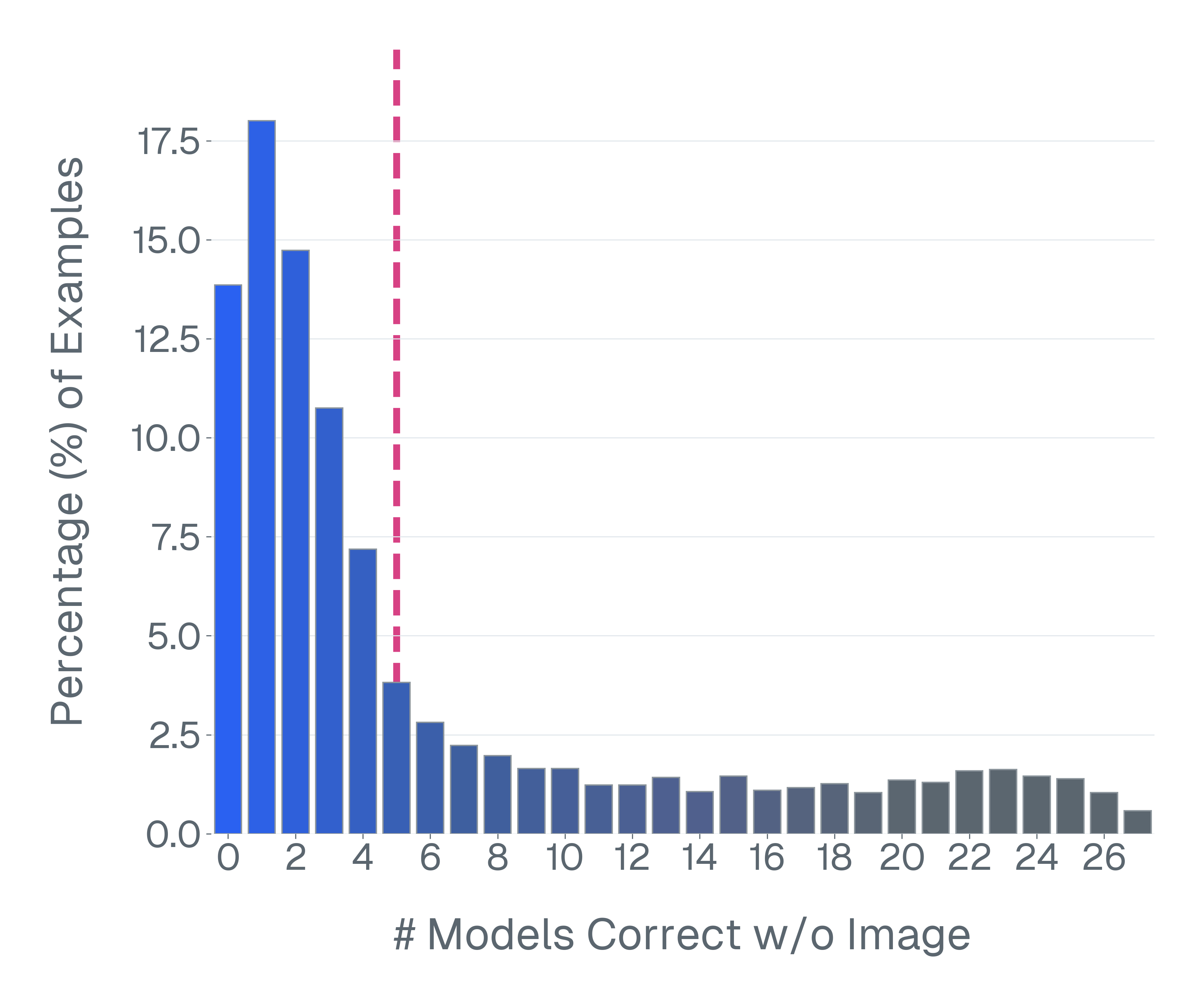}
        \caption{AI2D}
        \label{fig:blind_ai2d}
    \end{subfigure}
    \hfill
    \begin{subfigure}[b]{0.49\textwidth}
        \centering
        \includegraphics[width=\linewidth]{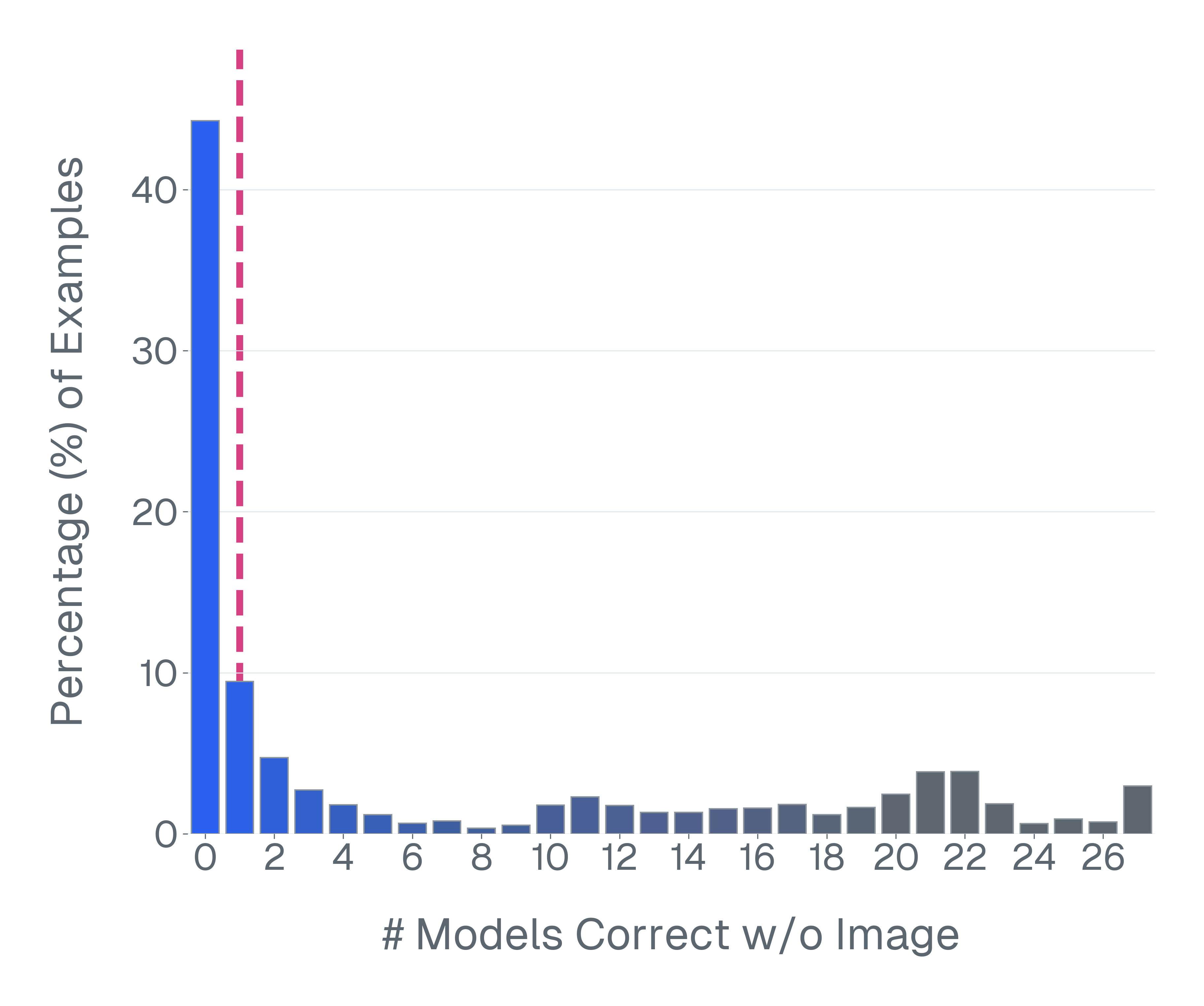}
        \caption{CharXiv-Descriptive}
        \label{fig:blind_charxiv}
    \end{subfigure}
    \caption{Language Priors are often \textit{all you need}: Histograms show the fraction of dataset examples ($y$-axis) correctly answered by a specific number of models ($x$-axis) using text alone. Red dashed lines denote rejection thresholds ($\tau$): examples solved by more models than $\tau$ are discarded. We apply a strict $\tau=1$ for generative tasks, while using higher thresholds for MCQ and constrained-answer tasks to adjust for random guessing baselines (c.f. Appendix~\ref{app:step2})}
    \label{fig:blind_eval}
\end{figure}

\begin{table*}[t]
\centering
\small
\renewcommand{\arraystretch}{1.3}
\setlength{\tabcolsep}{5pt}
\begin{tabularx}{\textwidth}{p{1.6cm} X X p{1.4cm}}
\toprule
\textbf{Dataset} & \textbf{Question (text-only part)} & \textbf{Model Output} & \textbf{Ground Truth} \\
\midrule
\textsc{AI2D} & How many stages are in the development of a mosquito? & 4 & 4 \\
\midrule
\textsc{MathVision} & Please solve the problem step by step. Mary has written all the numbers from 1 to 30. How many times has she written digit 2? & To determine how many times the digit 2 appears in the numbers from 1 to 30, we break it down by place value: Units place: The digit 2 appears in 2, 12, and 22 (3 times). Tens place: The digit 2 appears in 20--29 (10 times). Adding both contributions gives $3 + 10 = 13$. & 13 \\
\midrule
\textsc{VQA-v2} & What color is the toilet? & Most toilets are white, as white is the most common color for bathroom fixtures. Based on this common prior, the toilet is white. & white \\
\bottomrule
\end{tabularx}
\caption{Qualitative examples of \emph{blind-solvable} questions. Models correctly answer these questions without access to images, relying solely on language priors, world knowledge, and symbolic reasoning rather than visual understanding.}
\label{tab:blind_solvable_examples}
\end{table*}

\paragraph{Solution: Filtering Blind-Solvable Questions} 
To ensure \datbench measures genuine vision-language integration, we systematically identify and remove samples that models can solve ``blind.'' We conduct a comprehensive evaluation where all 27 models in our suite are queried using only the text prompts from each dataset, without the corresponding image inputs. For each dataset, we visualize this in a histogram (Figure~\ref{fig:blind_eval}) where the $x$-axis represents the number of models answering correctly and the $y$-axis represents the fraction of the dataset solved at that model-frequency.

As shown in Table~\ref{tab:blind_solvable_examples}, blind-solvable questions typically fall into three categories: (1) \textit{World Knowledge}, where the answer is physically or culturally standard  (e.g., a mosquito's four-stage life cycle); (2) \textit{Visual Stereo-typicality}, where models exploit the skewed distribution of attributes in natural images to predict answers without visual confirmation (e.g., toilets usually being white); and (3) \textit{Purely Symbolic Reasoning}, where the question contains all necessary information for a LLM to solve via logic or arithmetic (e.g., counting digits in a range).

We employ a systematic thresholding strategy ($\tau$) to define rejection criteria based on task format. For datasets with open-ended, generative answers where the probability of a model guessing the exact string is negligible, we set a strict threshold ($\tau = 1$); any sample solved by even a single model without visual input is discarded (e.g., \textit{CharXiv-Descriptive}). Conversely, for tasks with a constrained solution space—such as Multiple Choice Questions (MCQ) or specialized counting tasks—we set higher thresholds to account for the increased baseline of random guessing and language priors. This includes datasets like \textit{CountBench}, where answers are concentrated at low integers, or specific questions in \textit{AI2D} that feature a limited set of candidate solutions evident from the prompt (see Row 1 of Table~\ref{tab:blind_solvable_examples}).

As illustrated in Figure~\ref{fig:blind_eval}a for \textbf{AI2D}, the distribution shows a significant ``tail'' of questions solvable by nearly all models without an image. Even in more recent evaluations like \textbf{CharXiv Descriptive} (Figure~\ref{fig:blind_eval}b), a large fraction of samples are solvable through language priors alone despite the descriptive nature of the task. In the \textit{General} capability, this issue is most acute: over 70\% of examples can be answered without the image. By removing these samples, \datbench ensures the evaluation focuses on high-quality data where visual reasoning is mandatory for success.

\subsection{Incorrect Ground Truth and Ambiguity}
\label{subsec:data_quality}
\begin{figure}[h]
    \centering
    \includegraphics[width=0.83\linewidth]{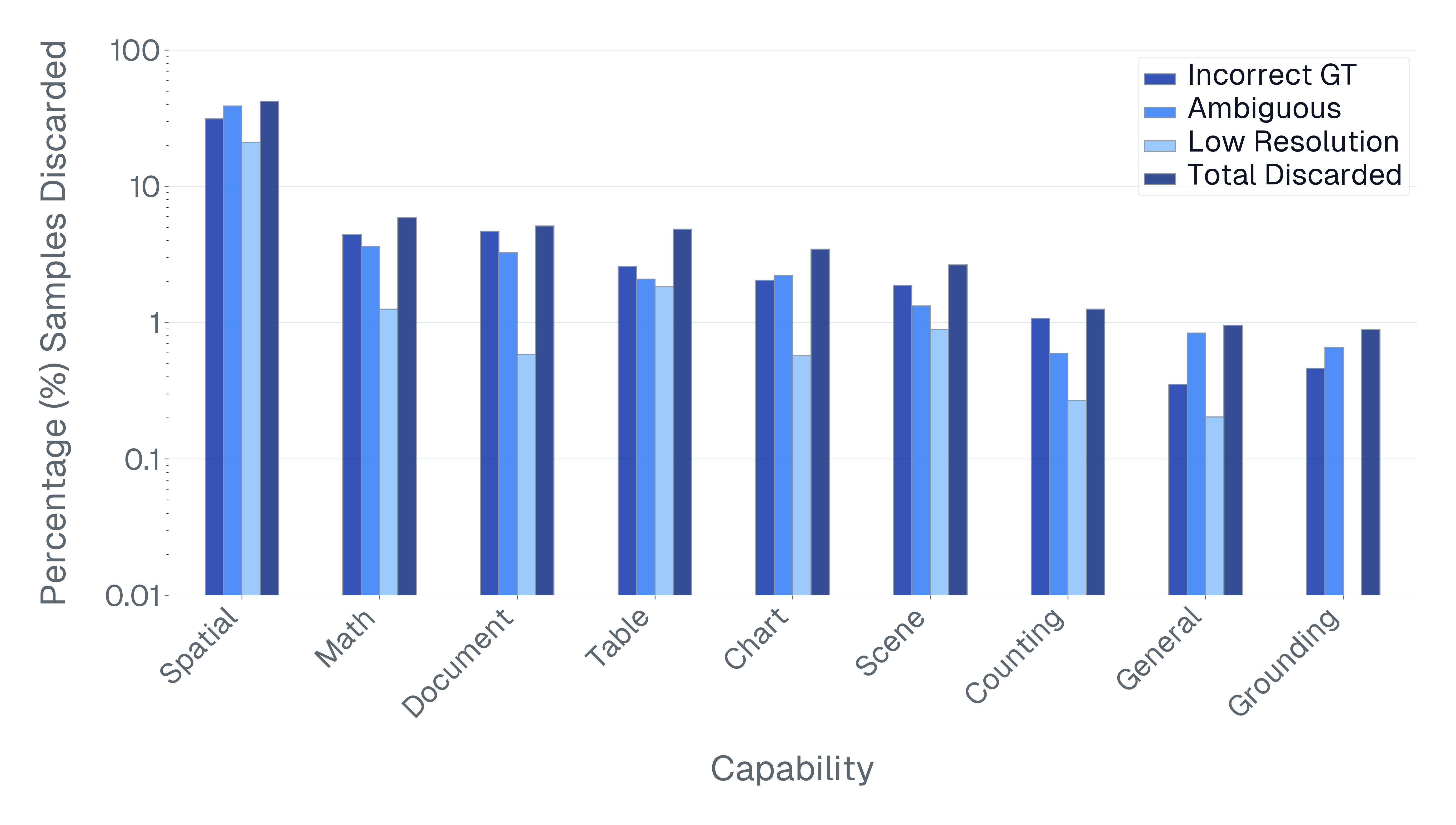}
    \caption{\textbf{VLM-as-Judge Quality Filtering.} Percentage of samples discarded per capability due to ambiguous questions, incorrect ground truth, and samples that are too low resolution (log scale). Spatial capability exhibits the highest discard rate (42.07\%), while well-curated capabilities like Grounding and General require minimal filtering ($<1\%$). Note that a single sample may be labeled under multiple discard categories and is counted in each applicable category.}
    \label{fig:groundtruthquality4bar}
\end{figure}

\begin{figure}[h]
    \centering
    \includegraphics[width=1\linewidth]{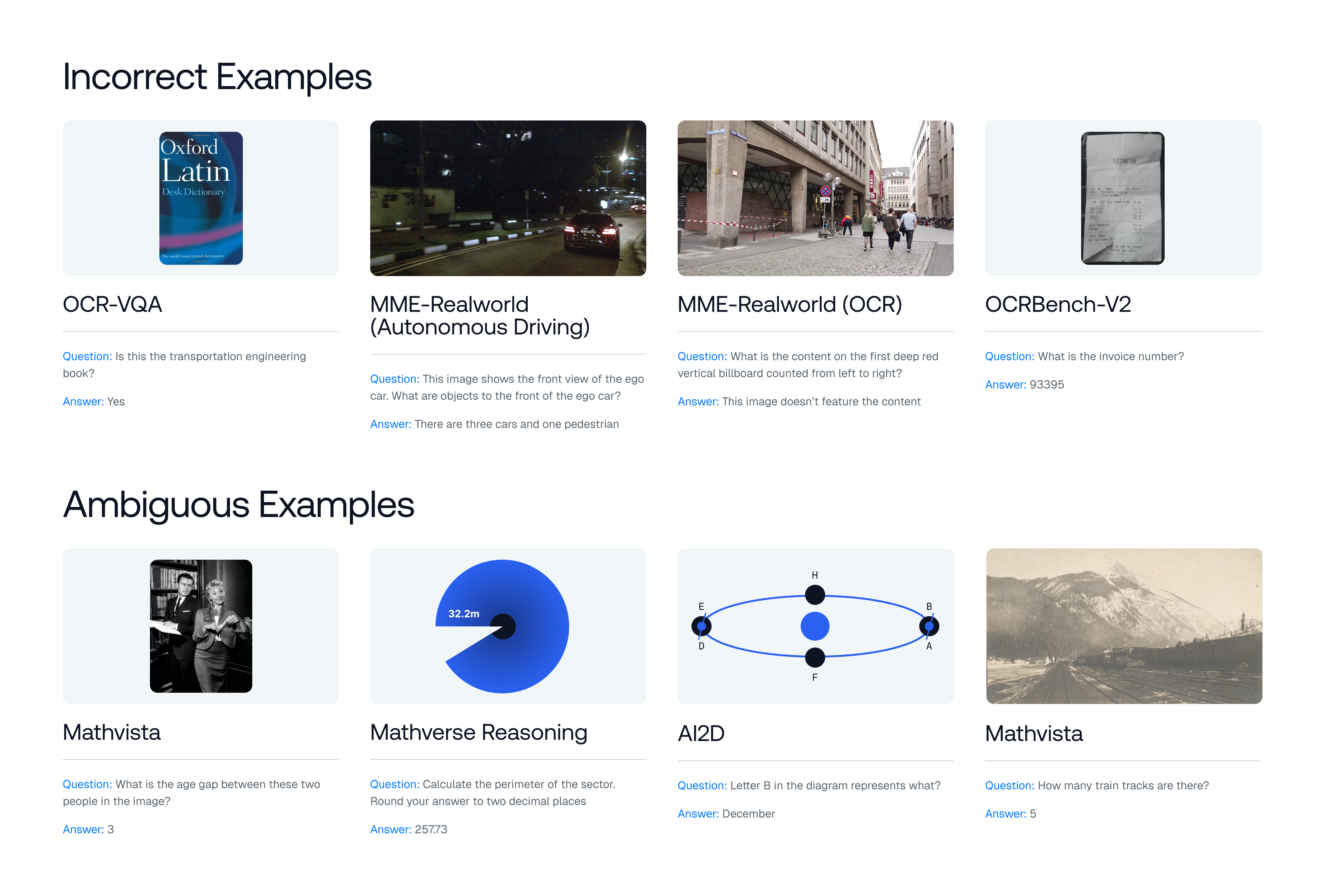}
    \caption{\textbf{Top half}: Judge identifying incorrect ground-truth samples. \textbf{Bottom half}: Quality control via VLM-as-judge filtering. Examples shown were flagged by unanimous model failure and confirmed as low-quality by the judge.}
    \label{fig:5}
\end{figure}

\paragraph{Problem: The Cost of Evaluative Noise}
Despite significant curation efforts, many existing VLM benchmarks contain non-trivial proportions of examples with incorrect ground-truth labels, ambiguous questions, or insufficient image resolution to support the required reasoning. Such noise fundamentally compromises benchmark validity; when a dataset punishes a model for providing a correct answer that contradicts a flawed label, it obscures genuine capability gains and encourages ``hill-climbing on noise''. Since we source \datbench from a massive aggregate pool of candidate datasets, we have a surplus of examples that allows us to prioritize rigorous data quality over raw quantity.

\paragraph{Solution: Two-Stage Quality Filtering with VLM-as-Judge} 
To identify and purge these artifacts, we employ a two-stage filtering pipeline. In the first stage, we flag examples that \textit{all} evaluated models (1--10B parameters) answer incorrectly. Unanimous failure across a diverse suite of state-of-the-art models typically indicates either a data quality issue or a genuinely difficult frontier case, both of which warrant closer inspection. In the second stage, a strong VLM judge (GPT-5.2) verifies each flagged sample with access to the ground-truth answer as privileged information. 

Our choice of a frontier model as a judge is motivated by prior work suggesting that models are significantly stronger verifiers than generators \citep{verifier_Liao_2025_CVPR, verifier_saad2025shrinking, verifier_venktesh2025trustverifysurveyverification, verifier-guan-etal-2024-language}; we therefore expect the judge to accurately identify errors even in cases that are too challenging for contemporary models to solve autonomously. Given that we operate in a regime of abundant evaluation data across our 9 capabilities, we intentionally err on the side of caution. We adopt a stringent filtering policy, discarding any item flagged as (1) ambiguous, (2) incorrectly labeled, or (3) unsolvable due to insufficient resolution, ensuring that the resulting \datbench subset represents only the highest quality of evaluation data. The impact of this pipeline is most visible in the \textbf{Spatial} capability, which exhibits a 42.07\% discard rate, primarily due to insufficient resolution in ``in-the-wild'' images. Similarly, complex expert-authored sets like \textbf{ChartQA Pro} (17.2\% removed) and \textbf{MMMU-Pro} (24.3\% removed) show significantly higher noise rates than standard benchmarks (c.f. Appendix \ref{app:step3} for per dataset / capability counts of filtered examples). While these high attrition rates reflect significant noise in frontier evaluations, we recognize that a judge might occasionally misinterpret specialized, valid reasoning as a data defect. To maintain evaluative headroom, we retain only the subset of these examples that the judge explicitly validates as correct and unambiguous. Our aggregate data surplus allows us to prioritize this high-fidelity subset, accepting the risk that a conservative filtering policy may sacrifice some valid frontier samples to ensure the remaining benchmark remains strictly noise-free.

\subsection{High Discrimination with Limited Compute}
\label{sec:high_disc_limited_compute}

\paragraph{Problem: The Computational Burden of Comprehensive Evaluation}
As VLMs grow in sophistication and expand their set of capabilities, comprehensive evaluation imposes a prohibitive computational burden. This is exacerbated by the emergence of ``thinking'' models; for instance, \cite{qwen3vl} utilize inference-time compute scaling, often generating chains-of-thought exceeding 32K tokens. Consequently, evaluating a single capability like OCR (often containing $>100$K examples) can require generating over 3 billion tokens, an untenable cost for iterative research. 

Selecting a representative subset of examples is a \textit{natural} approach to reducing evaluation costs. The intuitive heuristic for such a selection is to preserve the model ranking induced by the full dataset, typically quantified using rank correlation measures such as Spearman’s $\rho$ or Kendall’s $\tau$ \citep{spearman1904,voorhees2001,buckley2004}. While rank preservation is a necessary condition for a representative subset, it is theoretically insufficient: rank correlation is agnostic to \emph{which} specific samples are retained. In practice, even random subsets can preserve global model rankings by retaining items that separate coarse capability tiers (e.g., small versus large models), while failing to retain the high-discrimination examples needed to distinguish models along the Pareto frontier. More broadly, methods that optimize solely for rank preservation face a fundamental limitation, rank correlation saturates rapidly and is often achieved by subsets whose individual samples are weakly or inconsistently informative about underlying capabilities \citep{sakai2007,voorhees2001}. In such regimes, apparent ranking stability may be driven by spurious correlations or superficial artifacts rather than genuine reasoning ability. 

Instead, we turn to \textit{Item Response Theory (IRT)} for inspiration, originally formalized by \citet{lord1952theory}. IRT posits that items differ not just in difficulty, but in \textit{item discrimination}, a parameter that determines how sharply an item distinguishes between subjects of varying ability levels \citep{baker2001basics}. However, directly applying standard IRT methodologies \citep{fluid2025} to VLM evaluation is often infeasible due to the limited number of diverse observations available per sample in the current research landscape \citep{bean2025scales++, liao2025unifiedframeworkdataefficientevaluation}. Effectively fitting IRT models typically requires stable evaluations from hundreds of diverse state-of-the-art models; without this scale, IRT models become highly sensitive to hyperparameters and are notoriously difficult to fit stably.

Consequently, simply prioritizing rank stability risks overfitting to the evaluated model suite, without guaranteeing the quality or generalizability of the underlying examples. In effect, this produces a ``coarse'' measuring stick: it yields a subset that is discriminative enough to recover a specific ranking but lacks the \textit{resolution} to generalize to unseen models or distinguish those with similar capabilities. Therefore, the core optimization problem is not merely to maintain ranking stability, but to maximize \textit{total discrimination}. By ensuring every sampled example possesses high discriminative power, we can implicitly guarantee robust ranking while maximizing the information content per inference token.

\begin{figure}[h]
    \centering
    \begin{subfigure}[b]{0.48\textwidth}
        \centering
        \includegraphics[width=\linewidth]{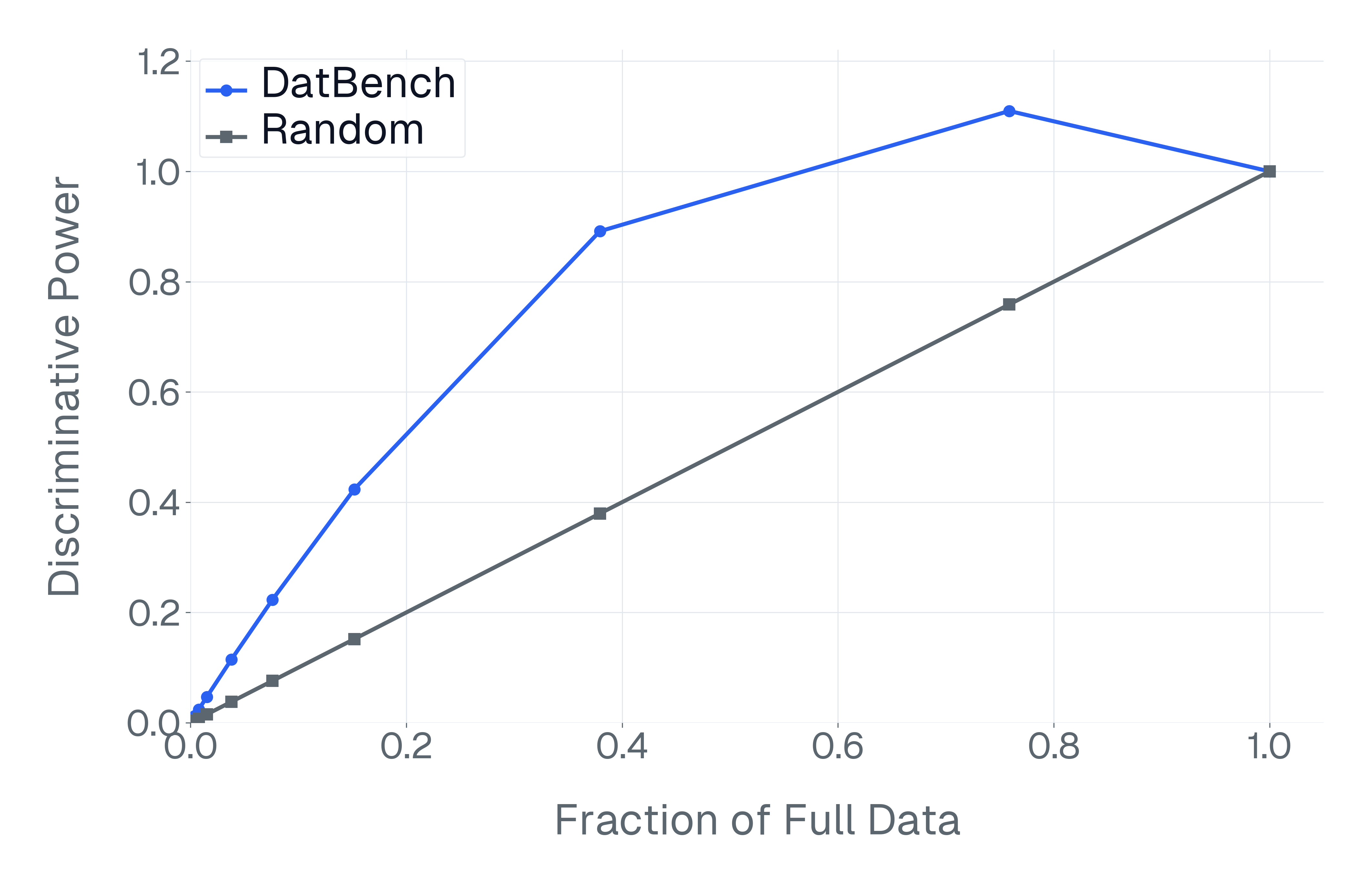}
        \caption{Discrimination vs. Budget}
        \label{fig:disc_vs_budget}
    \end{subfigure}
    \hfill
    \begin{subfigure}[b]{0.48\textwidth}
        \centering
        \includegraphics[width=\linewidth]{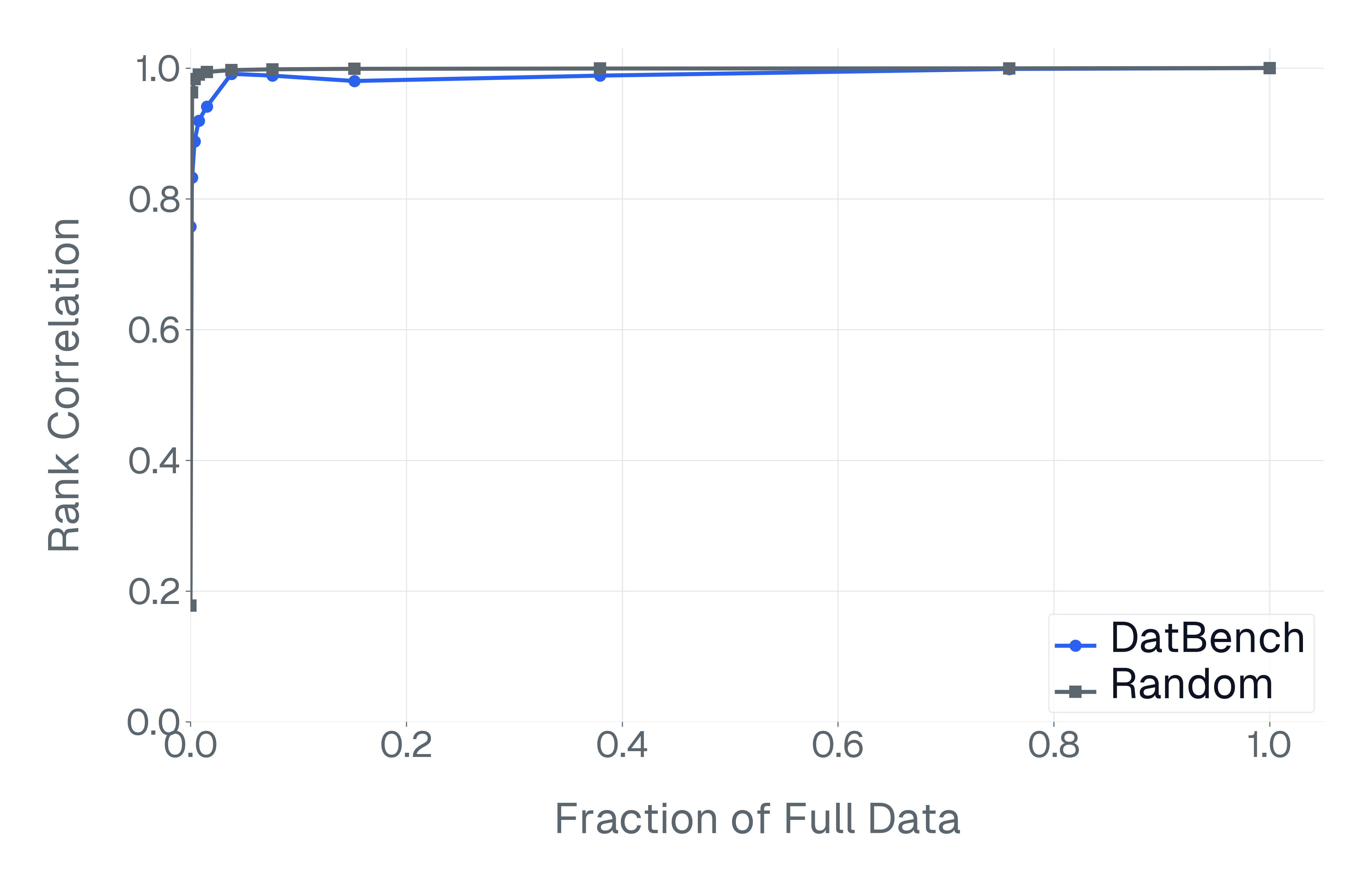}
        \caption{Rank Correlation vs. Budget}
        \label{fig:corr_vs_budget}
    \end{subfigure}
    \caption{\textbf{Efficiency and Discrimination Analysis.} (a) \datbench (blue) maintains significantly higher discriminative power at low sample budgets compared to random sampling (gray). The peak at ~75\% budget followed by a dip indicates the removal of \textit{anomalous items} (negative $r_{pb}$) that degrade evaluation quality. (b) Rank correlation saturates rapidly for both methods due to the distinct ability gaps in the model suite, highlighting why correlation alone is an insufficient metric for subset selection. Data shown for \textit{Document Understanding}; trends are consistent across all capabilities (c.f. Appendix~\ref{app:step4}).}
    \label{fig:discrimination}
\end{figure}

\paragraph{Solution: Item-Discrimination Based Subset Selection} 
To avoid the instability of IRT models that are sensitive to hyperparameters and sample size, we operationalize item-discrimination using the point-biserial correlation ($r_{pb}$): a robust, hyperparameter-free measure of the association between a binary item response and continuous model capability. Intuitively, $r_{pb}$ measures the extent to which success on a specific question acts as a proxy for global performance. An item with high $r_{pb}$ is one that strong models consistently answer correctly and weak models consistently miss; conversely, a low or negative $r_{pb}$ indicates a noisy item that fails to track with underlying capability. We define total discriminative power as the sum of discrimination of each example (item). 

We select subsets by prioritizing examples with the highest $r_{pb}$ to maximize information density (c.f. Appendix \ref{app:step4}). As demonstrated in Figure~\ref{fig:discrimination}a, \datbench achieves approximately 90\% of the total discriminative power using only 40\% of the full dataset, whereas random sampling scales linearly and provides less than half that signal at the same budget. Notably, our selection curve peaks above 1.0 before the full dataset is included; this occurs because we intentionally deprioritize "anomalous items" at the end of our selection process. These are questions with negative $r_{pb}$ where weaker models outperform stronger ones—likely due to spurious text-based correlations, prompt sensitivity, or test-set leakage—which effectively introduce noise into the evaluation.

While Figure~\ref{fig:discrimination}b shows that both random and discriminative subsets rapidly achieve high rank correlation, this similarity is deceptive. Because our model suite contains distinct performance tiers (e.g., 1B vs. 8B), the global ranking is easily recovered even by uninformative samples. Rank correlation is thus a "low-bar" metric that saturates too quickly to reflect subset quality. By maximizing discrimination, \datbench provides a higher-fidelity instrument that remains sensitive to marginal capability gains and ensures that evaluation remains stable across unseen model architectures.

\section{Introducing \datbench and \bbfull}
\label{sec:datbench_artifacts}

By applying our four-stage pipeline: MCQ transformation (Section \ref{subsec:mcq_noise}), blind-solvability filtering (Section \ref{subsec:blind_solvability}), quality filtering (Section \ref{subsec:data_quality}), and discriminative selection (Section \ref{sec:high_disc_limited_compute}), we transform noisy, redundant dataset aggregations into precise evaluation artifacts. These artifacts cover nine distinct capabilities: \textit{Chart Understanding, Document Understanding, Scene OCR, Grounding, Counting, Spatial Reasoning, Math \& Logic, Diagrams \& Tables, and General VQA}. We release two versions of the benchmark to cater to varying computational budgets.

For the final \datbench subset, we execute steps 1 through 4. However, the discrimination-based selection in Step 4 naturally discards "frontier" examples—items that all evaluated models fail—as they offer near-zero discrimination by construction. To prevent benchmark saturation and ensure evaluative headroom for future models, we manually allocate up to 20\% of the \datbench subsets for these valid frontier cases, specifically those verified by our VLM-as-judge as correct and unambiguous. This strategic inclusion ensures that \datbench maintains a high difficulty ceiling while remaining a robust instrument for measuring progress at the frontier of vision-language modeling.

\begin{itemize}[leftmargin=*, nosep]
    \item \textbf{\datbench}: Our primary, high-efficiency subset tailored for rapid iterative development. Constructed via item-wise point-biserial correlation ($r_{pb}$), this set maintains high ranking fidelity while minimizing inference costs. We explicitly retain a partition of verified, high-quality ``frontier'' examples—currently unsolvable by 1B--10B models—to ensure the benchmark remains an effective measuring stick as model capabilities scale.
    \item \textbf{\bbfull}: The complete aggregation of all high-quality samples remaining after our systematic filtering pipeline (Steps 1--3). While these sets include all examples validated as objectively high-quality, their scale varies significantly across capabilities based on the severity of the filtering required. For capabilities such as \textit{Counting} and \textit{Spatial Reasoning}, where high noise and blind-solvability rates resulted in massive attrition, \bbfull is comparable in size to the \datbench subset. However, for most capabilities, \bbfull evaluation sets are an order of magnitude larger, reaching up to $50\times$ the size of their efficient counterparts. These are intended for extensive, fine-grained error analysis and established as a comprehensive resource for deep-dive capability assessment.
\end{itemize}

\paragraph{Usage Guide.}
We recommend \textbf{\textsc{\datbench}} in high-iteration contexts such as training loops and ablation studies, in which compute costs for evaluation can rapidly balloon but discriminative signal should be maximized. \bbfull should be reserved for final model reporting where computational constraints are relaxed and maximum coverage is desired. Collectively, these artifacts transition multimodal evaluation from a regime of noisy data to one of precise measurement.

\begin{figure}[h]
    \centering
    \includegraphics[width=0.99\linewidth]{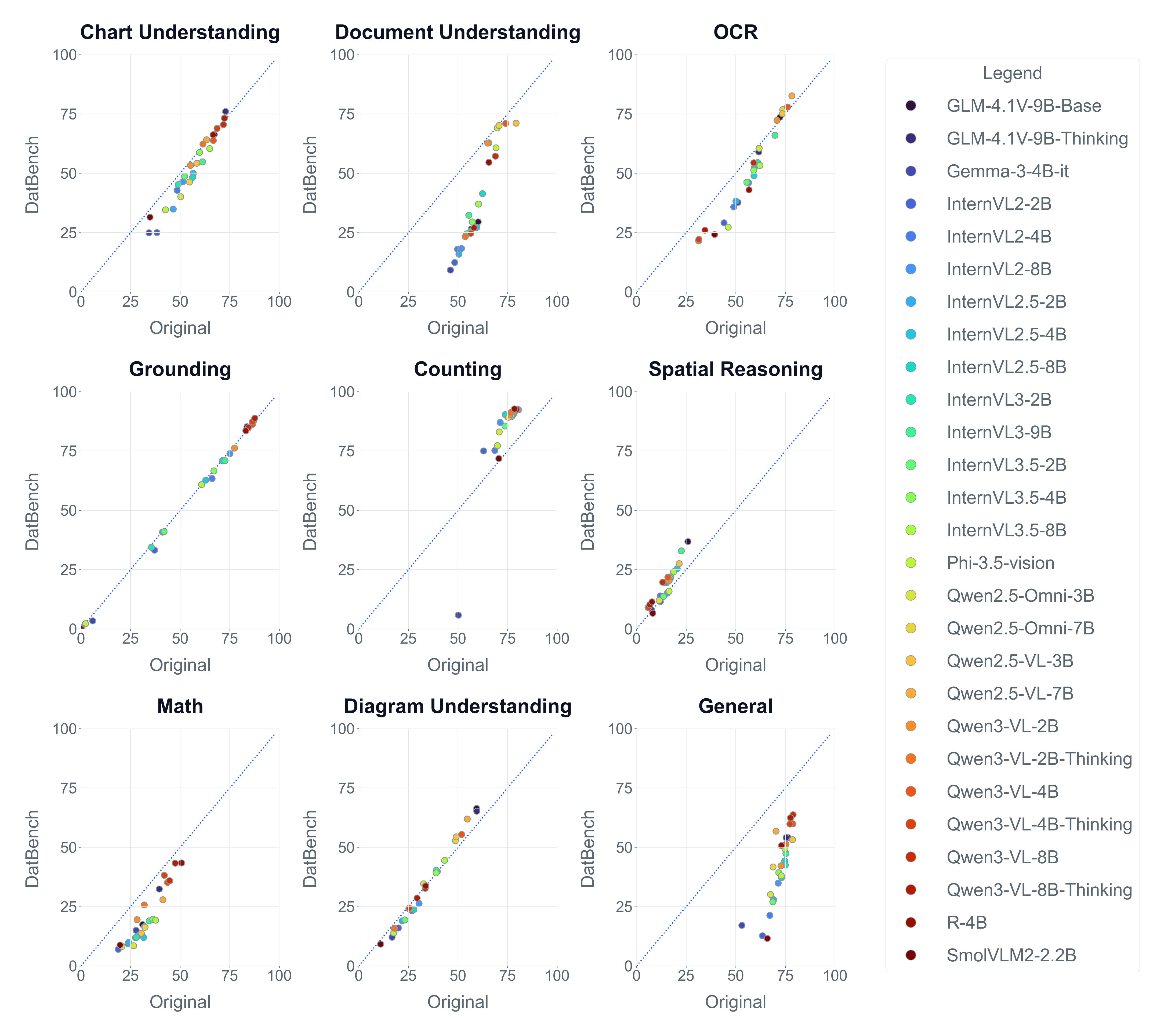}
    \caption{\textbf{\datbench vs. Original Performance.} We plot accuracy on \datbench ($y$-axis) against original baselines ($x$-axis) for 27 models, demonstrating the impact of our refinements on evaluative faithfulness and discrimination. Points below the diagonal indicate a more rigorous evaluation following the removal of non-discriminative and blind-solvable items, while a larger slope and higher dispersion—most notable in \textit{Document Understanding} and \textit{Math}—reveal a higher-resolution instrument that exposes latent differences between models previously masked by noise. Conversely, upward shifts in categories like \textit{Counting} reflect increased faithfulness achieved by purging incorrectly labeled artifacts that penalized correct reasoning. Despite these interventions, the consistency of model clusters across all nine plots confirms that our methodology establishes a discriminative and efficient evaluation that accurately captures model rankings while remaining sensitive to marginal performance gains.}
    \label{fig:big_table}
\end{figure}
\clearpage

Having established these artifacts, we provide a comprehensive statistical analysis of how \datbench transforms raw benchmark data into faithful and discriminative instruments for the efficient estimation of VLM capabilities. 

\paragraph{\datbench discards samples that are too easy / too hard.}
The most immediate impact of our filtering is the removal of samples that act as statistical noise. In the \textbf{General} capability, model performance is significantly shifted downward from the $y=x$ diagonal (Figure~\ref{fig:big_table}), a direct result of Stage 2 filtering (c.f. Appendix \ref{app:step2}) which discarded 72.07\% of samples solvable via language priors alone. Conversely, the \textbf{Spatial Reasoning} capability underwent rigorous quality filtering in Stage 3 (c.f. Appendix \ref{app:step3}), with 42.07\% of samples removed due to ambiguity or insufficient resolution. This systematic removal of evaluative noise shifts model assessments to a more faithful performance tier, ensuring that benchmark outcomes accurately reflect genuine multimodal reasoning.

\paragraph{\datbench is more discriminative.} Our item-selection methodology amplifies performance differences between models, increasing measurement resolution. On the original \textbf{General} benchmarks, models compress into a narrow 65–80\% accuracy band; on \datbench, they spread across 10–65\%, a nearly 4× expansion in effective score range. This ``stretching'' reflects our point-biserial selection criterion (Section 3.4): by retaining only items where strong models reliably succeed and weak models reliably fail, small capability differences that were previously masked now manifest as measurable gaps. The steep slopes observed in General and Document Understanding (Figure \ref{fig:big_table}) confirm this effect; equivalent spacing on the original benchmarks translates to greater separation on \datbench.

\paragraph{\datbench preserves discrimination power with far fewer samples.}
Despite aggressive filtering, ranking stability is maintained. For capabilities such as \textbf{Chart Understanding} and \textbf{Grounding}, \datbench points fall almost perfectly on the $y=x$ line (Figure~\ref{fig:big_table}), confirming that the subset preserves the discriminability and model rankings from the full dataset. As shown in our Stage 4 efficiency analysis (c.f. Appendix \ref{app:step4}), \datbench maintains high total discriminative power even at severely restricted budgets, whereas random sampling suffers from linear signal degradation.

\paragraph{Limitations and Future Directions.}
While our methodology offers a substantial leap forward, several avenues remain for future exploration:
\begin{itemize}[leftmargin=*, nosep]
    \item \textbf{Scaling to Larger Regimes:} Our current analysis focuses on models in the 1B--10B parameter range and inference traces within standard context windows. While the methodology is scale-invariant, the specific subsets of highly discriminative questions will likely shift for larger models and extended inference budgets (e.g., exceeding the 4096 tokens used in our work). Future work can apply this pipeline to larger model families and longer reasoning traces to identify the discriminative frontier for state-of-the-art systems.
    \item \textbf{Diversity Guarantees:} Our current subset selection prioritizes the highest individual discrimination scores, which implicitly relies on the inherent variety of the source data rather than an explicit diversity constraint. Consequently, this objective does not formally account for redundancy between samples; in pathological cases (e.g., duplicate but highly discriminative examples), the selection could theoretically yield a degenerate or repetitive subset. While we mitigate this through rigorous initial curation, future iterations could incorporate explicit diversity-aware objectives to ensure broader coverage of the capability space.
    
    \item \textbf{Expanding Capabilities:} We aim to extend our capability map beyond static images to include long-form video understanding, UI/GUI grounding, and robotics perception.
    
    \item \textbf{\textsc{DatBench-Live}:} Finally, discrimination is a moving target; questions that distinguish today's models will eventually become trivial. We envision a dynamic, ``living'' benchmark where subsets are recomputed periodically as new models shift the capability distribution and new datasets emerge.
\end{itemize}

\section{Diagnosing VLM Pathologies with \datbench}
\label{sec:insights}

In this section, we leverage the high-signal artifacts produced by the \datbench pipeline to diagnose the behavioral pathologies of modern VLMs. By analyzing performance across 27 state-of-the-art models spanning the 1B–10B parameter range, we uncover fundamental trade-offs between semantic reasoning and perceptual grounding, risks and rewards of inference-time scaling, and the impact of language priors on evaluation metrics.

\begin{figure}[h]
    \centering
    \vspace{-0.5em}
    \begin{subfigure}[b]{0.57\textwidth}
        \centering
        \includegraphics[width=\linewidth]{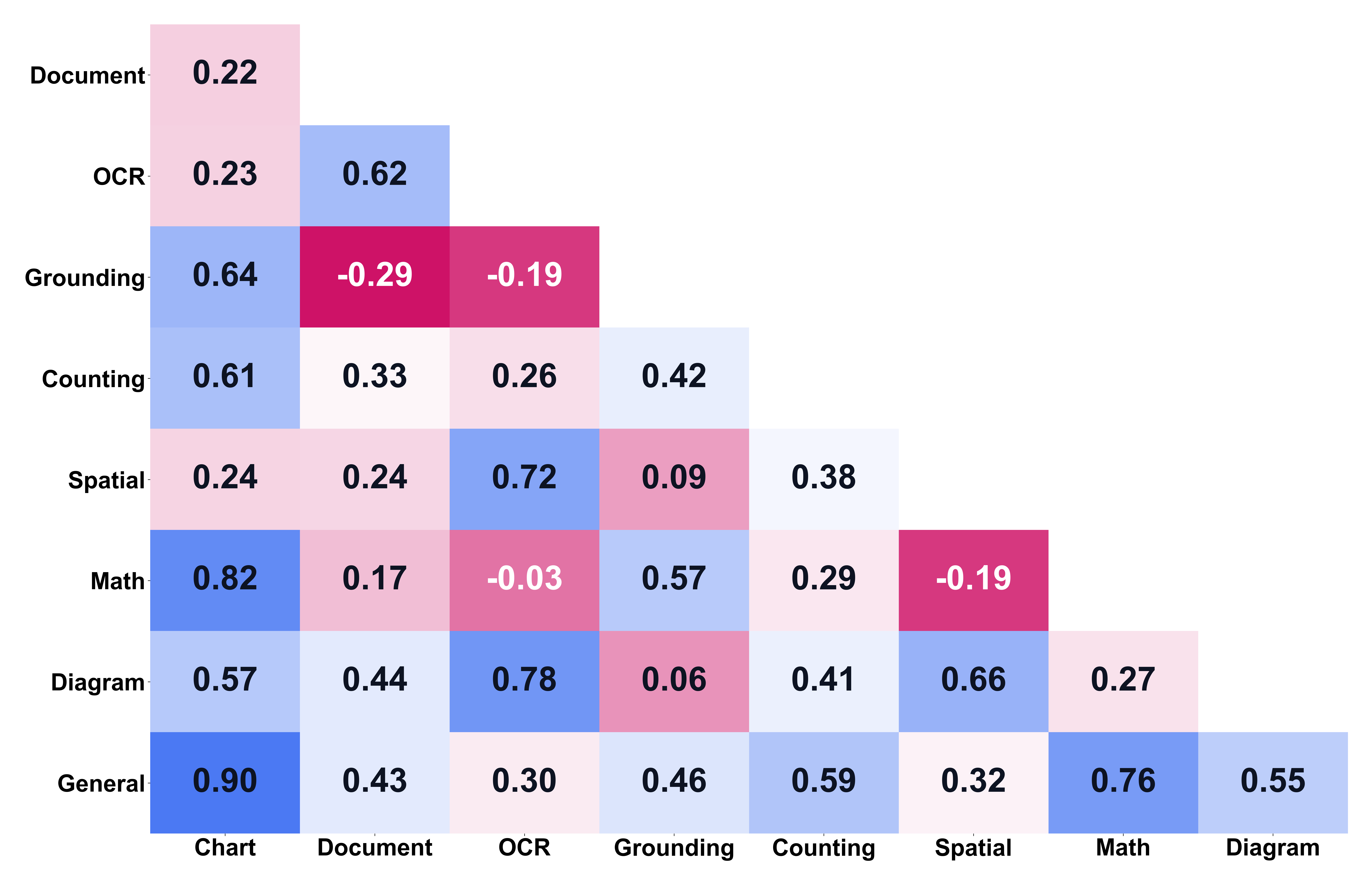}
        \vspace{-1.5em}
        \caption{Strong links exist between reasoning tasks, while spatial tasks often conflict with text-heavy ones.}
        \label{fig:capability_heatmap}
    \end{subfigure}
    \hfill
    \begin{subfigure}[b]{0.41\textwidth}
        \centering
        \includegraphics[width=\linewidth]{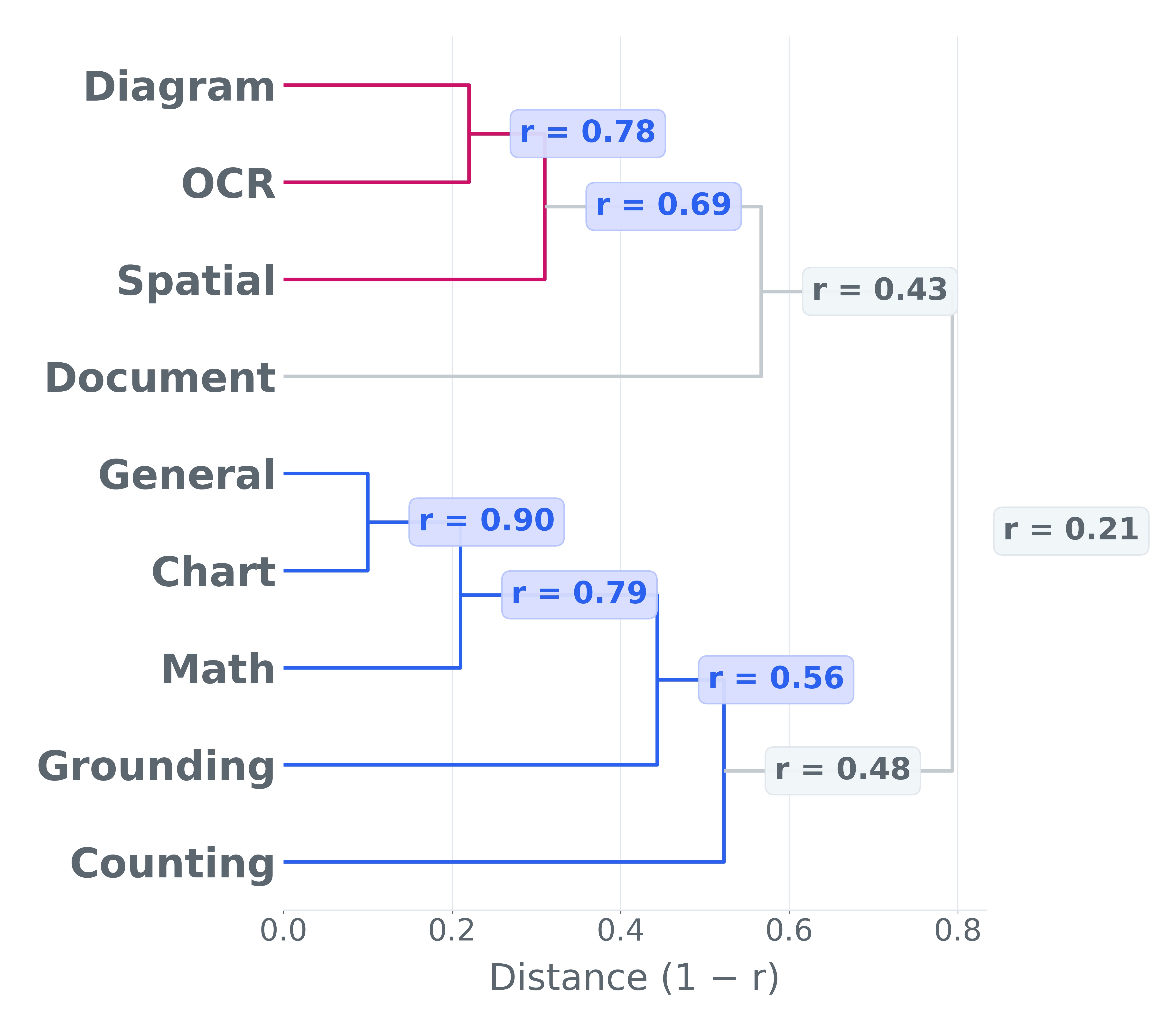}
        \vspace{-1.5em}
        \caption{Hierarchical clustering (average linkage, distance $= 1-r$) reveals two main clusters: reasoning (Chart, Math, General) and perception (OCR, Spatial, Diagram).}
        \label{fig:capability_dendrogram}
    \end{subfigure}
    \vspace{-0.5em}
    \caption{\textbf{Correlation analysis of model capabilities across 26 vision-language models.} Pairwise Pearson correlations are computed between mean accuracy scores for each capability. (a)~Capability correlations. (b)~Capability clustering.}
    \label{fig:capability_analysis}
    \vspace{-1em}
\end{figure}

\paragraph{Takeaway 1: Capability Correlations Reveal a "Reasoning vs. Perception" Trade-off.}
To identify hidden relationships between tasks, we calculated Pearson correlations ($r$) between all capability scores across our model suite (Figure \ref{fig:capability_heatmap}). We identify a tight \textbf{Reasoning Cluster} in which \textit{Chart Understanding}, \textit{Math}, and \textit{General VQA} exhibit exceptionally high pairwise correlations, such as $r=0.90$ between Chart and General tasks. This analysis confirms that \textit{General} VQA benchmarks, such as MMBench and MMMU-Pro, primarily test abstract reasoning capabilities that are also fundamental to \textit{Math}, evidenced by a strong correlation of $r=0.76$ between these two domains. Furthermore, a distinct \textbf{Spatial-Semantic Trade-off} exists: \textit{Grounding} correlates negatively with text-heavy tasks like \textit{Document Understanding} ($r=-0.29$) and \textit{OCR} ($r=-0.19$). These negative relationships, alongside the inverse correlation between Math and \textit{Spatial Reasoning} ($r=-0.19$), suggest a latent conflict in current training paradigms between high-level semantic processing and low-level perceptual fidelity. Hierarchical clustering (Figure \ref{fig:capability_dendrogram}) corroborates this dichotomy, revealing two distinct clusters: reasoning (Chart, Math, General) and perception (OCR, Spatial, Diagram).

\begin{figure}[h]
    \centering
        \centering
        \includegraphics[width=0.5\linewidth]{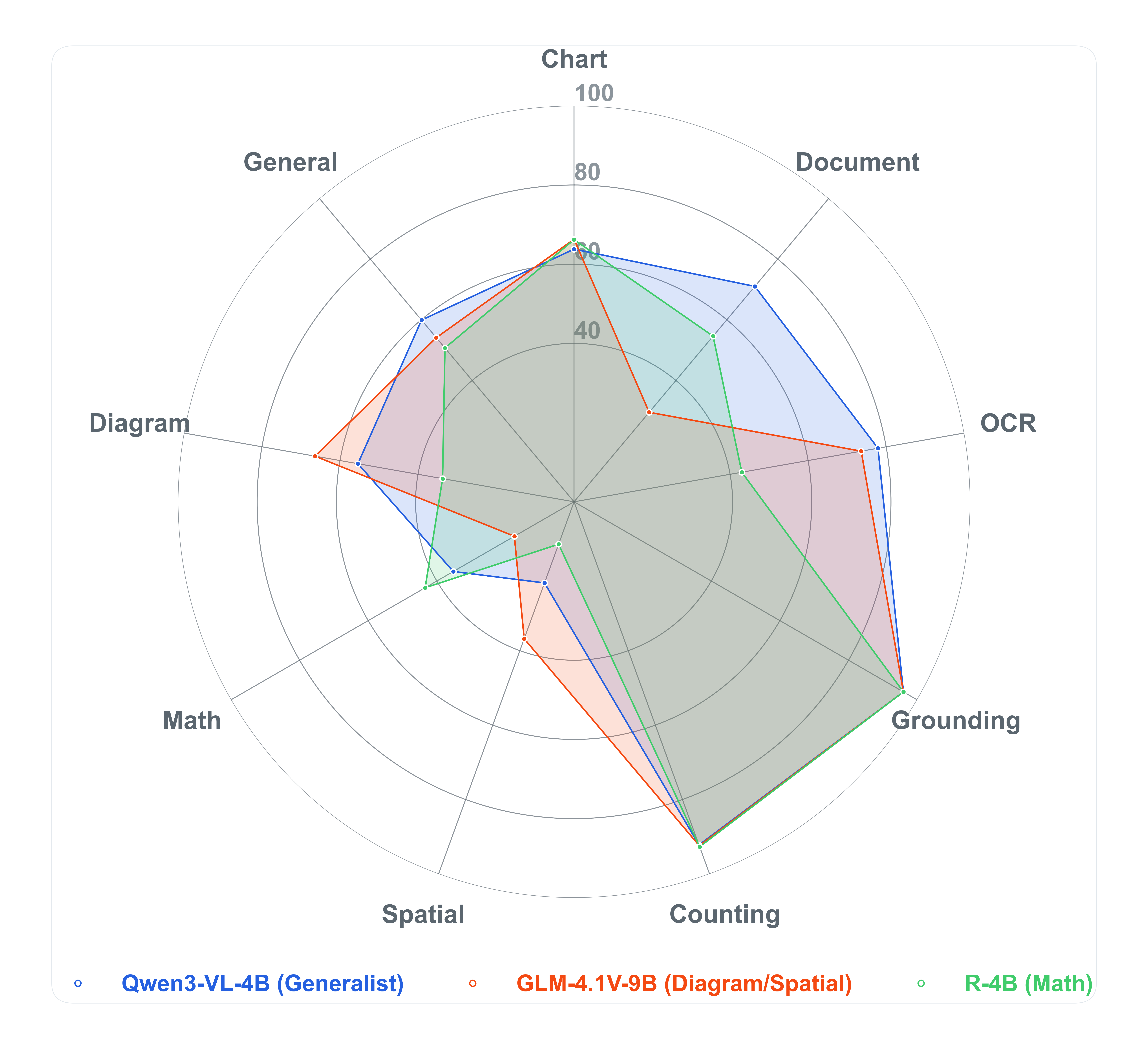}
        \caption{\textbf{Capability Profiles.} Models show clear trade-offs between general reasoning and specialized visual tasks.}
        \label{fig:radar}
\end{figure}

\paragraph{Takeaway 2: Capability Profiles Reveal Specialist-Generalist Trade-offs.} This trade-off manifests in distinct model archetypes (Figure~\ref{fig:radar}). \textit{GLM-4.1V-9B} acts as a perception specialist, leading in diagram understanding (66.4\%) and spatial reasoning (36.8\%) but struggling with math (17.4\%). Balanced generalists are rare: \textit{Qwen3-VL-4B} is a notable exception, maintaining strong document understanding (71.0\%), OCR (77.9\%), and reasoning (59.9\%). Most tellingly, \textit{R-4B} reaches the highest math score (43.4\%) at the cost of the lowest spatial performance (11.4\%), suggesting that (current) reasoning-focused training can degrade visual grounding.

\begin{figure}[h]
    \centering
    \begin{subfigure}[b]{0.48\textwidth}
        \centering
        \includegraphics[width=\linewidth]{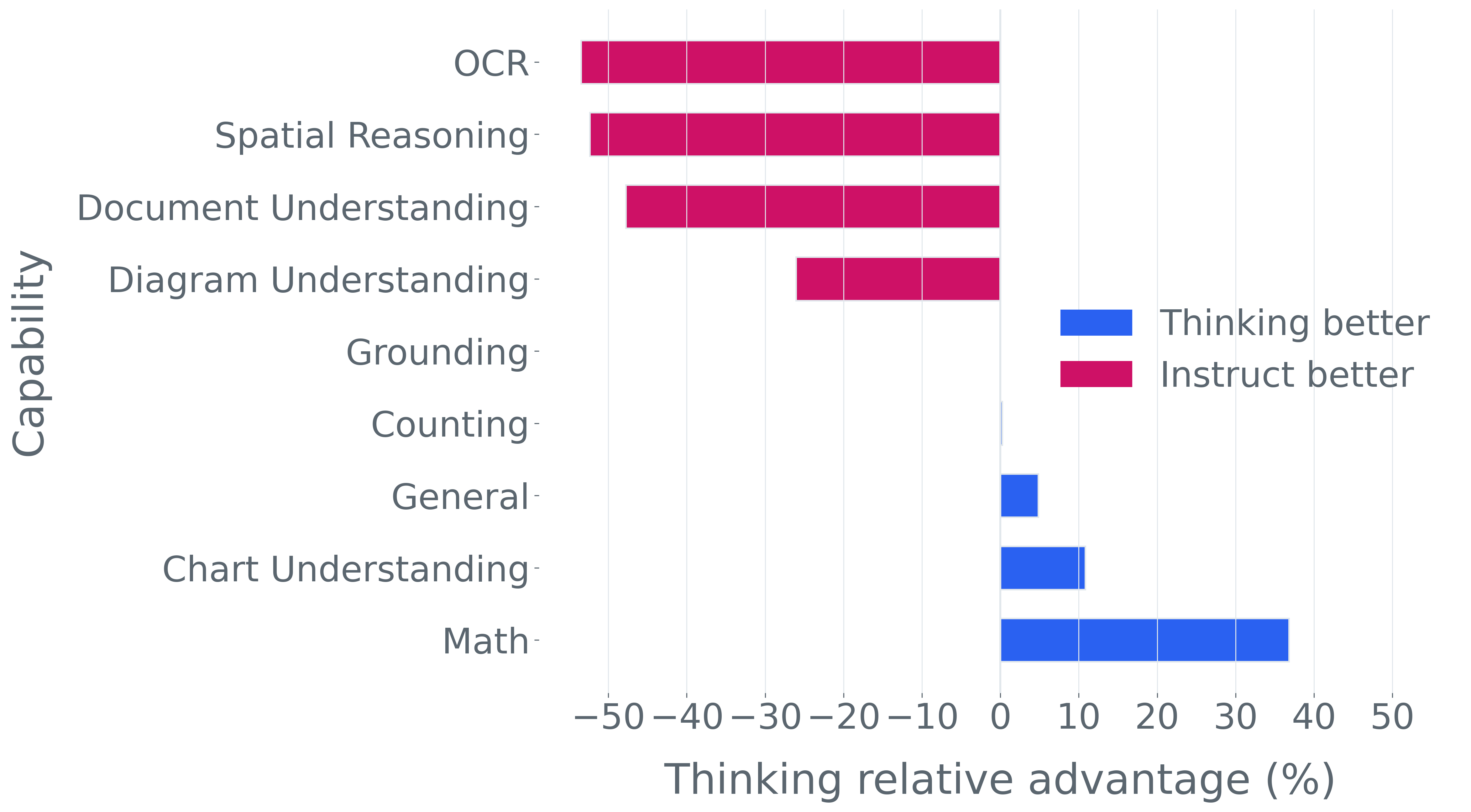}
        \caption{Inference Scaling Impact}
        \label{fig:thinking_vs_nothinking}
    \end{subfigure}
    \hfill
    \begin{subfigure}[b]{0.48\textwidth}
        \centering
        \includegraphics[width=\linewidth]{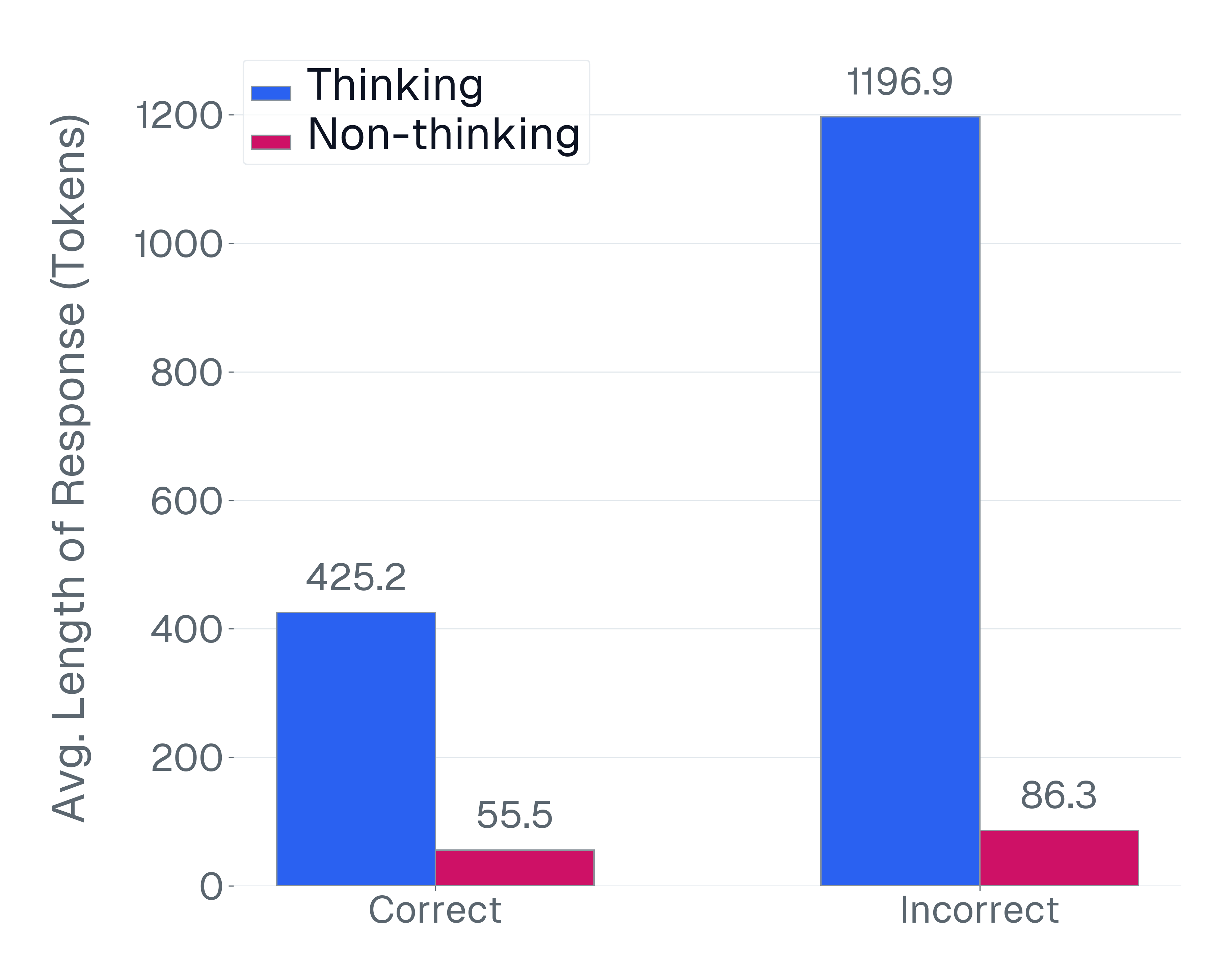}
        \caption{Computational Cost of Errors}
        \label{fig:cost_of_overthinking}
    \end{subfigure}
    \caption{\textbf{The Overthinking Penalty.} (a) Scaling compute helps reasoning but hurts perception. (b) Incorrect "thinking" responses use $\approx 14\times$ more tokens than standard models.}
    \label{fig:thinking_analysis_combined}
\end{figure}

\paragraph{Takeaway 3: The "Overthinking" Penalty: Inference-Time Scaling Degrades Perception at High Cost.}
Comparing \textit{Thinking} models to standard counterparts reveals that extra test-time compute is a double-edged sword (Figure \ref{fig:thinking_vs_nothinking}). To quantify this, we define the Thinking relative advantage as the percentage gain in accuracy of the thinking model over its instruct counterpart, normalized by the instruct baseline: $(Acc_{thinking} - Acc_{instruct}) / Acc_{instruct} \times 100$. Scaling helps \textit{Math} ($\approx +36.8\%$) and \textit{Charts} ($\approx +10.8\%$) but causes massive regressions in \textit{OCR} ($\approx -53.5\%$) and \textit{Document Understanding} ($\approx -47.8\%$). This regression is also extremely computationally wasteful: while correct \textit{thinking} answers use $\approx425$ tokens, incorrect attempts balloon to $\approx1196.9$ tokens, a $\approx 14\times$ increase over non-thinking models (Figure \ref{fig:cost_of_overthinking}). We observed that this is due to models entering \textit{unproductive thinking} loops on perceptual tasks they cannot solve. Prior work has observed a similar \textit{overthinking penalty} for language models \citep{overthink_hochlehnert2025soberlookprogresslanguage, overthink_su2025underthinkingoverthinkingempiricalstudy, overthink_wang2025thoughtsplaceunderthinkingo1like, overthink_wu2025lessunderstandingchainofthoughtlength}. 

\begin{figure}[h]
    \centering
    \includegraphics[width=0.5\linewidth]{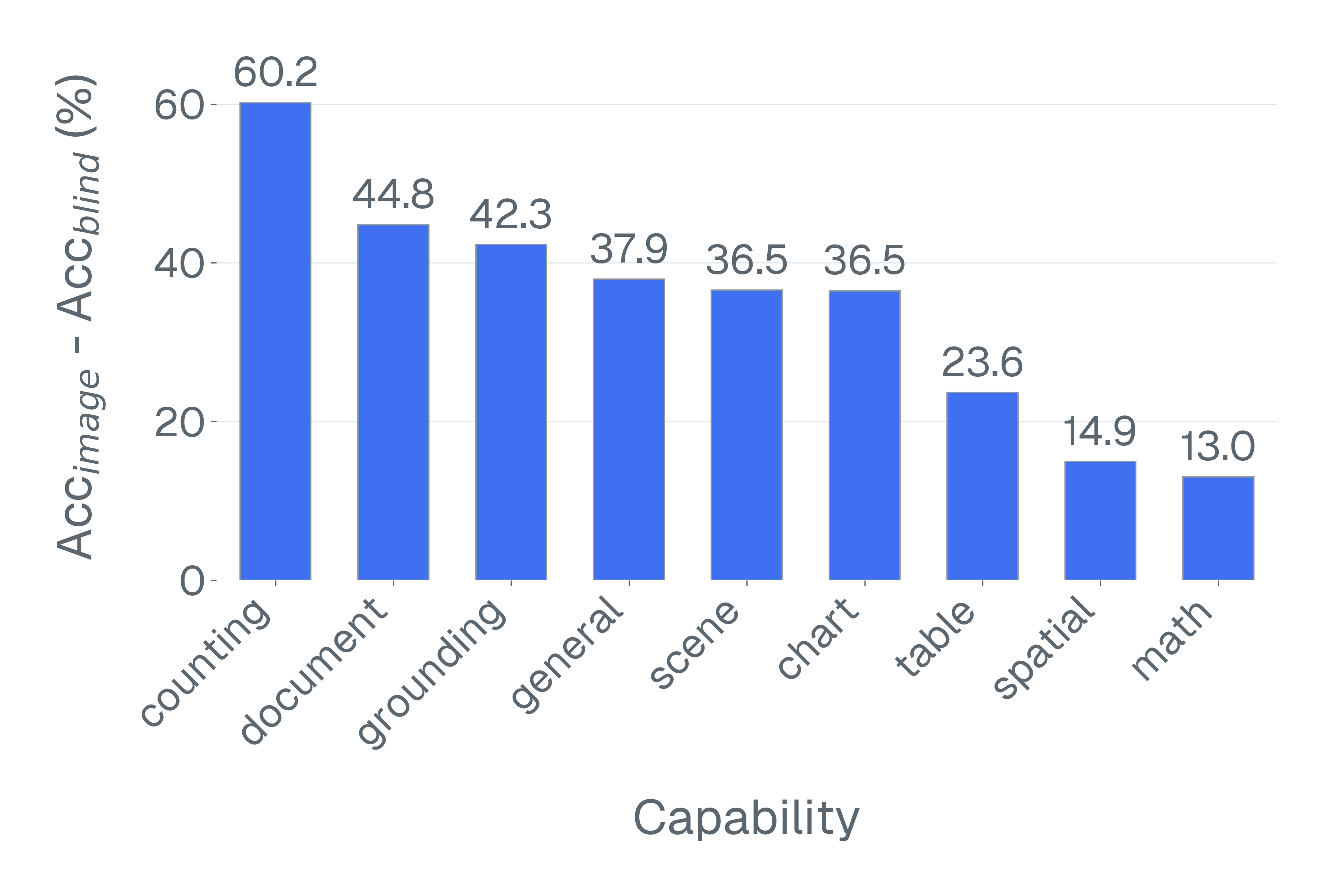}
    \caption{\textbf{Vision vs. Text Priors.} Tasks like Counting require seeing the image, whereas Math can often be solved by language prior alone.}
    \label{fig:vision_delta_capability}
\end{figure}

\paragraph{Takeaway 4: Language Priors Mask True Multimodal Performance across Capabilities.}
To isolate the actual visual requirement of each task, we analyze the \textit{vision delta} ($V_{\Delta}$), defined as the performance gap between standard multimodal evaluation and a \textit{blind} text-only baseline (Figure \ref{fig:vision_delta_capability}). Our results show that reliance on language priors varies drastically by capability, often distorting perceived progress in multimodal reasoning. Capabilities such as \textit{Counting} ($V_{\Delta} = 60.2\%$) and \textit{Grounding} ($V_{\Delta} = 42.3\%$) exhibit high vision dependency, making them the most faithful indicators of true perceptual accuracy. Conversely, \textit{Math} ($V_{\Delta} = 13.0\%$) and \textit{Spatial Reasoning} ($V_{\Delta} = 14.9\%$) show significant language prior distortion, relying heavily on textual patterns that allow models to guess correctly without the image. These findings confirm that without the rigorous filtering introduced in \datbench, i.e. discarding samples that can be solved with the language prior alone, high scores in capabilities like \textit{Math} may inadvertently reward stronger language models rather than superior vision-language integration.

\section{Conclusion}

In this work, we addressed the dual challenges of data quality and computational cost in the evaluation of Vision-Language Models (VLMs). We introduced a framework of three desiderata that evaluations should satisfy: (1) \textbf{faithfulness} to the modality and application, (2) \textbf{discriminability} between models of varying quality, and (3) \textbf{efficiency} in compute. We then applied this lens to expose four critical pathologies in existing benchmarks: multiple-choice formats are both unfaithful and weakly discriminative; many VLM benchmarks can be solved without vision; incorrect and ambiguous ground truth introduces substantial noise; and existing evaluation suites are inefficient. We used these insights to distill these benchmarks into high-signal evaluation suites.

Our primary contribution, \datbench, serves as a precise, psychometrically grounded instrument for measuring multimodal capability. Motivated by Item Response Theory (IRT) and operationalizing discrimination via point‑biserial correlation (rpb), we demonstrated that maximizing \textit{total test discrimination} yields subsets that are not only computationally lightweight but also significantly more robust and generalizable than those derived via random sampling or simple rank correlation. Our accompanying analysis of ``thinking'' models and language priors further validates that \datbench is capable of surfacing nuanced behavioral insights that are often obscured in aggregate metrics. We release two versions of the benchmark, the efficiency-focused \textsc{DatBench} for rapid iterative development (yielding 13$\times$ average speedup), and the comprehensive \textsc{DatBench-Full} for final reporting, to standardize comparison and accelerate progress at the pareto frontier. Our work provides a path towards evaluation practices that are both rigorous and sustainable as VLMs continue to scale.

\section{Contributions and Acknowledgements}
\label{sec:contri}

\begin{tabularx}{\textwidth}{@{}p{0.19\textwidth}X@{}}
\textbf{Core Contributors} & Siddharth Joshi, Haoli Yin, Rishabh Adiga, and Ricardo Monti. \\[0.25em]
& \emph{for fusing modalities, wrangling the datasets, and ensuring the evaluation pipeline didn't hallucinate.} \\
\noalign{\vspace{0.75em}}

\textbf{Technical Contributors} & Aldo Carranza, Alex Fang, Alvin Deng, Amro Abbas, Brett Larsen, Cody Blakeney, Darren Teh, David Schwab, Fan Pan, Haakon Mongstad, Jack Urbanek, Jason Lee, Jason Telanoff, Josh Wills, Kaleigh Mentzer, Luke Merrick, Parth Doshi, Paul Burstein, Pratyush Maini, Scott Loftin, Spandan Das, Tony Jiang, Vineeth Dorna, and Zhengping Wang. \\
\noalign{\vspace{0.25em}}
& \emph{the ensemble of experts who maximized our few-shot performance and ablated every hyperparameter.}\\
\noalign{\vspace{0.75em}}

\textbf{Leadership} & Bogdan Gaza, Ari Morcos, and Matthew Leavitt. \\[0.25em]
& \emph{the ground truth oracles who steered the project and prevented collective mode collapse.} \\
\noalign{\vspace{0.75em}}

\textbf{Acknowledgements} & Liz Gatapia (\textit{for incredible logo design}),
Jacqueline Liu, Tiffanie Pham, Sylvia Hoang, Jayla Lindsey, Kylie Clement, Elise Clark \\
\noalign{\vspace{0.25em}} 
& \emph{the human-in-the-loop feedback that provided essential regularization and support.}
\end{tabularx}

\clearpage
\bibliographystyle{abbrvnat}  
\bibliography{references}

\clearpage
\appendix
\appendix

\clearpage
\section{Main Results}\label{app:main_results}

\begin{table*}[h]
\centering
\tiny 
\renewcommand{\arraystretch}{1.15}
\setlength{\tabcolsep}{1.5pt} 
\caption{   \textbf{Comprehensive Evaluation.} Comparison across \textsc{DatBench} (DB), \textsc{DatBench-Full} (Full), and the Original datasets (Orig). Values are percentages.}
\label{tab:giant_comparison}

\resizebox{\textwidth}{!}{%
\begin{tabular}{@{}l ccc ccc ccc ccc ccc ccc ccc ccc ccc@{}}
\toprule
& \multicolumn{3}{c}{\textbf{Chart}} & \multicolumn{3}{c}{\textbf{Doc}} & \multicolumn{3}{c}{\textbf{OCR}} & \multicolumn{3}{c}{\textbf{Grd}} & \multicolumn{3}{c}{\textbf{Cnt}} & \multicolumn{3}{c}{\textbf{Spa}} & \multicolumn{3}{c}{\textbf{Mth}} & \multicolumn{3}{c}{\textbf{Dia}} & \multicolumn{3}{c}{\textbf{Gen}} \\
\cmidrule(lr){2-4} \cmidrule(lr){5-7} \cmidrule(lr){8-10} \cmidrule(lr){11-13} \cmidrule(lr){14-16} \cmidrule(lr){17-19} \cmidrule(lr){20-22} \cmidrule(lr){23-25} \cmidrule(lr){26-28}
\textbf{Model} & DB & Full & Orig & DB & Full & Orig & DB & Full & Orig & DB & Full & Orig & DB & Full & Orig & DB & Full & Orig & DB & Full & Orig & DB & Full & Orig & DB & Full & Orig \\
\midrule

\multicolumn{28}{l}{\textit{\textbf{Qwen2.5-VL}}} \\
~Qwen2.5-VL-3B-Instruct & 54.2 & 56.2 & 58.4 & 71.1 & 78.5 & 79.3 & 75.1 & 70.2 & 73.5 & 0.3 & 0.2 & 0.4 & 90.2 & 71.8 & 77.0 & 27.5 & 27.5 & 21.5 & 13.8 & 14.9 & 30.4 & 54.4 & 51.1 & 49.1 & 53.2 & 59.8 & 78.6 \\
~Qwen2.5-VL-7B-Instruct & 64.1 & 63.8 & 63.3 & 62.8 & 66.3 & 65.6 & 82.6 & 75.4 & 78.3 & 0.2 & 0.3 & 0.4 & 92.4 & 76.5 & 80.4 & 20.8 & 20.8 & 16.7 & 27.9 & 26.1 & 41.3 & 61.9 & 56.2 & 54.7 & 56.8 & 62.1 & 70.3 \\

\multicolumn{28}{l}{\textit{\textbf{Qwen3-VL}}} \\
~Qwen3-VL-2B-Instruct & 53.3 & 54.8 & 55.2 & 62.7 & 62.6 & 65.0 & 72.3 & 67.9 & 70.8 & 76.2 & 75.6 & 77.4 & 90.3 & 70.8 & 76.2 & 20.3 & 20.3 & 16.3 & 19.5 & 17.6 & 28.3 & 24.1 & 23.1 & 25.3 & 42.2 & 51.5 & 72.8 \\
~Qwen3-VL-4B-Instruct & 63.8 & 63.5 & 66.7 & 71.0 & 71.6 & 74.1 & 77.9 & 72.3 & 76.1 & 86.2 & 85.7 & 86.4 & 92.2 & 74.6 & 79.1 & 21.8 & 21.8 & 15.8 & 35.2 & 30.7 & 43.6 & 55.4 & 51.6 & 51.9 & 59.9 & 59.4 & 78.8 \\
~Qwen3-VL-8B-Instruct & 70.5 & 68.4 & 71.8 & 57.2 & 63.5 & 68.9 & 54.4 & 54.7 & 59.1 & 84.9 & 83.7 & 84.4 & 92.6 & 75.4 & 80.0 & 19.7 & 19.7 & 13.2 & 35.9 & 31.2 & 44.7 & 32.7 & 31.3 & 33.5 & 63.7 & 59.8 & 78.9 \\

\multicolumn{28}{l}{\textit{\textbf{Qwen2.5-Omni}}} \\
~Qwen2.5-Omni-3B & 40.1 & 44.9 & 50.3 & 69.1 & 71.3 & 69.8 & 60.5 & 58.2 & 61.8 & 0.4 & 0.4 & 0.5 & 83.0 & 64.0 & 70.8 & 12.0 & 12.0 & 11.4 & 8.5 & 10.8 & 26.5 & 34.6 & 31.0 & 32.7 & 30.1 & 45.8 & 67.5 \\
~Qwen2.5-Omni-7B & 46.3 & 50.1 & 54.7 & 70.1 & 74.7 & 70.7 & 76.8 & 70.6 & 73.7 & 0.5 & 0.4 & 0.5 & 89.2 & 69.8 & 75.5 & 21.2 & 21.2 & 17.0 & 16.3 & 17.5 & 32.3 & 52.8 & 48.1 & 48.7 & 41.7 & 49.7 & 68.8 \\

\multicolumn{28}{l}{\textit{\textbf{InternVL2}}} \\
~InternVL2-2B & 25.0 & 32.5 & 38.3 & 12.4 & 39.2 & 48.4 & 29.1 & 35.8 & 44.1 & 33.2 & 32.9 & 37.1 & 75.0 & 64.8 & 62.9 & 11.5 & 11.5 & 12.0 & 7.0 & 9.8 & 18.8 & 16.0 & 17.2 & 20.0 & 12.7 & 41.7 & 63.5 \\
~InternVL2-4B & 42.8 & 47.2 & 48.5 & 18.0 & 39.1 & 49.8 & 35.8 & 40.3 & 49.0 & 63.4 & 62.6 & 66.1 & 75.2 & 65.3 & 68.6 & 13.9 & 13.9 & 11.9 & 9.4 & 11.5 & 23.7 & 23.2 & 23.0 & 26.8 & 21.3 & 45.3 & 67.2 \\
~InternVL2-8B & 46.4 & 50.4 & 51.3 & 18.3 & 41.6 & 51.8 & 46.0 & 47.7 & 56.5 & 73.9 & 73.0 & 75.1 & 87.0 & 68.6 & 71.3 & 19.4 & 19.4 & 14.6 & 11.9 & 14.4 & 27.5 & 26.3 & 26.3 & 30.4 & 34.9 & 50.9 & 71.4 \\

\multicolumn{28}{l}{\textit{\textbf{InternVL2.5}}} \\
~InternVL2\_5-2B & 34.9 & 40.7 & 46.5 & 15.9 & 42.3 & 50.5 & 38.3 & 43.0 & 50.1 & 40.8 & 39.9 & 41.1 & 89.3 & 71.4 & 75.1 & 15.1 & 15.1 & 15.5 & 9.9 & 11.8 & 24.0 & 19.0 & 20.5 & 21.9 & 27.9 & 47.9 & 69.1 \\
~InternVL2\_5-4B & 50.0 & 52.8 & 56.6 & 27.2 & 51.0 & 59.5 & 52.4 & 52.1 & 59.4 & 62.7 & 60.7 & 62.9 & 90.3 & 72.6 & 77.9 & 25.4 & 25.4 & 20.5 & 12.0 & 13.9 & 31.6 & 39.8 & 37.9 & 39.6 & 37.2 & 49.7 & 73.1 \\
~InternVL2\_5-8B & 48.2 & 51.5 & 56.2 & 41.4 & 57.9 & 62.4 & 49.0 & 51.5 & 59.2 & 70.9 & 69.2 & 71.3 & 90.3 & 72.7 & 73.8 & 21.8 & 21.8 & 16.9 & 12.7 & 14.6 & 28.6 & 23.8 & 23.1 & 27.8 & 44.3 & 55.4 & 74.7 \\

\multicolumn{28}{l}{\textit{\textbf{InternVL3}}} \\
~InternVL3-2B-Instruct & 45.2 & 48.9 & 49.0 & 25.7 & 50.2 & 56.5 & 54.5 & 55.3 & 61.1 & 34.4 & 33.4 & 35.5 & 89.5 & 71.9 & 77.1 & 22.0 & 22.0 & 17.3 & 12.1 & 13.7 & 27.9 & 24.0 & 24.0 & 26.3 & 42.5 & 54.3 & 75.0 \\
~InternVL3-9B-Instruct & 54.8 & 56.1 & 61.3 & 32.3 & 56.4 & 55.5 & 66.0 & 63.8 & 69.8 & 70.9 & 69.3 & 72.6 & 91.4 & 74.5 & 78.5 & 32.9 & 32.9 & 22.6 & 19.0 & 19.5 & 34.4 & 40.3 & 39.6 & 39.1 & 47.4 & 54.6 & 75.3 \\

\multicolumn{28}{l}{\textit{\textbf{InternVL3.5}}} \\
~InternVL3\_5-2B-Instruct & 48.7 & 51.4 & 52.2 & 24.5 & 46.7 & 54.4 & 46.2 & 48.6 & 55.6 & 41.0 & 40.3 & 41.9 & 85.5 & 67.6 & 73.7 & 13.7 & 13.7 & 13.6 & 16.6 & 17.0 & 32.0 & 19.4 & 19.2 & 23.3 & 27.0 & 46.2 & 68.7 \\
~InternVL3\_5-4B-Instruct & 58.8 & 59.5 & 59.8 & 29.5 & 51.3 & 57.2 & 51.3 & 51.3 & 59.1 & 66.6 & 65.5 & 67.0 & 90.6 & 71.3 & 76.8 & 21.4 & 21.4 & 16.7 & 19.7 & 19.7 & 36.3 & 39.2 & 36.6 & 39.0 & 39.3 & 49.8 & 71.7 \\
~InternVL3\_5-8B-Instruct & 60.4 & 61.1 & 64.9 & 37.0 & 56.2 & 60.4 & 53.3 & 54.2 & 62.2 & 60.8 & 59.1 & 60.8 & 91.0 & 72.5 & 77.8 & 24.1 & 24.1 & 18.7 & 19.3 & 19.9 & 37.6 & 44.5 & 41.1 & 43.3 & 49.3 & 54.6 & 74.8 \\

\multicolumn{28}{l}{\textit{\textbf{Other Models}}} \\
~GLM-4.1V-9B-Base & 66.3 & 65.5 & 67.2 & 29.5 & 48.6 & 60.2 & 73.6 & 69.0 & 72.3 & 85.2 & 83.3 & 83.7 & 92.4 & 75.6 & 80.0 & 36.8 & 36.8 & 25.9 & 17.4 & 16.5 & 31.1 & 66.4 & 60.0 & 59.4 & 54.1 & 56.6 & 76.3 \\
~SmolVLM2-2.2B-Instruct & 31.5 & 37.1 & 34.8 & -- & -- & -- & 24.2 & 30.3 & 39.4 & 0.0 & 0.0 & 0.1 & 71.8 & 65.2 & 70.6 & 6.6 & 6.6 & 8.2 & 8.8 & 11.6 & 19.7 & 9.2 & 7.5 & 11.0 & 11.6 & 41.8 & 65.9 \\
~Phi-3.5-vision-instruct & 34.6 & 40.3 & 42.6 & 60.7 & 68.3 & 69.2 & 27.3 & 35.0 & 46.2 & 2.2 & 1.9 & 2.3 & 77.2 & 65.3 & 69.9 & 15.9 & 15.9 & 16.4 & 8.2 & 10.8 & 20.7 & 13.9 & 12.7 & 17.6 & 38.0 & 50.2 & 73.0 \\
~Gemma-3-4B-it & 24.9 & 31.7 & 34.3 & 9.2 & 37.1 & 46.2 & 37.7 & 43.3 & 51.1 & 3.4 & 3.4 & 5.9 & 5.8 & 51.5 & 50.2 & 8.1 & 8.1 & 7.6 & 15.0 & 15.2 & 27.8 & 12.1 & 11.4 & 16.8 & 17.1 & 41.6 & 53.1 \\

\midrule
\multicolumn{28}{c}{\textsc{Thinking Models}} \\
\midrule

~GLM-4.1V-9B-Thinking & 76.0 & 73.8 & 72.9 & 26.5 & 44.3 & 56.7 & 59.0 & 55.2 & 61.6 & 87.9 & 86.3 & 87.2 & 92.3 & 76.7 & 80.5 & 19.4 & 19.4 & 15.1 & 32.4 & 28.5 & 39.5 & 65.2 & 60.2 & 59.5 & 54.1 & 54.0 & 75.4 \\
~R-4B & 66.2 & 65.4 & 66.6 & 54.6 & 60.9 & 65.6 & 43.0 & 47.1 & 56.7 & 83.5 & 81.7 & 83.1 & 92.7 & 74.7 & 78.5 & 11.4 & 11.4 & 7.8 & 43.4 & 37.8 & 50.7 & 33.7 & 30.4 & 33.7 & 50.7 & 52.2 & 73.0 \\
~Qwen3-VL-2B-Thinking & 62.3 & 61.9 & 61.5 & 23.3 & 45.5 & 53.7 & 21.5 & 22.2 & 31.3 & 84.6 & 83.2 & 84.3 & 91.2 & 72.8 & 76.7 & 9.0 & 9.0 & 5.8 & 25.7 & 22.6 & 31.9 & 15.9 & 13.3 & 18.0 & 51.3 & 55.2 & 75.3 \\
~Qwen3-VL-4B-Thinking & 68.9 & 66.9 & 68.6 & 24.7 & 46.9 & 56.5 & 22.1 & 22.7 & 31.4 & 87.3 & 85.8 & 86.4 & 92.3 & 74.4 & 79.1 & 9.1 & 9.1 & 6.1 & 38.2 & 33.0 & 42.0 & 24.4 & 21.0 & 25.4 & 59.8 & 57.4 & 77.3 \\
~Qwen3-VL-8B-Thinking & 73.2 & 70.7 & 72.3 & 27.0 & 48.6 & 58.1 & 26.0 & 25.6 & 34.5 & 88.8 & 87.1 & 87.6 & 92.5 & 75.6 & 79.6 & 10.2 & 10.2 & 6.7 & 43.3 & 37.4 & 47.5 & 28.6 & 25.7 & 29.4 & 62.4 & 58.2 & 77.6 \\

\bottomrule
\end{tabular}%
}
\end{table*}

\clearpage

\clearpage
\section{Benchmark Computational Cost}\label{app:cost}

\begin{table*}[htbp]
\centering
\footnotesize
\begin{tabular}{lccccccccc}
\toprule
\textbf{Model} & \textbf{Chart} & \textbf{Count} & \textbf{Doc} & \textbf{Gen} & \textbf{Ground} & \textbf{Math} & \textbf{Scene} & \textbf{Spat} & \textbf{Table} \\
\midrule
\textbf{Samples} & \textit{\footnotesize 12,249} & \textit{\footnotesize 39,080} & \textit{\footnotesize 118,581} & \textit{\footnotesize 246,475} & \textit{\footnotesize 36,961} & \textit{\footnotesize 12,368} & \textit{\footnotesize 38,950} & \textit{\footnotesize 40,131} & \textit{\footnotesize 32,753} \\
\midrule
SmolVLM2-2.2B & 0.35 & 1.11 & --- & 9.02 & 0.74 & 0.71 & 1.05 & 0.60 & 0.38 \\
InternVL2-2B & 0.30 & 0.77 & 2.36 & 5.97 & 0.59 & 0.71 & 0.88 & 0.56 & 0.35 \\
InternVL2-4B & 0.34 & 0.78 & 3.13 & 6.28 & 0.79 & 1.28 & 0.89 & 0.50 & 0.31 \\
InternVL2-8B & 0.38 & 0.95 & 3.51 & 8.03 & 0.84 & 1.58 & 1.10 & 0.58 & 0.37 \\
InternVL2.5-2B & 0.28 & 0.70 & 2.13 & 6.38 & 0.64 & 0.70 & 0.82 & 0.55 & 0.34 \\
InternVL2.5-4B & 0.29 & 0.70 & 2.45 & 5.90 & 0.71 & 0.92 & 0.84 & 0.51 & 0.34 \\
InternVL2.5-8B & 0.39 & 0.95 & 3.71 & 8.07 & 0.84 & 1.28 & 1.04 & 0.60 & 0.38 \\
InternVL3-2B & 0.34 & 0.63 & 2.05 & 5.55 & 0.77 & 0.69 & 0.80 & 0.54 & 0.34 \\
InternVL3-9B & 0.46 & 1.00 & 4.02 & 8.88 & 1.00 & 1.61 & 1.13 & 0.61 & 0.41 \\
InternVL3.5-2B & 0.45 & 0.68 & 2.47 & 6.66 & 0.67 & 1.40 & 0.91 & 0.66 & 0.36 \\
InternVL3.5-4B & 0.60 & 0.77 & 3.12 & 8.01 & 0.84 & 1.31 & 0.99 & 0.56 & 0.38 \\
InternVL3.5-8B & 0.80 & 0.88 & 3.89 & 12.86 & 0.98 & 2.62 & 1.03 & 0.68 & 0.40 \\
Qwen2.5-Omni-3B & 0.29 & 0.56 & 2.35 & 4.68 & 0.64 & 0.51 & 1.18 & 0.74 & 0.46 \\
Qwen2.5-Omni-7B & 0.35 & 0.55 & 3.06 & 6.15 & 0.77 & 1.11 & 1.26 & 0.80 & 0.46 \\
Qwen2.5-VL-3B & 0.29 & 0.71 & 2.52 & 4.65 & 0.63 & 1.19 & 1.19 & 0.70 & 0.43 \\
Qwen2.5-VL-7B & 0.56 & 0.76 & 3.42 & 6.04 & 0.70 & 2.42 & 1.23 & 0.81 & 0.48 \\
Qwen3-VL-2B & 0.98 & 0.72 & 2.88 & 8.65 & 0.58 & 3.62 & 1.29 & 0.71 & 2.89 \\
Qwen3-VL-2B-T & 5.63 & 4.32 & 18.94 & 34.78 & 4.25 & 10.37 & 5.55 & 9.86 & 13.88 \\
Qwen3-VL-4B & 0.89 & 0.74 & 3.55 & 9.32 & 0.68 & 4.89 & 1.62 & 3.65 & 1.83 \\
Qwen3-VL-4B-T & 7.71 & 8.75 & 40.79 & 57.55 & 6.81 & 14.35 & 8.49 & 16.86 & 13.62 \\
Qwen3-VL-8B & 1.39 & 0.76 & 4.82 & 12.74 & 1.00 & 6.89 & 3.28 & 4.49 & 3.81 \\
Qwen3-VL-8B-T & 7.91 & 6.41 & 27.30 & 57.59 & 7.90 & 17.53 & 8.98 & 19.43 & 14.29 \\
R-4B & 2.04 & 4.70 & 8.07 & 32.70 & 4.16 & 5.24 & 2.18 & 4.67 & 3.47 \\
gemma-3-4b-it & 0.34 & 0.49 & 2.42 & 3.94 & 0.59 & 2.75 & 0.64 & 0.36 & 0.27 \\
Phi-3.5-vision-instruct & 0.17 & 0.46 & 2.19 & 3.60 & 0.61 & 0.50 & 0.79 & 0.31 & 0.22 \\
GLM-4.1V-9B & 1.22 & 0.40 & 4.08 & 11.41 & 0.58 & 4.79 & 1.34 & 0.98 & 0.57 \\
GLM-4.1V-9B-T & 4.78 & 9.13 & 18.36 & 48.30 & 4.10 & 9.61 & 7.51 & 12.27 & 8.22 \\
\bottomrule
\end{tabular}
\caption{H100 hours per model and capability. Sample counts are shown in the second row (italic). Values represent total H100 hours required to process all samples for each capability. Missing entries (---) indicate the capability was not evaluated for that model.}
\label{tab:h100_hours}
\end{table*}

\clearpage
\section{Considerations converting from MCQ to Generative}\label{app:step1}
\subsection{Qualitative Example}
\begin{figure}[h]
    \centering
    \includegraphics[width=0.9\linewidth]{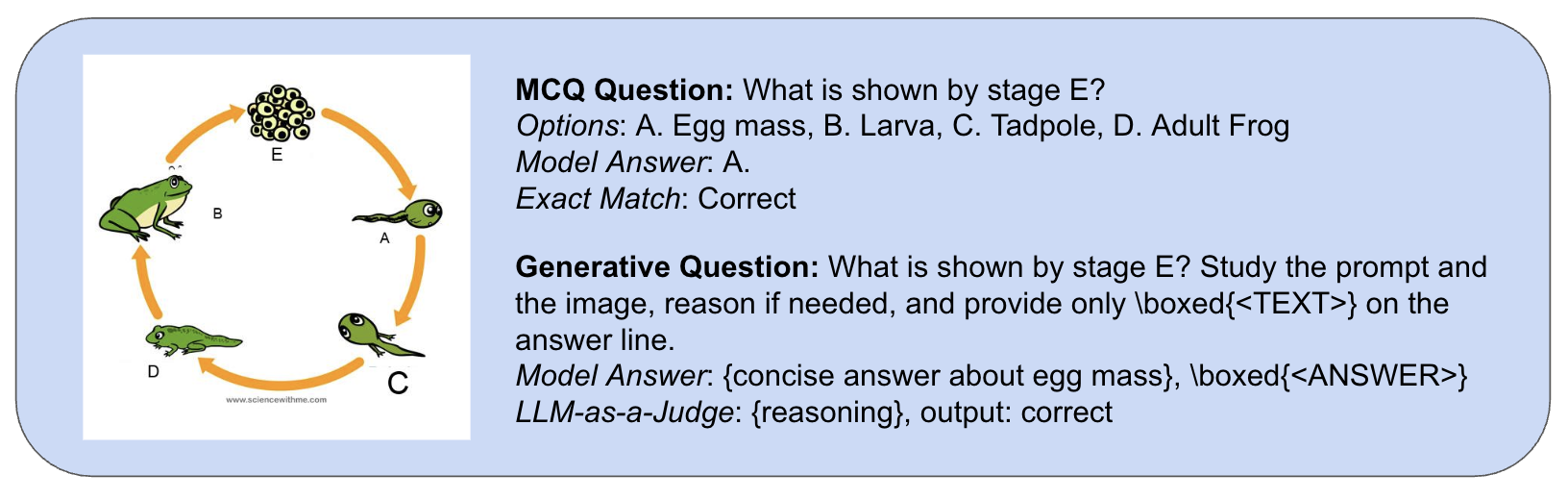}
    \caption{Example of converting a sample from AI2D from its native MCQ format to the generative setting}
    \label{fig:ai2dmcqgen}
\end{figure}

We provide an example of converting an eval sample from AI2D from MCQ to generative in Fig. \ref{fig:ai2dmcqgen}

\subsection{Evals that we had to keep as MCQ}
\begin{figure}[h]
    \centering
    \begin{subfigure}[b]{0.5\textwidth}
        \centering
        \includegraphics[width=\linewidth]{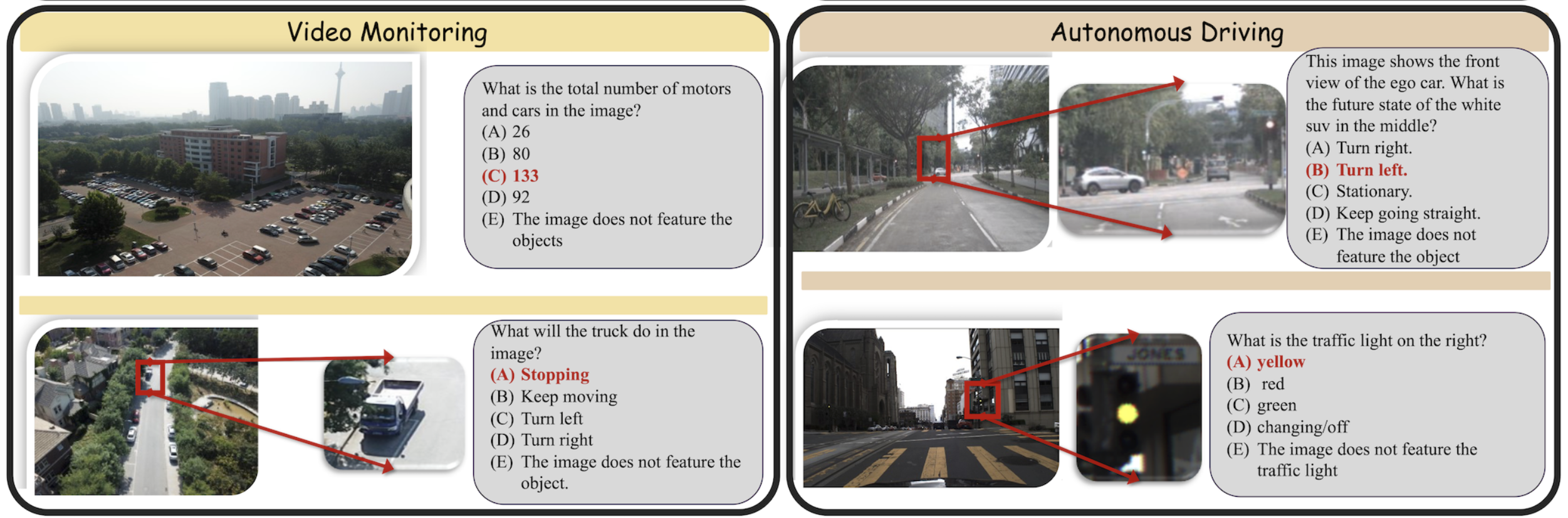} 
        \caption{Samples from MME-Realworld}
    \end{subfigure}
    \hfill
    \begin{subfigure}[b]{0.44\textwidth}
        \centering
        \includegraphics[width=\linewidth]{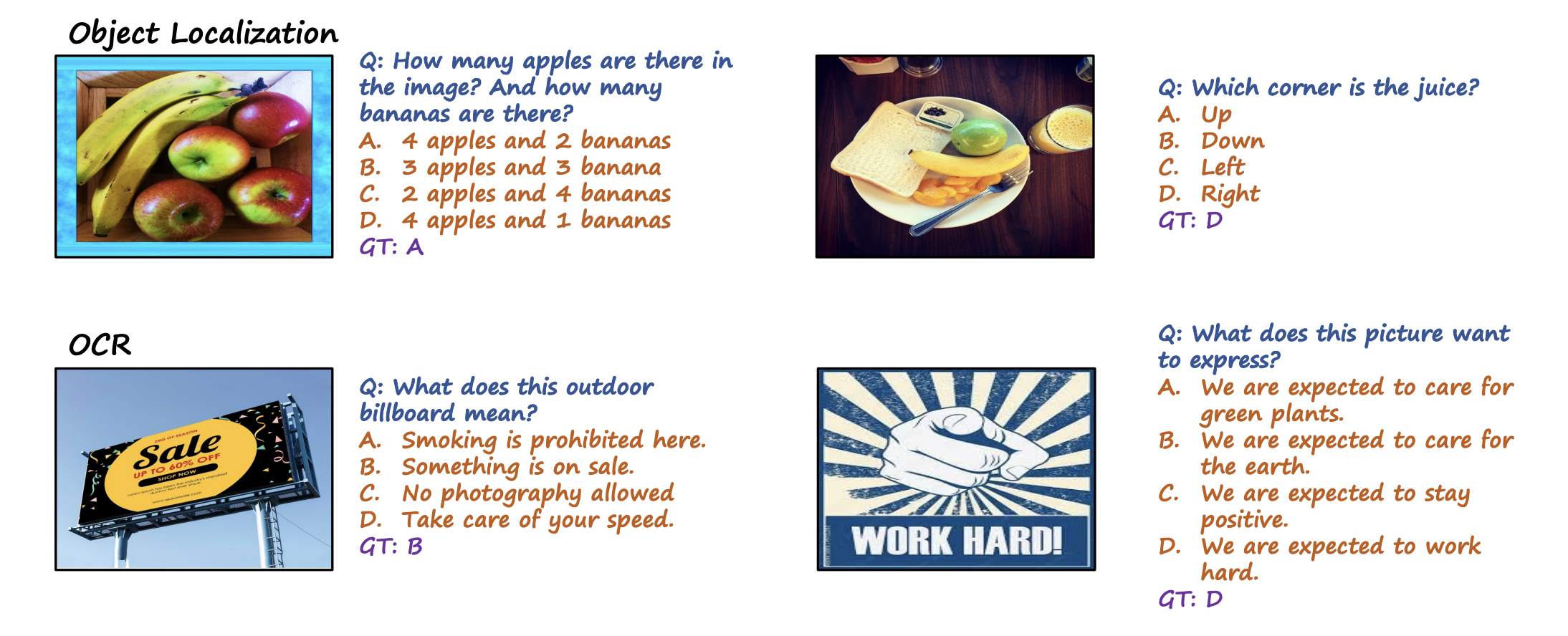} 
        \caption{Samples from MMBench}
    \end{subfigure}
    \caption{Dependence on options for solving various tasks in MME-Realworld and MMbench}
    \label{fig:dataset_samples}
\end{figure}

\begin{itemize}
    \item LogicVista - questions are mensa style puzzles where options are the next image in the sequence so cannot be generated de-novo
    \item MME-Realworld and MMBench - We keep these in their original MCQ format because many questions are underspecified as free-form prompts: the answer choices provide crucial context about what kind of response is expected. Converting these items to generative QA would not give models sufficient signal about the task definition (see Figure~\ref{fig:dataset_samples}). Further, many counting questions involve scenes with a very large number of objects, where an exact count is both ambiguous and brittle to minor visual uncertainty. In these cases, the benchmark is primarily probing whether the model can distinguish coarse scales (e.g., few vs many, tens vs hundreds) rather than recover a precise integer, so enforcing an exact-match generative answer would add noise and misrepresent the intended capability being measured.
\end{itemize}

\clearpage
\section{VLM-as-Judge Filtering Results}\label{app:step3}

Table~\ref{tab:filtering_stats} reports filtering statistics for examples that all evaluated models answered incorrectly. Examples flagged by the VLM judge as ambiguous, incorrectly labeled, or requiring higher resolution are removed from the benchmark. Retained examples represent ``frontier'' cases where current models uniformly fail on valid, high-quality data, indicating the benchmark has not yet saturated.

\begin{table}[h]
\centering
\caption{Two-Stage Quality Filtering Statistics}
\label{tab:filtering_stats}
\resizebox{\textwidth}{!}{%
\begin{tabular}{lcccccc}
\toprule
\multirow{2}{*}{Dataset} & \multirow{2}{*}{Total Samples} & \multirow{2}{*}{Flagged} & \multicolumn{4}{c}{Removed by Judge} \\
\cmidrule(lr){4-7}
& & & Ambiguous & Incorrect & Low Resolution & Total (\%) \\
\midrule
AI2D & 3,088 & 342 & 111 & 72 & 8 & 131 (4.2\%) \\
CC-OCR (kie) & 2,008 & 1 & 0 & 0 & 0 & 0 (0.0\%) \\
CC-OCR (Multi Scene OCR) & 2,750 & 4 & 4 & 2 & 4 & 4 (0.1\%) \\
ChartQA & 2,500 & 57 & 19 & 25 & 1 & 36 (1.4\%) \\
ChartQAPro & 1,948 & 693 & 234 & 190 & 56 & 335 (17.2\%) \\
CharXiv (DQ) & 4,000 & 37 & 2 & 15 & 3 & 17 (0.4\%) \\
CharXiv (RQ) & 1,000 & 69 & 10 & 17 & 4 & 23 (2.3\%) \\
CountBench & 491 & 4 & 3 & 4 & 2 & 4 (0.8\%) \\
DocVQA & 5,349 & 6 & 1 & 1 & 1 & 2 (0.0\%) \\
InfoVQA & 2,801 & 37 & 7 & 4 & 6 & 12 (0.4\%) \\
LogicVista & 448 & 8 & 2 & 1 & 0 & 2 (0.4\%) \\
MathVerse (reasoning) & 3,940 & 608 & 141 & 185 & 50 & 234 (5.9\%) \\
MathVerse (wo) & 3,940 & 580 & 130 & 169 & 36 & 221 (5.6\%) \\
MathVision & 3,040 & 512 & 133 & 153 & 65 & 220 (7.2\%) \\
MathVista & 1,000 & 59 & 41 & 38 & 4 & 48 (4.8\%) \\
MME-RealWorld (Autonomous Driving) & 5,004 & 2560 & 2076 & 1723 & 759 & 2198 (43.9\%) \\
MME-RealWorld (Diagram / Table) & 5,933 & 577 & 77 & 161 & 157 & 307 (5.2\%) \\
MME-RealWorld (Video Monitoring) & 2,694 & 1555 & 1197 & 898 & 1015 & 1342 (49.8\%) \\
MME-RealWorld (OCR-in-the-Wild) & 6,240 & 478 & 141 & 229 & 101 & 311 (5.0\%) \\
MMBench & 4,329 & 152 & 82 & 75 & 0 & 107 (2.5\%) \\
MMMU-Pro & 1,730 & 855 & 331 & 256 & 74 & 420 (24.3\%) \\
OCR-VQA & 100,424 & 6068 & 3506 & 5245 & 547 & 5531 (5.5\%) \\
OCRBench\_$v2$ & 10,000 & 1360 & 340 & 302 & 146 & 533 (5.3\%) \\
Pixmo-Pointing & 394 & 80 & 29 & 24 & 0 & 36 (9.1\%) \\
RealWorldQA & 764 & 38 & 12 & 18 & 1 & 20 (2.6\%) \\
Ref-COCO-M & 5,598 & 35 & 19 & 8 & 0 & 24 (0.4\%) \\
RefCOCO+ (testA) & 5,726 & 65 & 34 & 21 & 0 & 45 (0.8\%) \\
RefCOCO+ (testB) & 4,889 & 120 & 60 & 33 & 1 & 76 (1.6\%) \\
RefCOCO (testA) & 5,657 & 45 & 24 & 16 & 0 & 32 (0.6\%) \\
RefCOCO (testB) & 5,095 & 77 & 38 & 27 & 0 & 52 (1.0\%) \\
RefCOCO-G & 9,602 & 137 & 38 & 42 & 1 & 63 (0.7\%) \\
TallyQA & 38,589 & 798 & 230 & 416 & 103 & 487 (1.3\%) \\
TextVQA & 5,000 & 94 & 40 & 31 & 20 & 56 (1.1\%) \\
VQA-V2 & 214,354 & 2297 & 1433 & 447 & 373 & 1585 (0.7\%) \\
\bottomrule
\end{tabular}%
}
\end{table}

\clearpage

\begin{tcolorbox}[
    title=VLM-as-Judge Verification Prompt,
    colback=white,
    colframe=black,
    boxrule=0.5pt,
    arc=2pt,
    left=6pt,
    right=6pt,
    top=6pt,
    bottom=6pt,
    breakable
]
\small
\begin{verbatim}
You are an expert Quality Assurance verifier for a Vision-Language Benchmark.

Task: You will be shown an Image, a Question, and the dataset's Ground Truth Answer.
Your job is to verify if the Ground Truth Answer is strictly correct and unambiguous
based on the image.

Context: None of the current state-of-the-art models were able to answer this
question correctly. We need to know if this is because the task is hard (valid
Frontier Example) or because the Ground Truth is flawed (Invalid).

Criteria for marking as 'Invalid' (ground_truth_wrong=true):
1. The image is missing, corrupted, or unreadable.
2. The question refers to details not present in the image.
3. The Ground Truth Answer is factually incorrect based on the image.
4. The question is ambiguous and has multiple valid answers, but the Ground Truth
   only accepts one specific phrasing.

Output Format: You must return ONLY a JSON object with three boolean fields and
one concise rationale field:
{
  "needs_high_resolution": true|false,
  "ground_truth_wrong": true|false,
  "question_is_ambiguous": true|false,
  "reason": "<one short sentence explaining your decision>"
}

Conservative Strategy: When in doubt about whether the ground truth is correct,
prefer marking it as invalid (ground_truth_wrong=true). We prefer False Positives
(discarding a valid hard example) over False Negatives (keeping a bad example).

Is the Ground Truth valid?
\end{verbatim}
\end{tcolorbox}

\clearpage

\begin{table}[h]
\centering
\caption{Quality Filtering Statistics Aggregated by Capability}
\label{tab:capability_filtering_summary}
\begin{tabular}{lrrr}
\toprule
\textbf{Capability} & \textbf{Total Samples} & \textbf{Samples Removed} & \textbf{Discarded (\%)} \\
\midrule
Spatial             & 8,462                  & 3,560                    & 42.07\% \\
Math / Logic        & 12,368                 & 725                      & 5.86\%  \\
Document            & 107,781                & 5,533                    & 5.13\%  \\
Table            & 9,021                  & 438                      & 4.86\%  \\
Chart               & 12,249                 & 423                      & 3.45\%  \\
Scene OCR           & 13,990                 & 371                      & 2.65\%  \\
Counting            & 39,080                 & 491                      & 1.26\%  \\
General             & 220,413                & 2,112                    & 0.96\%  \\
Grounding           & 36,961                 & 328                      & 0.89\%  \\
\midrule
\textbf{Total}      & \textbf{460,325}       & \textbf{13,981}          & \textbf{3.04\%} \\
\bottomrule
\end{tabular}
\end{table}

\clearpage
\section{Item-Discrimination Subset Selection}\label{app:step4}

\begin{figure}[h]
    \centering
    \includegraphics[width=0.8\linewidth]{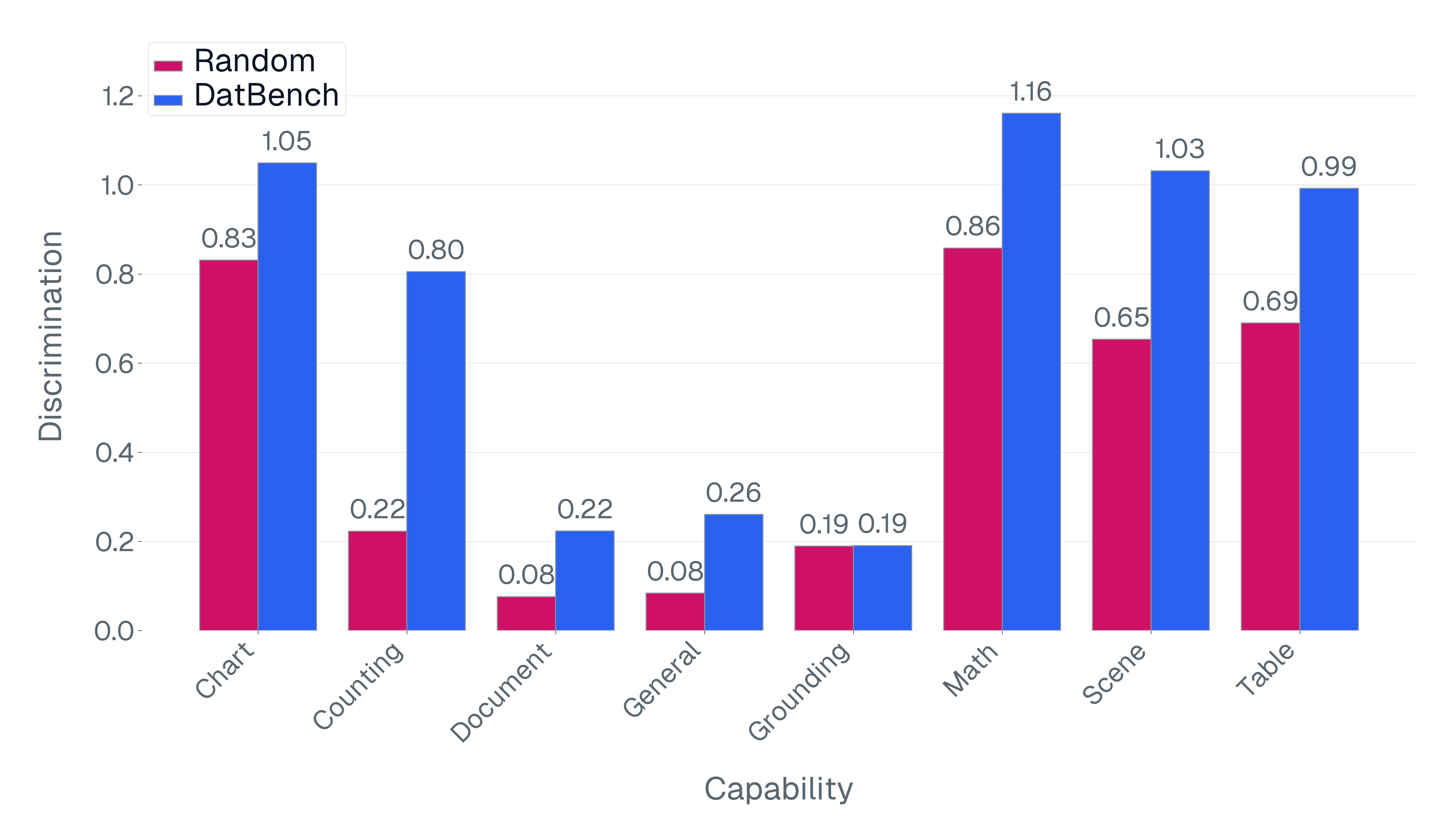}
    \caption{Discriminative power of DatBench compared with Random subsets ($k$=5000)}
    \label{fig:discrimination_by_capability_k5000}
\end{figure}

\begin{figure}[t]
    \centering

    \begin{subfigure}{0.332\textwidth}
        \centering
        \includegraphics[width=\linewidth]{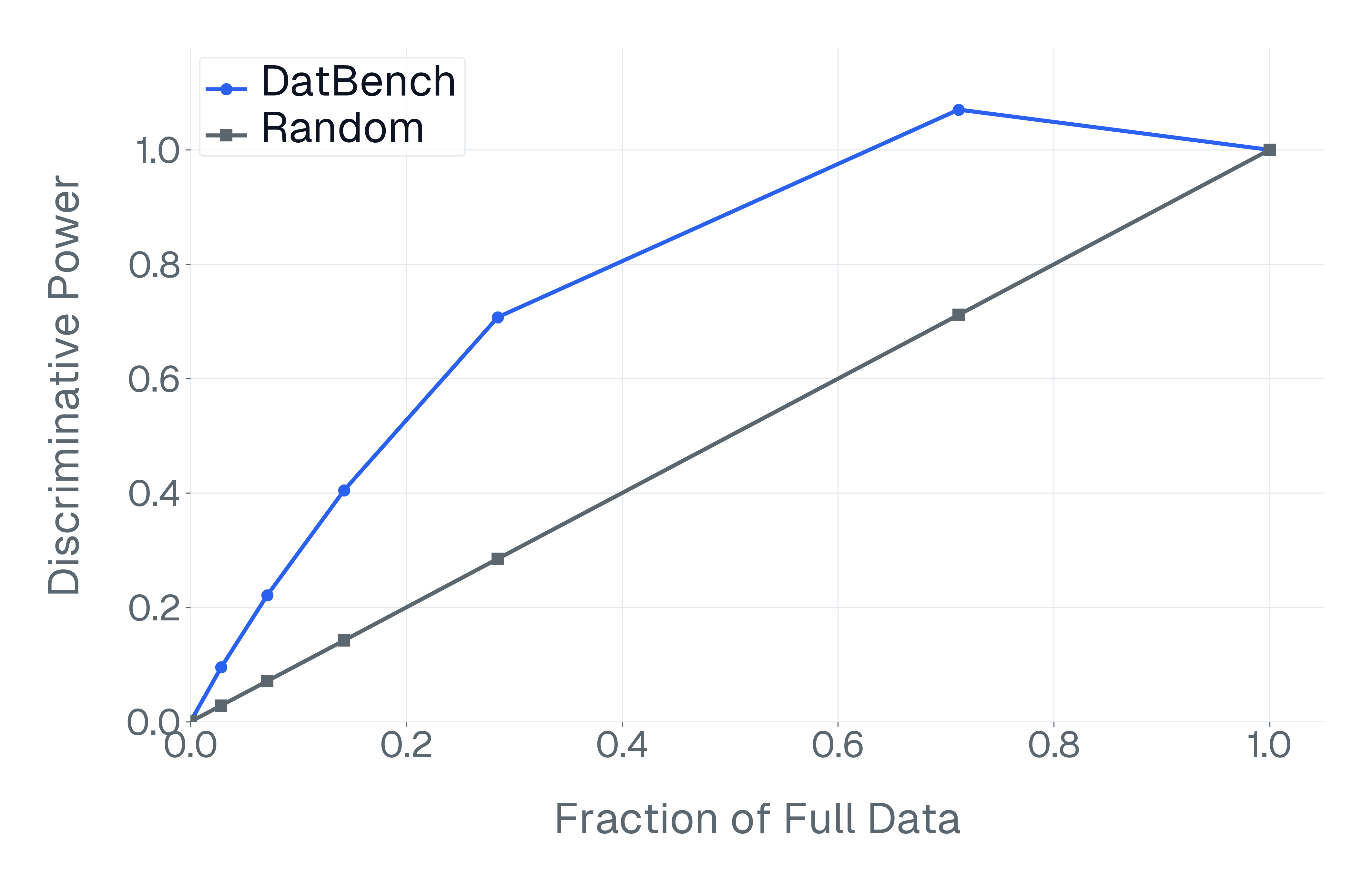}
        \caption{Spatial}
    \end{subfigure}\hfill
    \begin{subfigure}{0.332\textwidth}
        \centering
        \includegraphics[width=\linewidth]{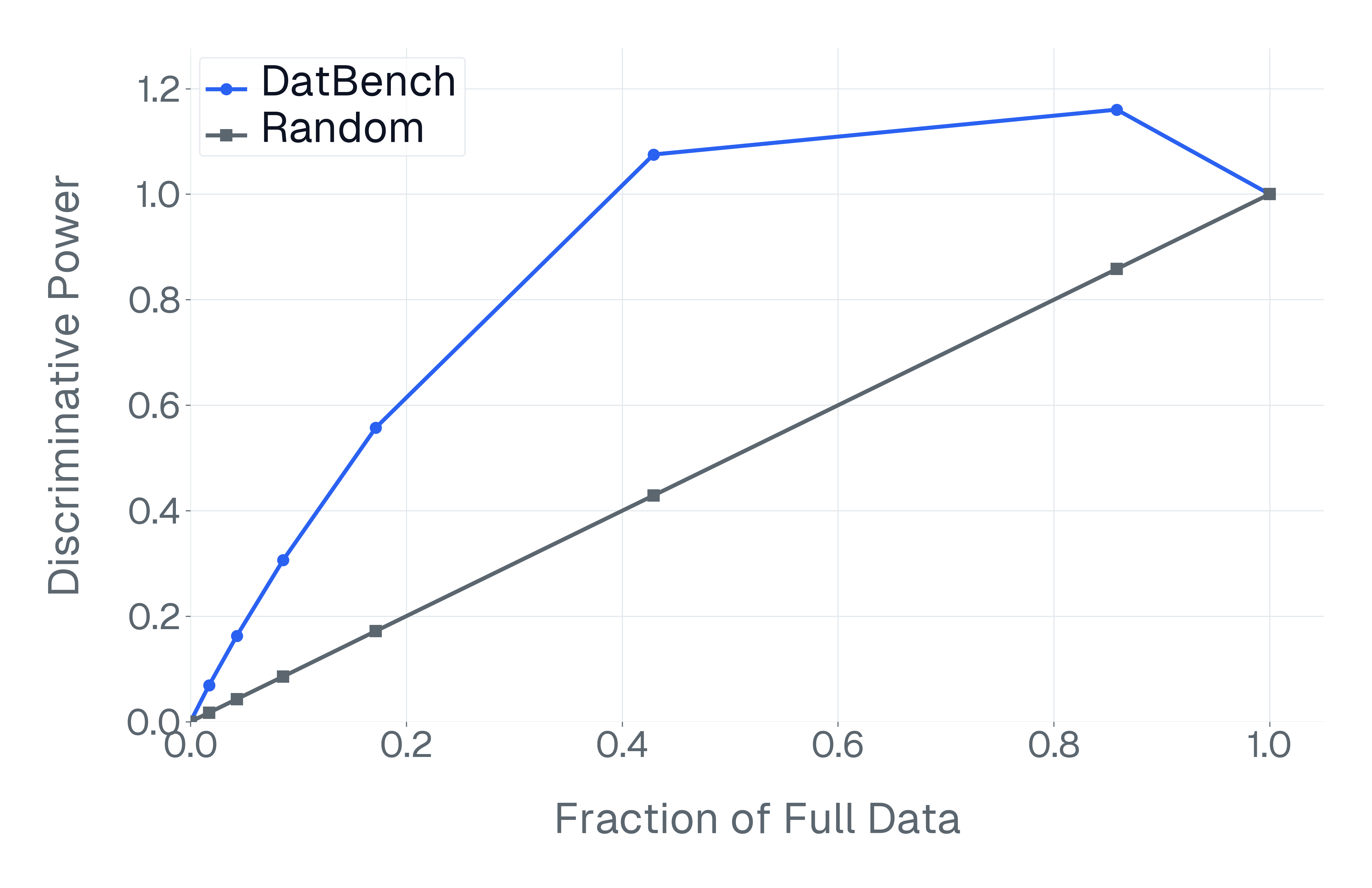}
        \caption{Math}
    \end{subfigure}\hfill
    \begin{subfigure}{0.332\textwidth}
        \centering
        \includegraphics[width=\linewidth]{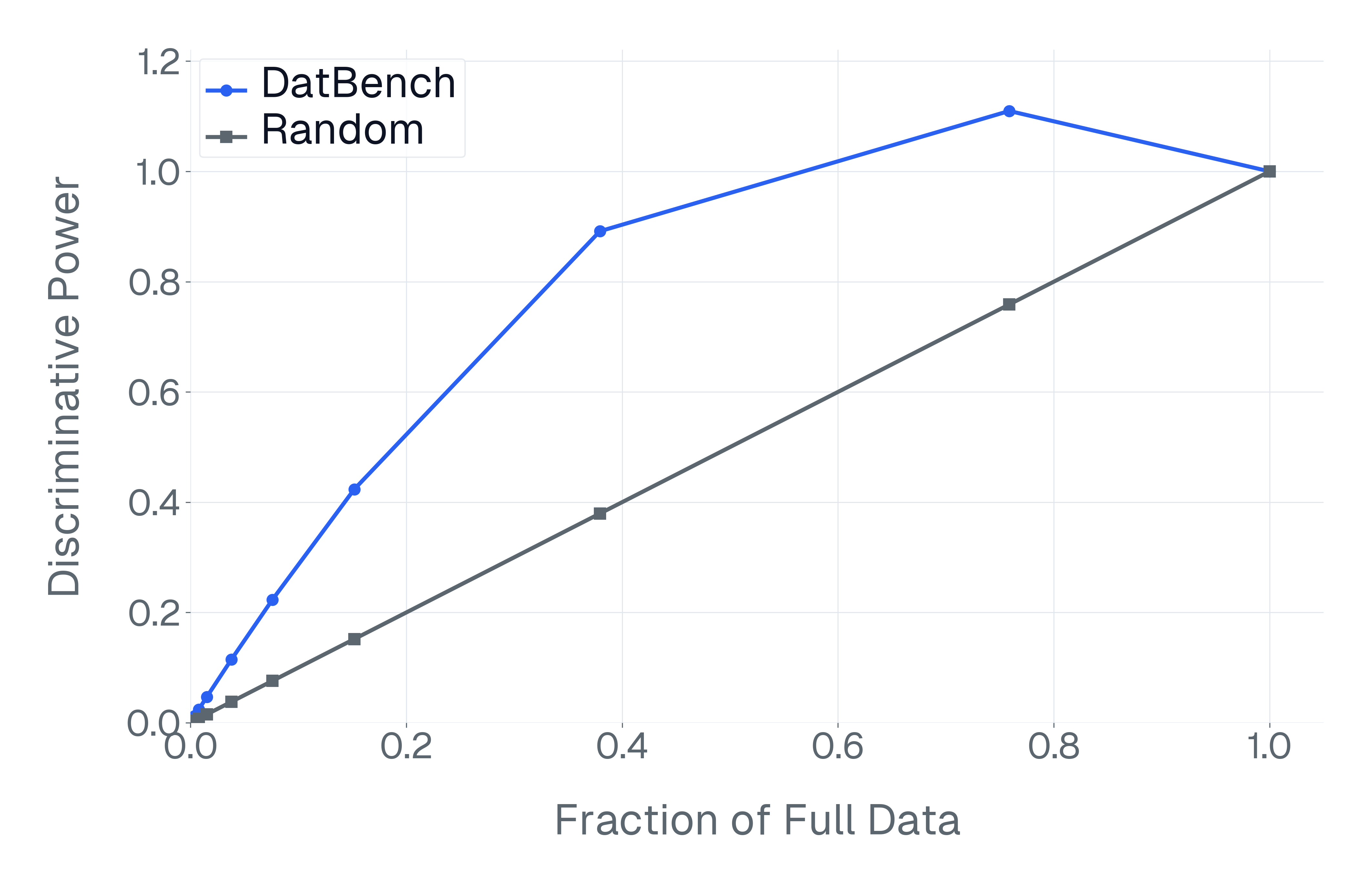}
        \caption{Document}
    \end{subfigure}

    \vspace{0.4em}

    \begin{subfigure}{0.332\textwidth}
        \centering
        \includegraphics[width=\linewidth]{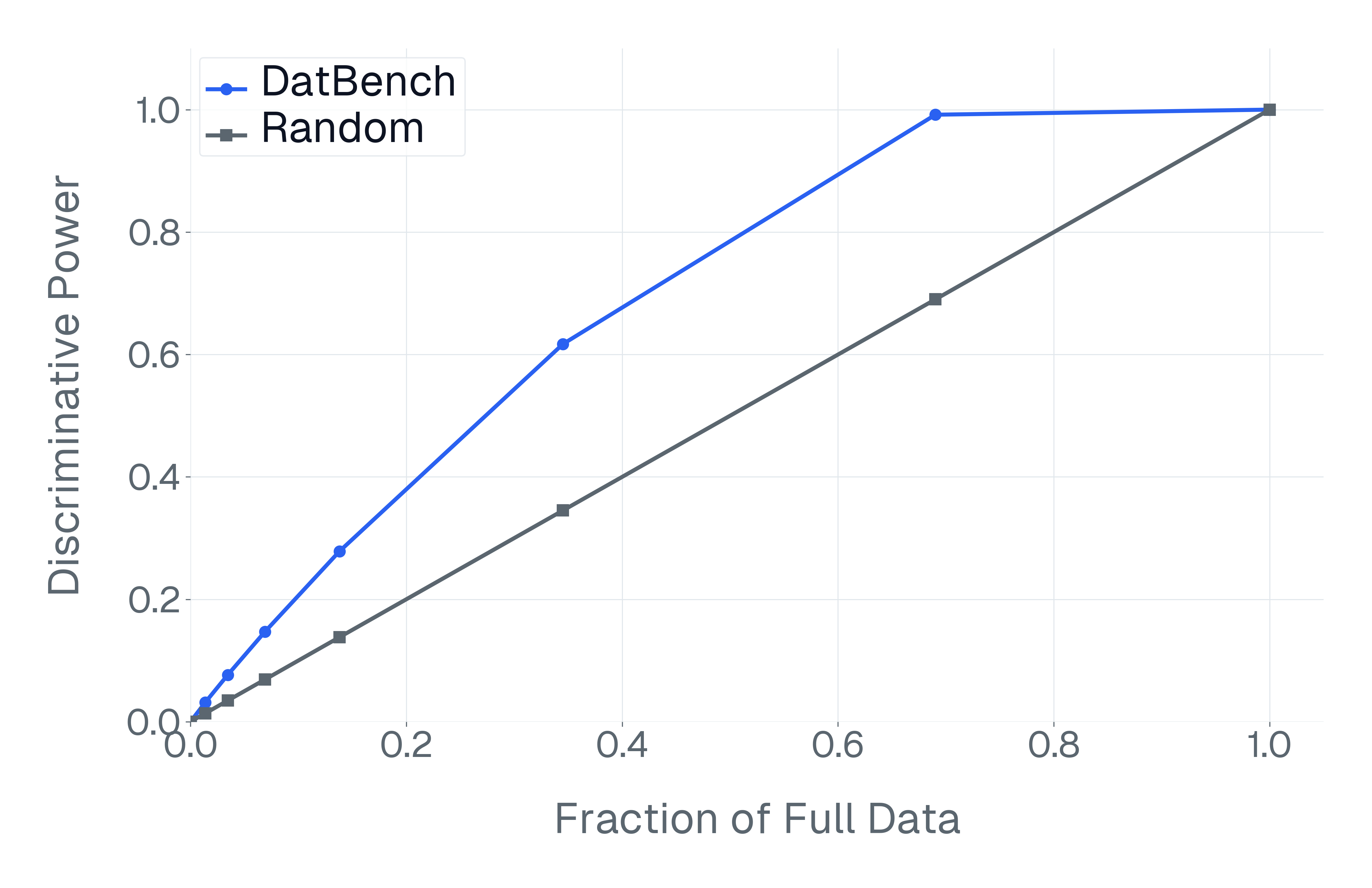}
        \caption{Table}
    \end{subfigure}\hfill
    \begin{subfigure}{0.332\textwidth}
        \centering
        \includegraphics[width=\linewidth]{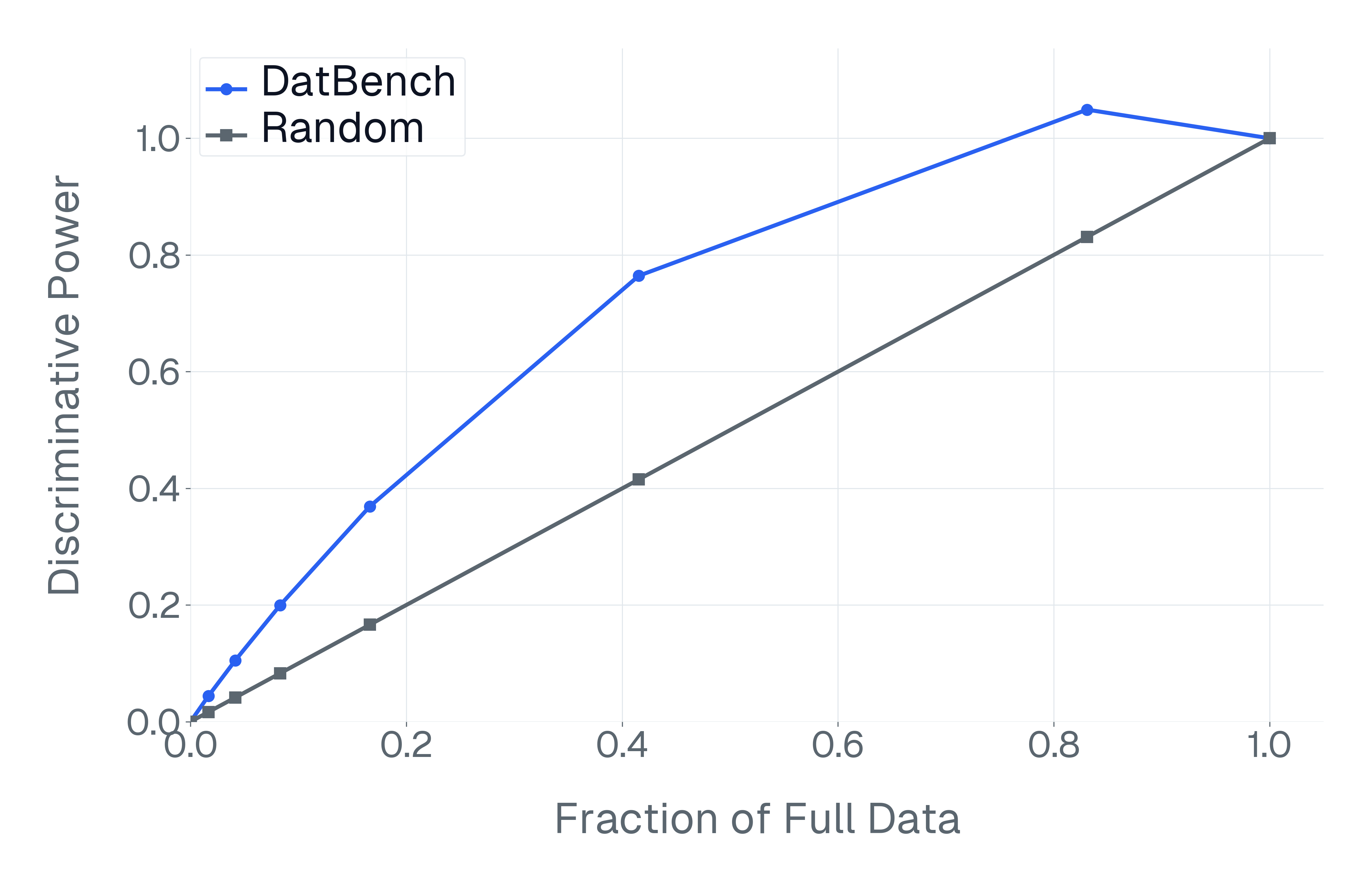}
        \caption{Chart}
    \end{subfigure}\hfill
    \begin{subfigure}{0.332\textwidth}
        \centering
        \includegraphics[width=\linewidth]{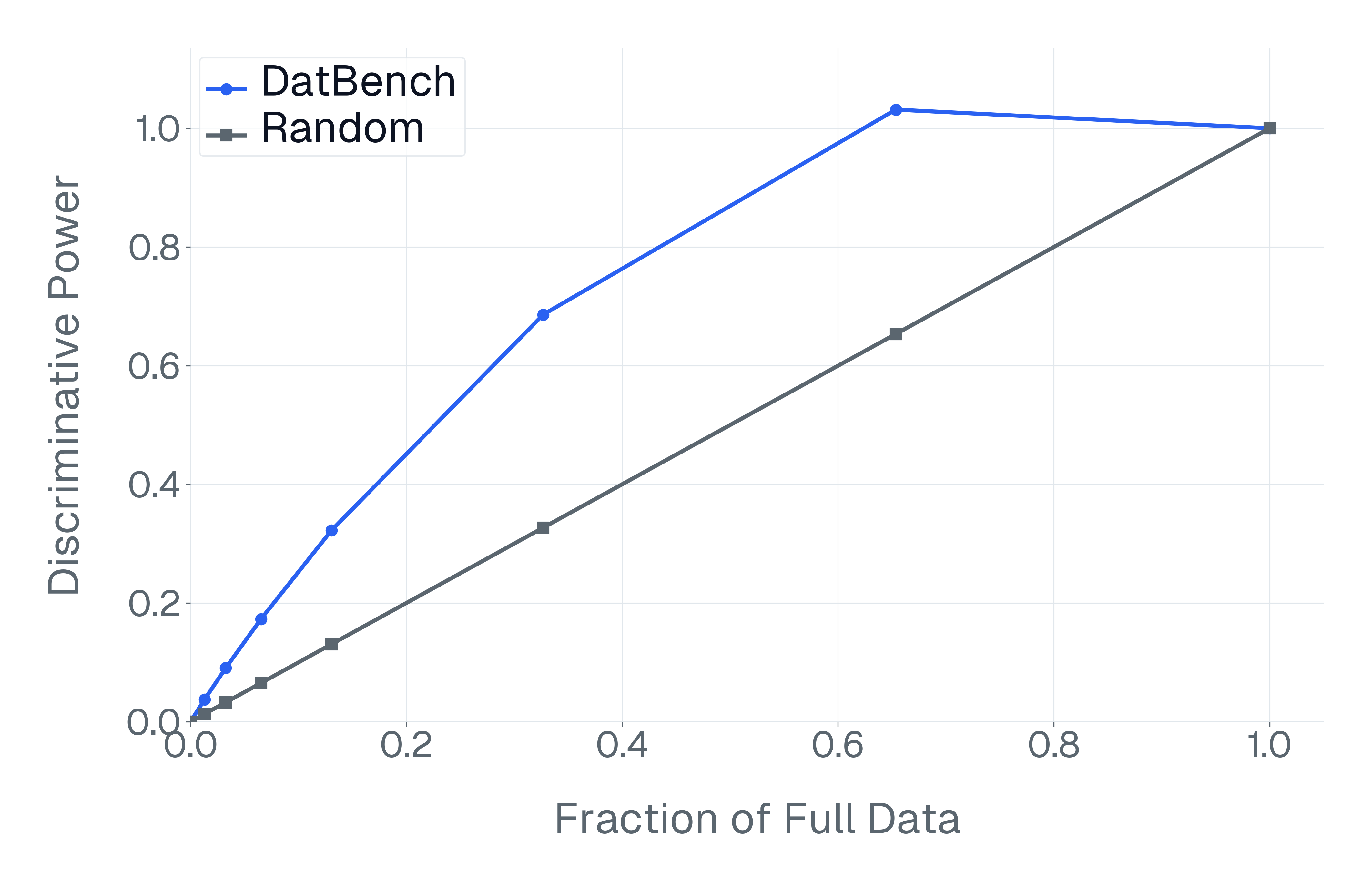}
        \caption{Scene}
    \end{subfigure}

    \vspace{0.4em}

    \begin{subfigure}{0.332\textwidth}
        \centering
        \includegraphics[width=\linewidth]{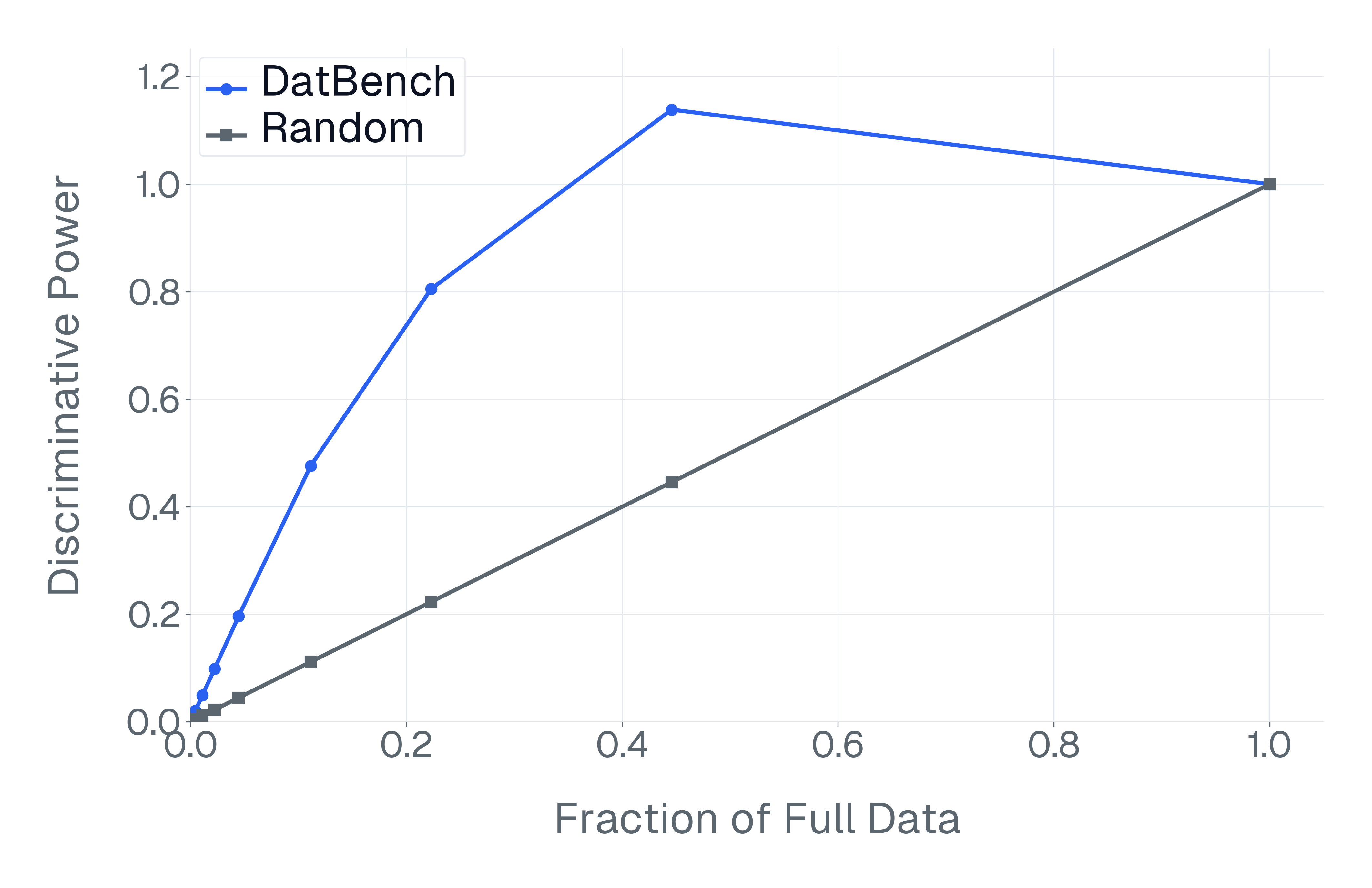}
        \caption{Counting}
    \end{subfigure}\hfill
    \begin{subfigure}{0.332\textwidth}
        \centering
        \includegraphics[width=\linewidth]{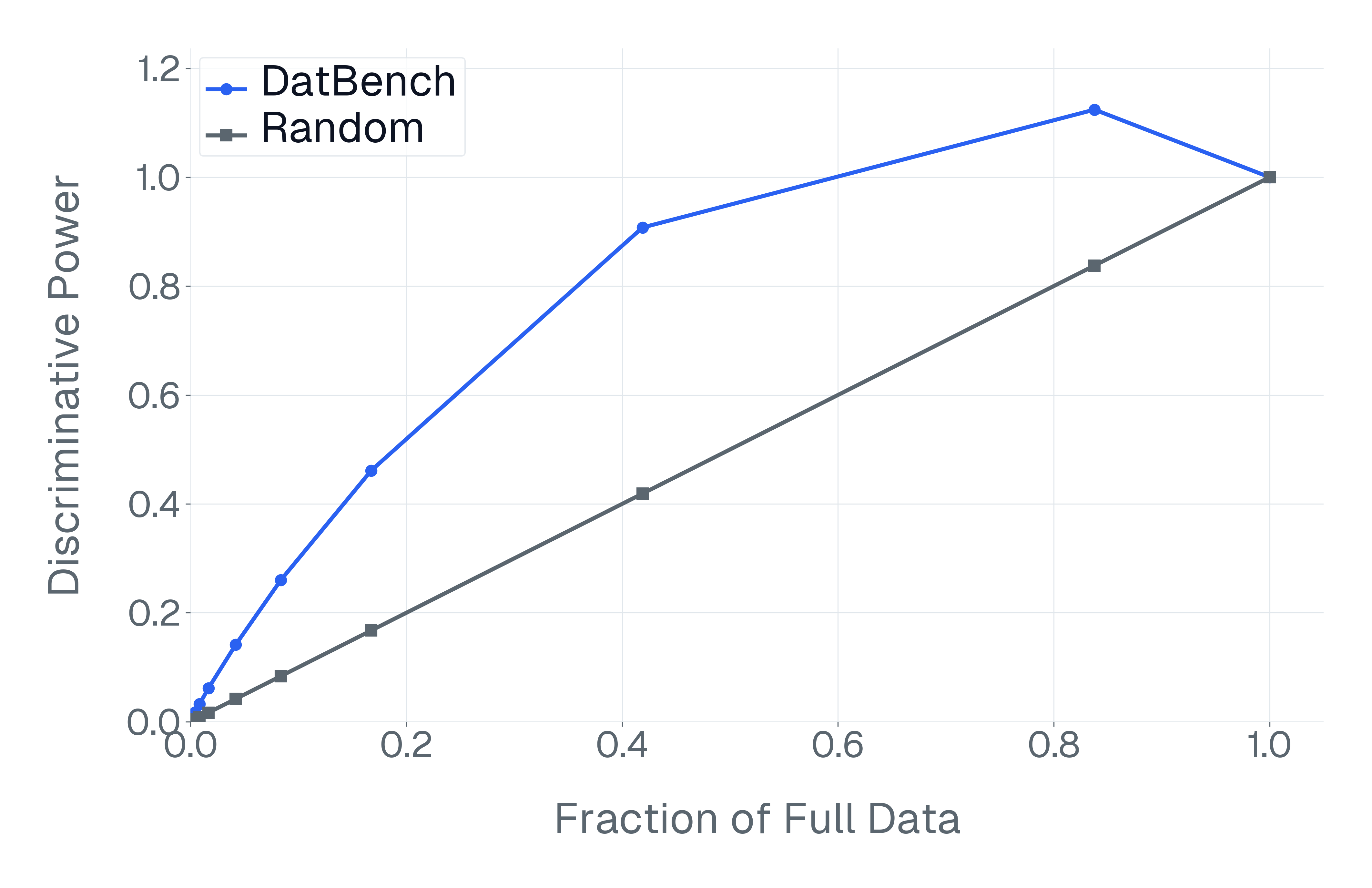}
        \caption{General}
    \end{subfigure}\hfill
    \begin{subfigure}{0.332\textwidth}
        \centering
        \includegraphics[width=\linewidth]{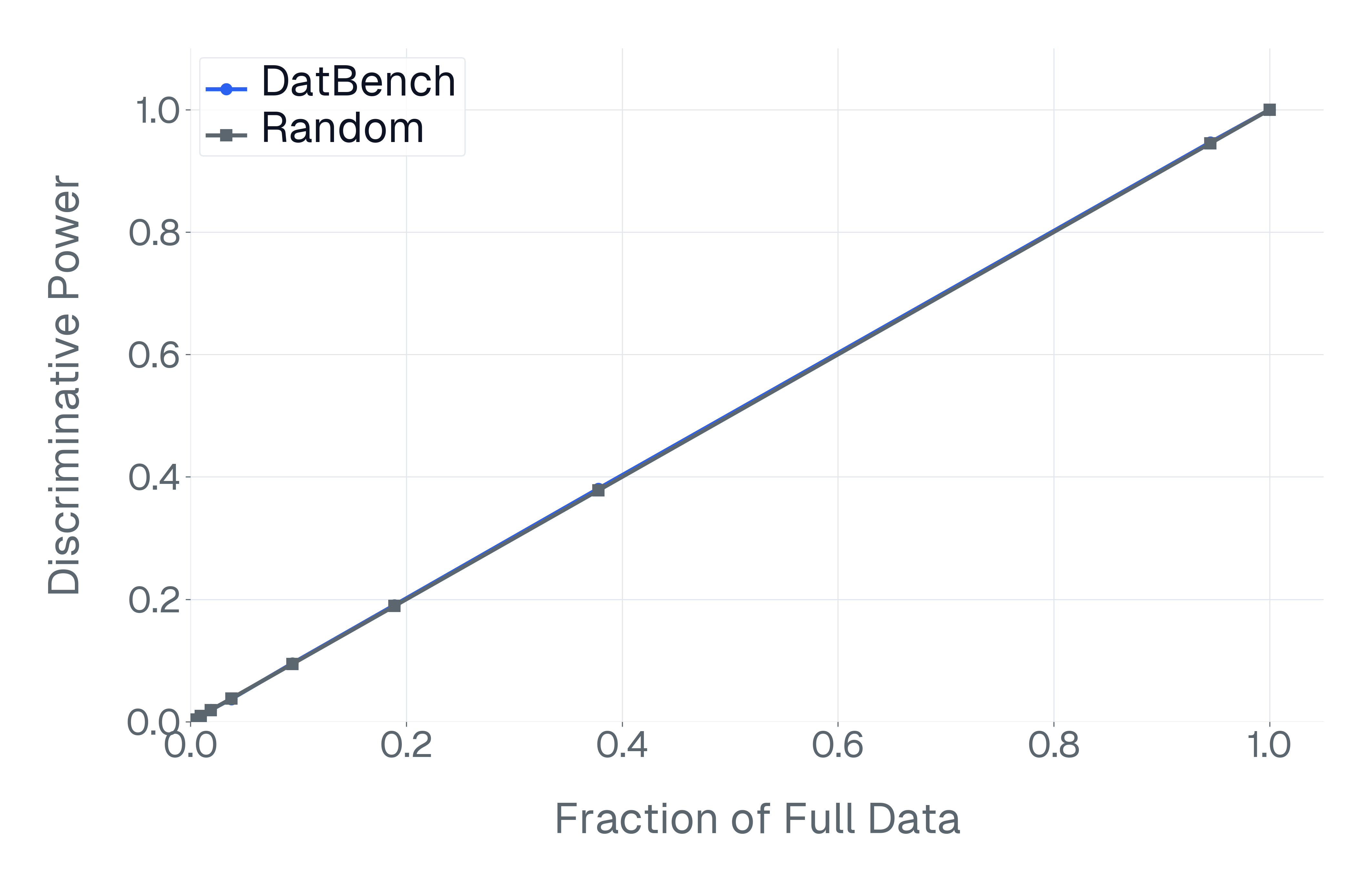}
        \caption{Grounding}
    \end{subfigure}

    \caption{Discriminative power as a function of retained data across all capabilities
    }
    \label{fig:discriminative_power_all_cap}
\end{figure}

\begin{figure}[p]
    \centering

    \begin{subfigure}{0.332\textwidth}
        \centering
        \includegraphics[width=\linewidth]{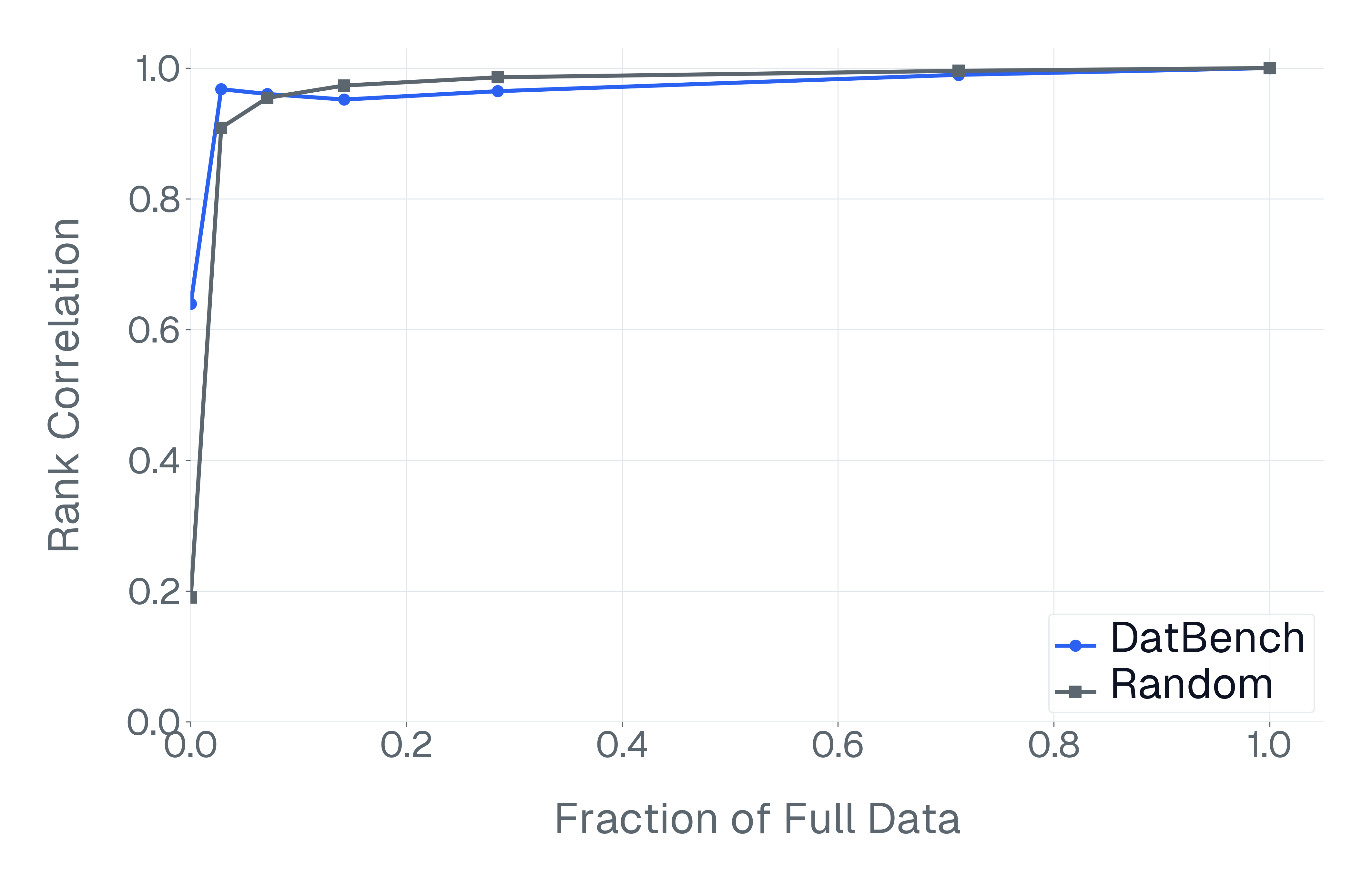}
        \caption{Spatial}
    \end{subfigure}\hfill
    \begin{subfigure}{0.332\textwidth}
        \centering
        \includegraphics[width=\linewidth]{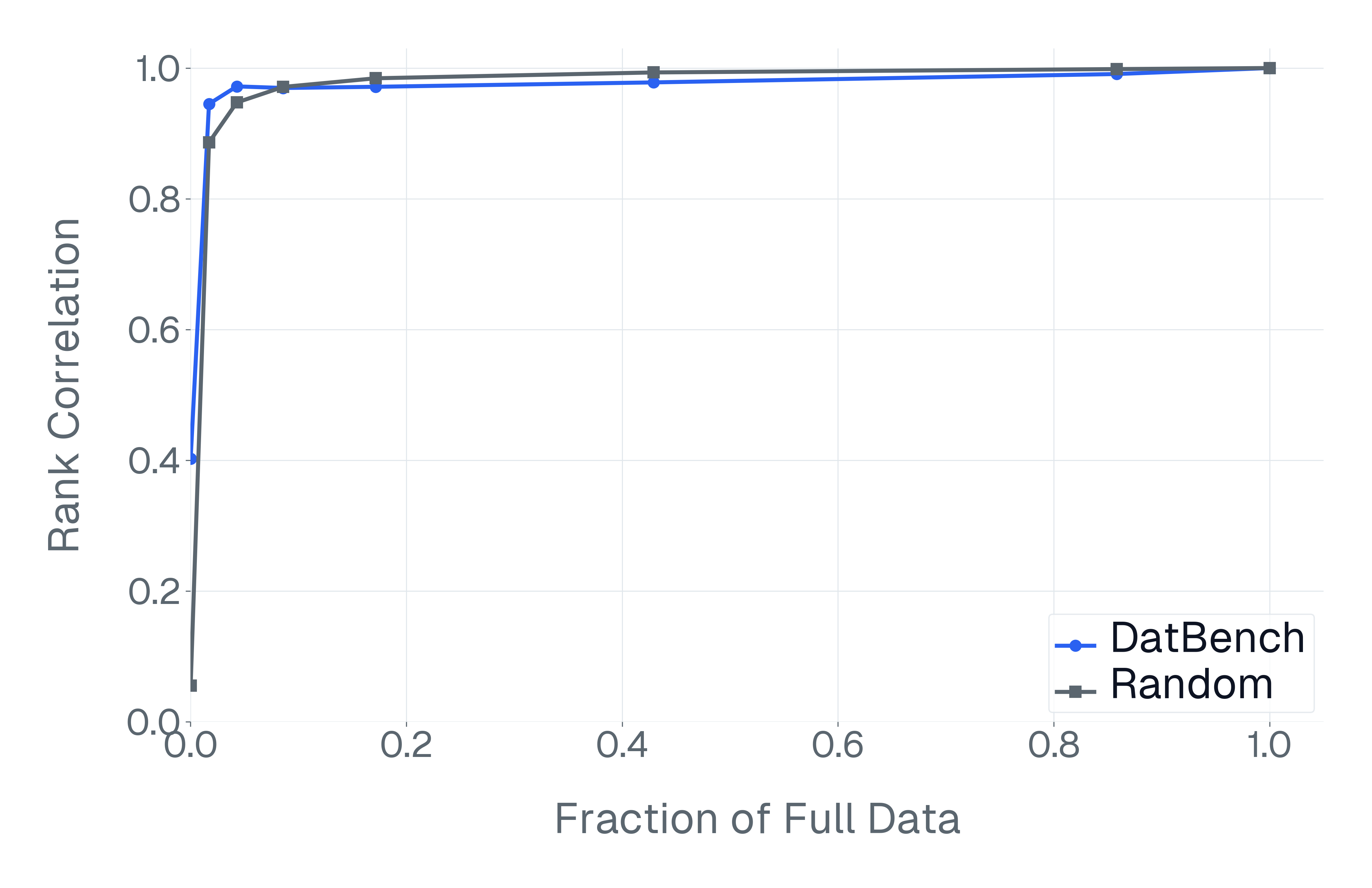}
        \caption{Math}
    \end{subfigure}\hfill
    \begin{subfigure}{0.332\textwidth}
        \centering
        \includegraphics[width=\linewidth]{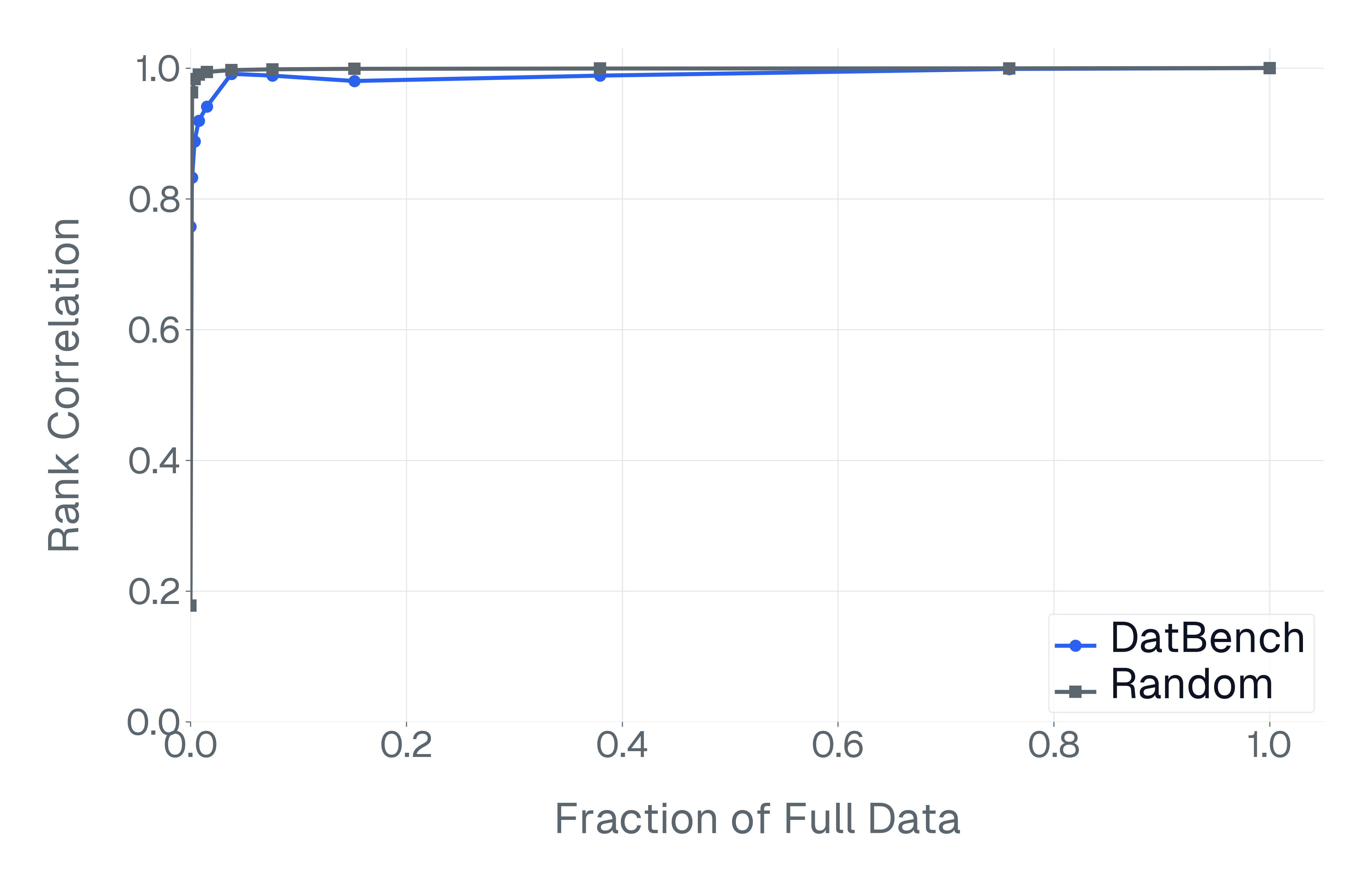}
        \caption{Document}
    \end{subfigure}

    \vspace{0.4em}

    \begin{subfigure}{0.332\textwidth}
        \centering
        \includegraphics[width=\linewidth]{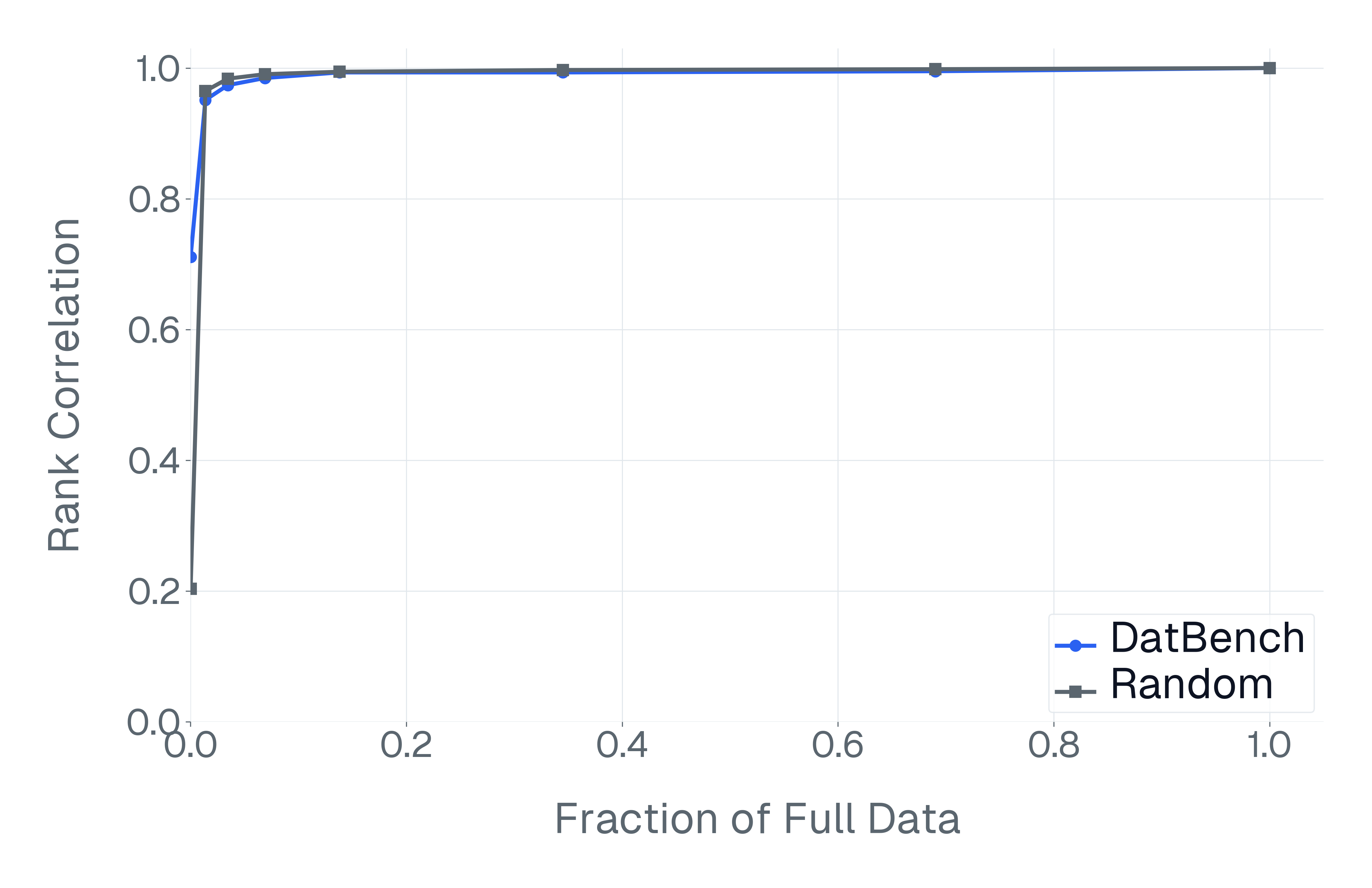}
        \caption{Table}
    \end{subfigure}\hfill
    \begin{subfigure}{0.332\textwidth}
        \centering
        \includegraphics[width=\linewidth]{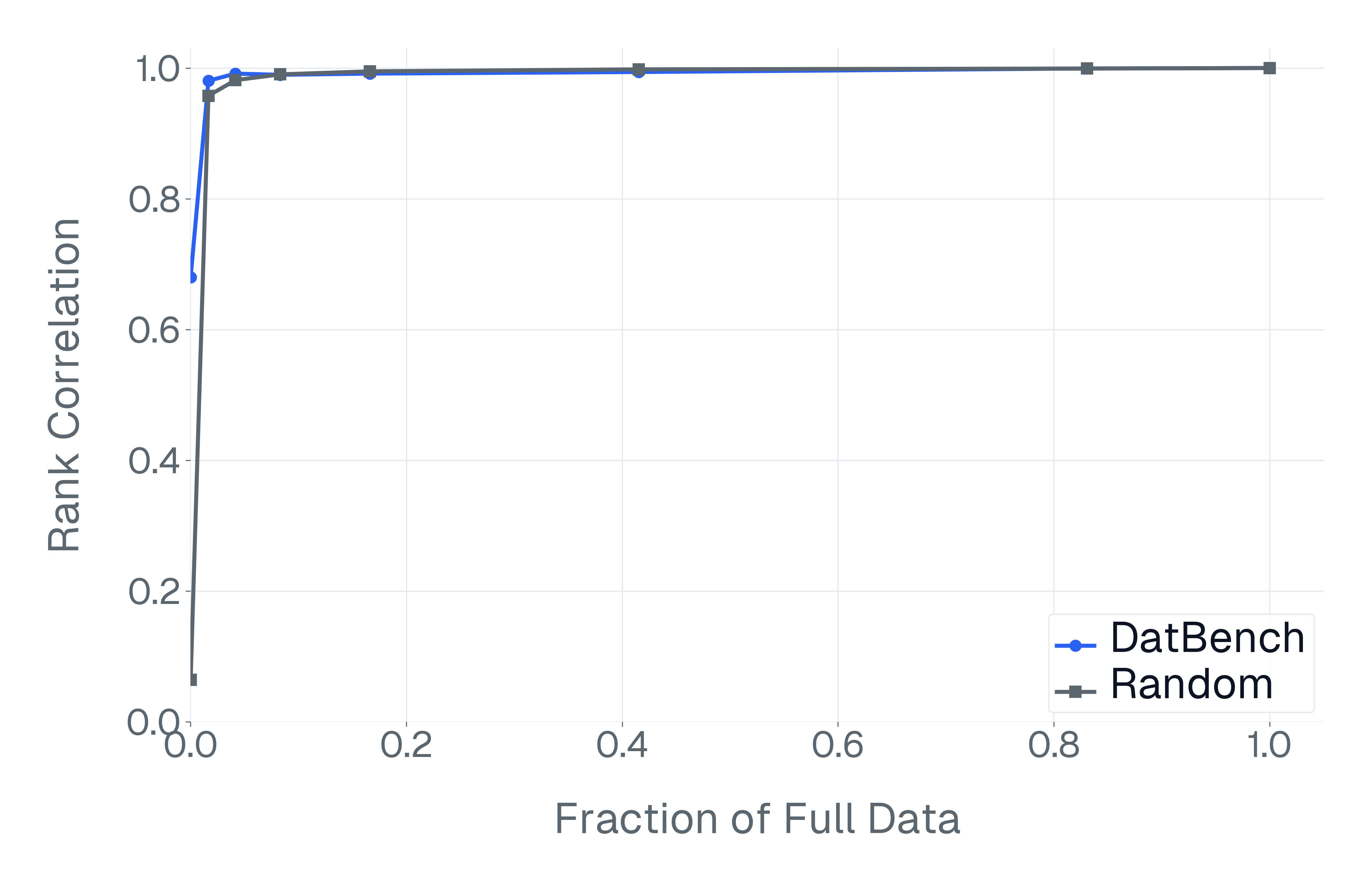}
        \caption{Chart}
    \end{subfigure}\hfill
    \begin{subfigure}{0.332\textwidth}
        \centering
        \includegraphics[width=\linewidth]{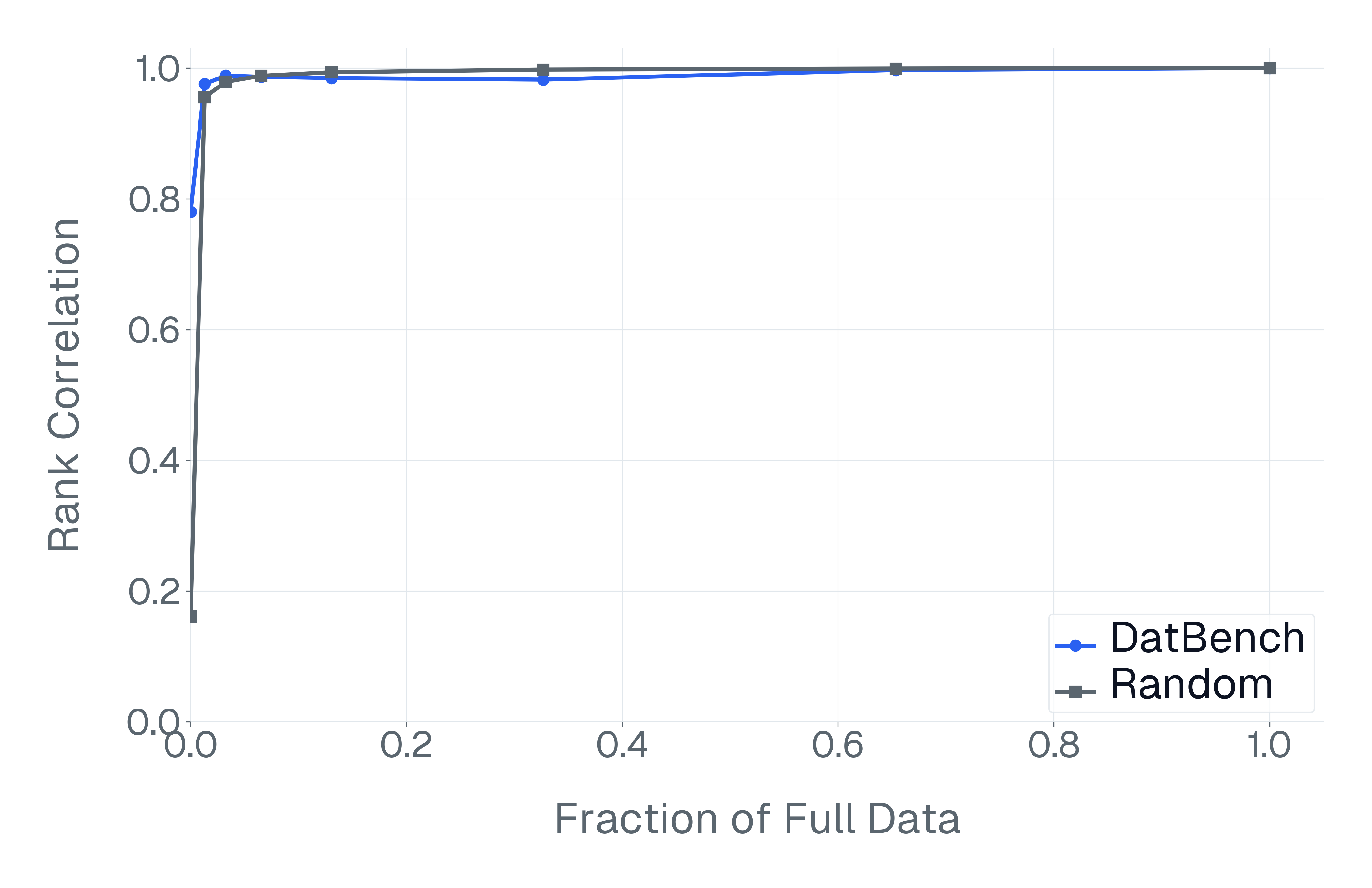}
        \caption{Scene}
    \end{subfigure}

    \vspace{0.4em}

    \begin{subfigure}{0.332\textwidth}
        \centering
        \includegraphics[width=\linewidth]{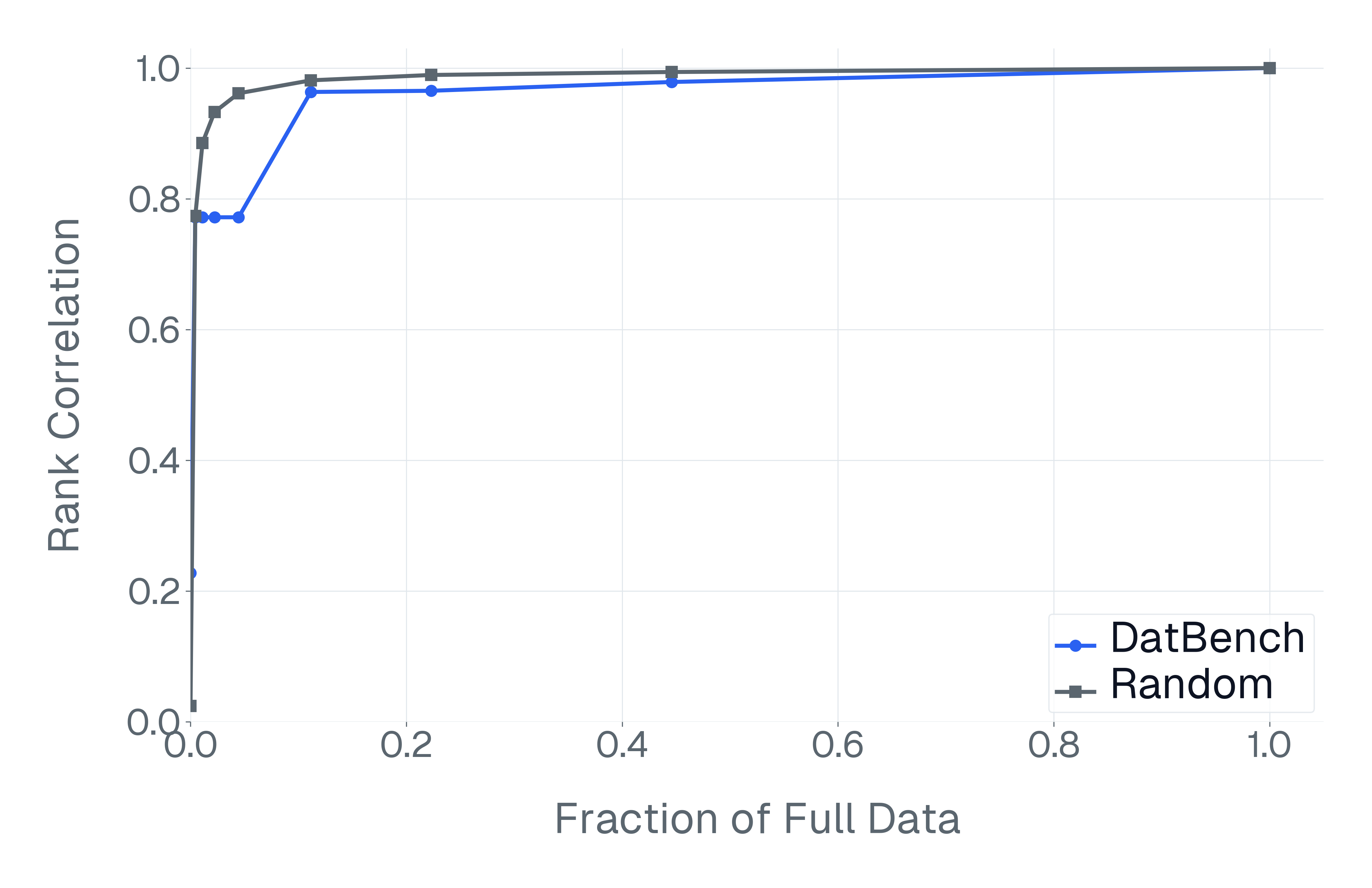}
        \caption{Counting}
    \end{subfigure}\hfill
    \begin{subfigure}{0.332\textwidth}
        \centering
        \includegraphics[width=\linewidth]{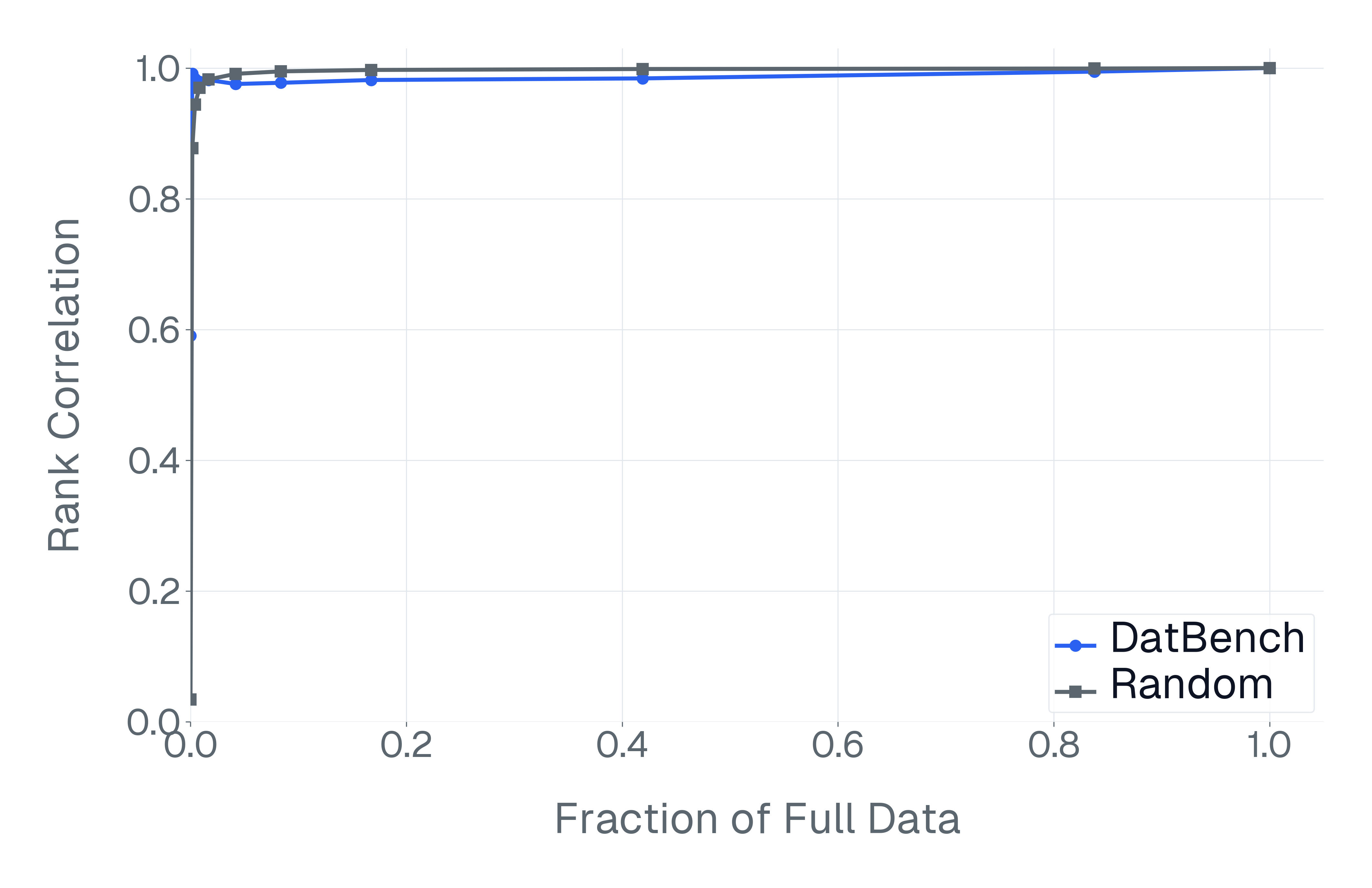}
        \caption{General}
    \end{subfigure}\hfill
    \begin{subfigure}{0.332\textwidth}
        \centering
        \includegraphics[width=\linewidth]{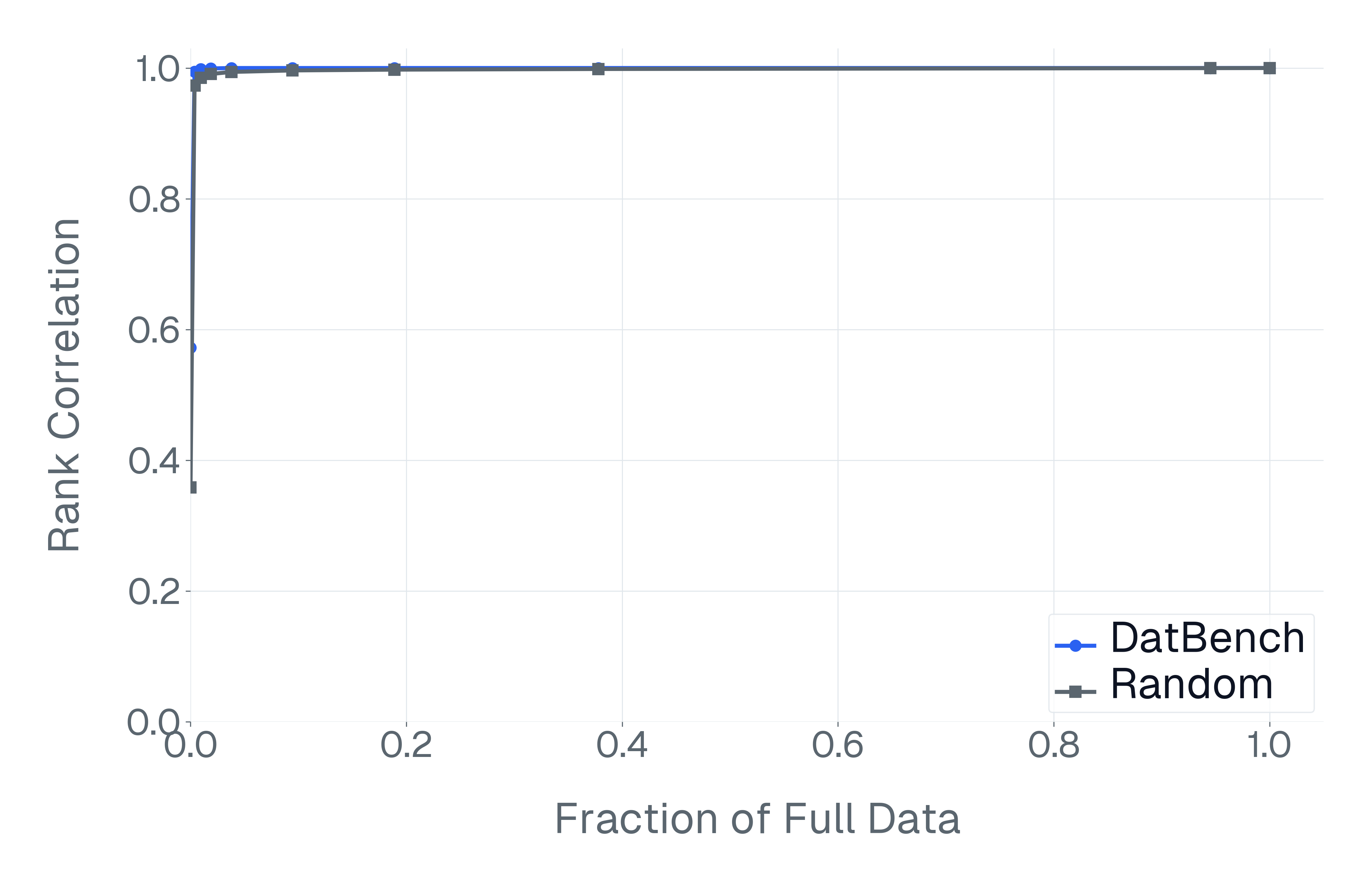}
        \caption{Grounding}
    \end{subfigure}

    \caption{Rank correlation as a function of retained data across all capabilities
    }
    \label{fig:rank_correlation_curves_all_cap}
\end{figure}

\clearpage
\section{Can be solved blind threshold}\label{app:step2}

\begin{table*}[h]
\centering
\small
\renewcommand{\arraystretch}{1.1}
\begin{tabular}{lll}
\toprule
\textbf{Capability} & \textbf{Dataset} & \textbf{Blind Threshold} \\
\midrule

Chart 
& ChartQA; ChartQA Pro; CharXiv; InfoVQA 
& 1 \\

Counting 
& CountBench 
& 4 \\
& TallyQA 
& 6 \\

Diagrams / Tables 
& MME-RW (Diagrams / Tables) 
& 1 \\
& AI2D 
& 5 \\

Document 
& OCR-VQA; CC-OCR (Document Parsing and KIE); DocVQA; OCRBench\_{v2} 
& 1 \\

Scene OCR 
& CC-OCR (Multi-Scene OCR) 
& 5 \\
& MME-RW (OCR in the wild) 
& 1 \\
& TextVQA 
& 5 \\

Spatial 
& MME-RW (Autonomous Driving, Remote Sensing) 
& 1 \\
& RealWorldQA 
& 8 \\

Math / Logic 
& LogicVista; MathVerse 
& 6 \\
& MathVision; MathVista (generative) 
& 1 \\
& MathVision; MathVista (MCQ) 
& 6 \\

Grounding 
& RefCOCO; RefCOCO+; RefCOCO-G; RefCOCO-M; Pixmo-Point 
& 1 \\

General 
& VQA-v2 
& 1 \\
& MMBench; MMMU-Pro 
& 6 \\

\bottomrule
\end{tabular}

\caption{Can-be-solved-blind thresholds for each evaluation dataset. Thresholds indicate the number of models that can correctly answer a question without visual input, above which the question is considered potentially solvable blind.}
\label{tab:blind_thresholds}
\end{table*}

\begin{table}[h]
\centering
\label{tab:capability_subset_selection}
\begin{tabular}{lrrr}
\toprule
\textbf{Capability} & \textbf{Total Samples} & \textbf{Samples Removed} & \textbf{Fraction Removed (\%)} \\
\midrule
General     & 220,413 & 158,841 & 72.07\% \\
Math        & 12,367  & 5,908   & 47.77\% \\
Chart       & 12,248  & 5,873   & 47.95\% \\
Scene OCR   & 13,990  & 6,054   & 43.27\% \\
Counting    & 39,079  & 16,263  & 41.62\% \\
Document    & 118,580 & 47,898  & 40.39\% \\
Grounding   & 36,960  & 10,221  & 27.65\% \\
Spatial     & 8,462   & 1,606   & 18.98\% \\
Table       & 9,021   & 1,482   & 16.43\% \\
\midrule
\textbf{Total} & \textbf{471,120} & \textbf{254,146} & \textbf{53.95\%} \\
\bottomrule
\end{tabular}
\caption{Blind solvable samples filtering statistics aggregated by capability.}
\end{table}

For each evaluation subset, we define a \emph{can-be-solved-blind} threshold, corresponding to the number of models that correctly answer a question without access to the image. Thresholds are chosen based on observed inflection points in blind accuracy curves and known sources of bias such as multiple-choice guessing, answer distribution skew, or lenient scoring functions. Thresholds for all datasets are summarized in Table~\ref{tab:blind_thresholds}.

As representative examples, inherently visual tasks such as chart understanding (e.g., ChartQA and related variants) exhibit near-zero blind solvability, with a clear inflection at a single model, motivating a minimal threshold of one. In contrast, multiple-choice evaluations such as RealWorldQA or MMMU-Pro admit non-trivial blind success due to chance-level guessing; in these cases, thresholds are set above random baselines (e.g., exceeding $\lfloor 0.25 \times N \rfloor$ models for four-option questions). Similarly, counting benchmarks such as CountBench and TallyQA show systematic biases toward small-number answers, leading to higher blind accuracy despite missing visual input; thresholds are therefore selected at empirical inflection points rather than at one. Finally, for datasets with lenient or continuous scoring metrics (e.g., multi-scene OCR), higher thresholds mitigate false positives arising from partial string matches or overly permissive correctness criteria.

\begin{figure}[p]
    \centering

    \begin{subfigure}{0.44\textwidth}
        \centering
        \includegraphics[width=\linewidth]{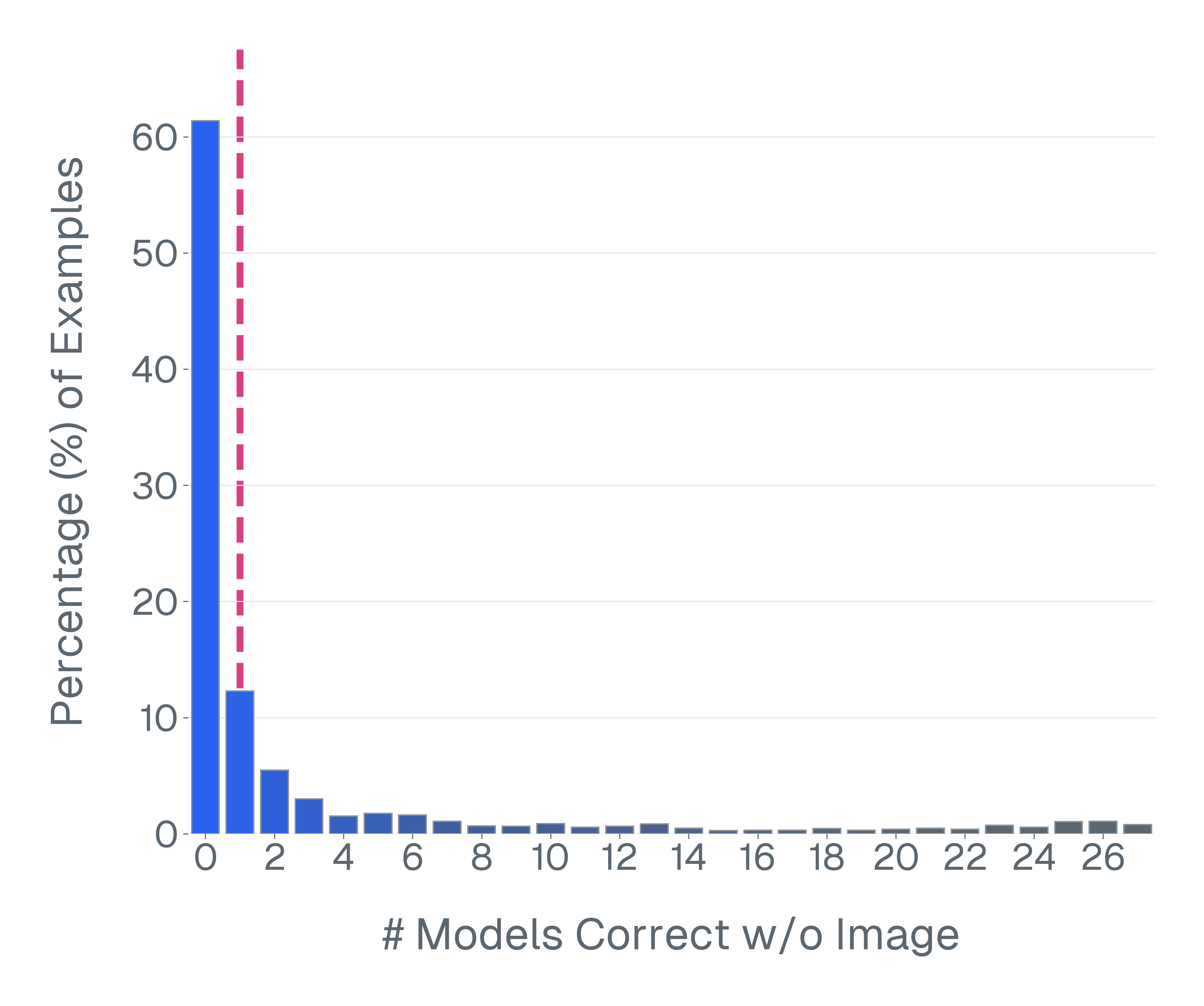}
        \caption{ChartQA}
    \end{subfigure}\hfill
    \begin{subfigure}{0.44\textwidth}
        \centering
        \includegraphics[width=\linewidth]{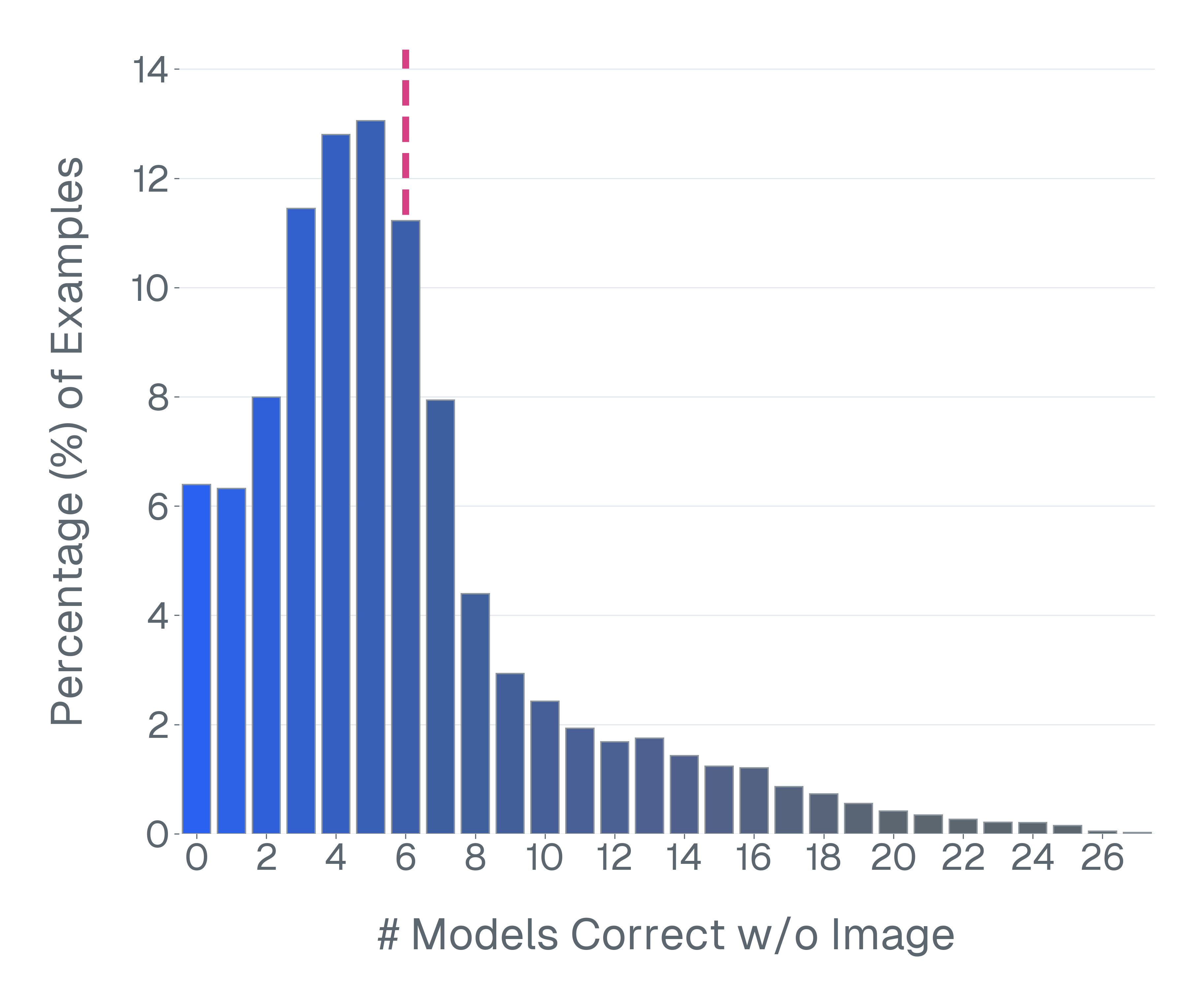}
        \caption{TallyQA}
    \end{subfigure}

    \vspace{0.5em}

    \begin{subfigure}{0.44\textwidth}
        \centering
        \includegraphics[width=\linewidth]{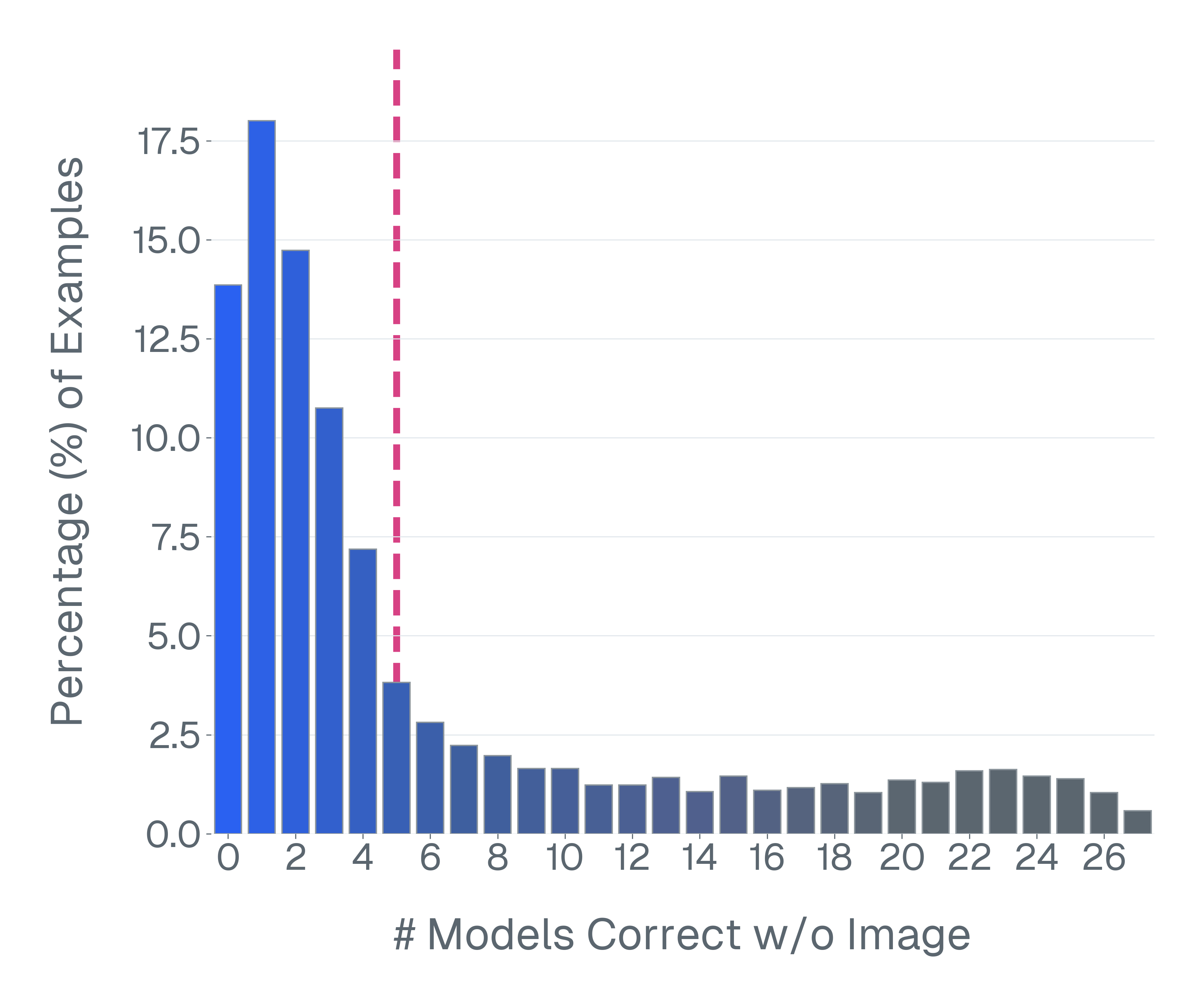}
        \caption{AI2D}
    \end{subfigure}\hfill
    \begin{subfigure}{0.44\textwidth}
        \centering
        \includegraphics[width=\linewidth]{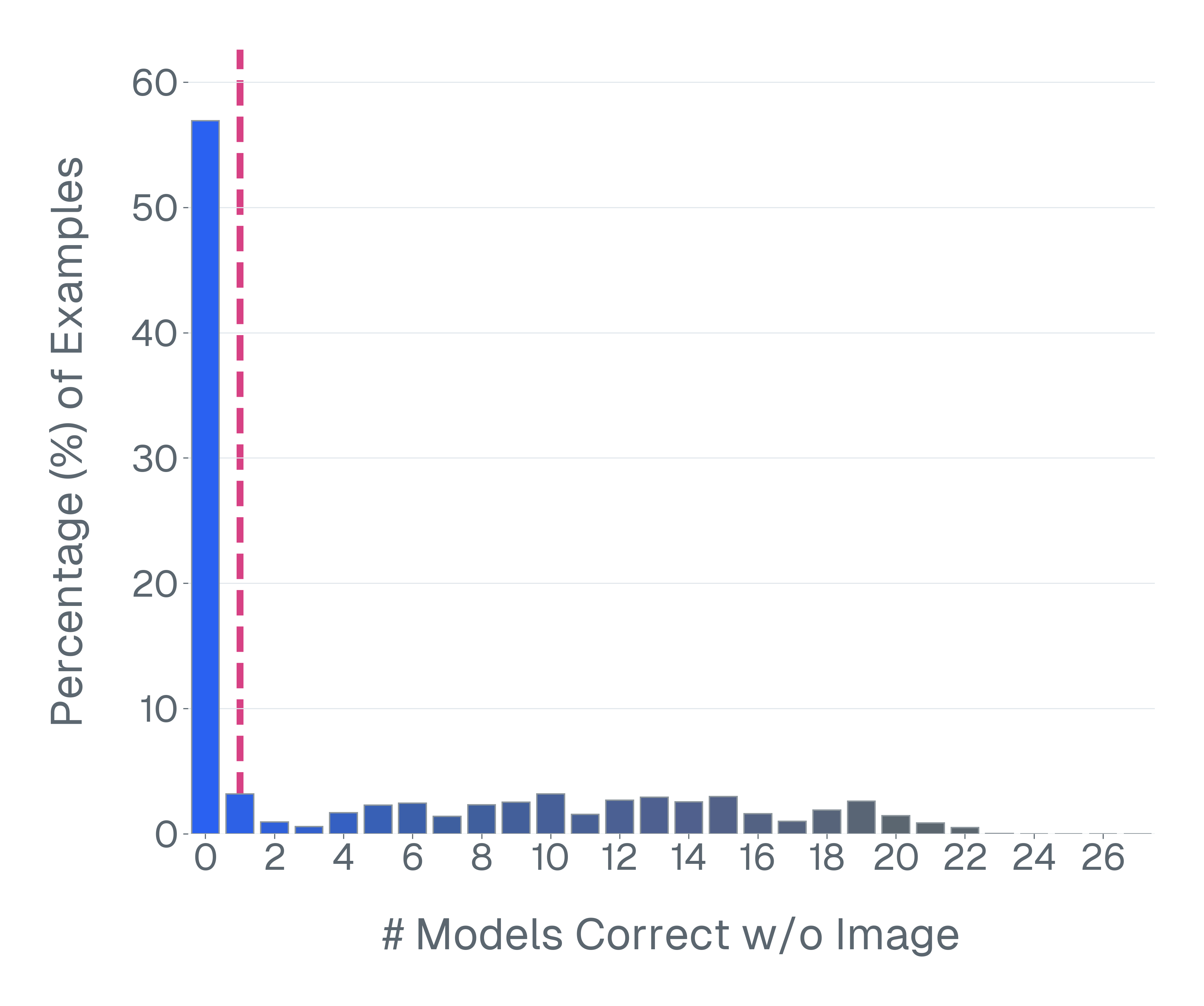}
        \caption{OCR-VQA}
    \end{subfigure}

    \vspace{0.5em}

    \begin{subfigure}{0.44\textwidth}
        \centering
        \includegraphics[width=\linewidth]{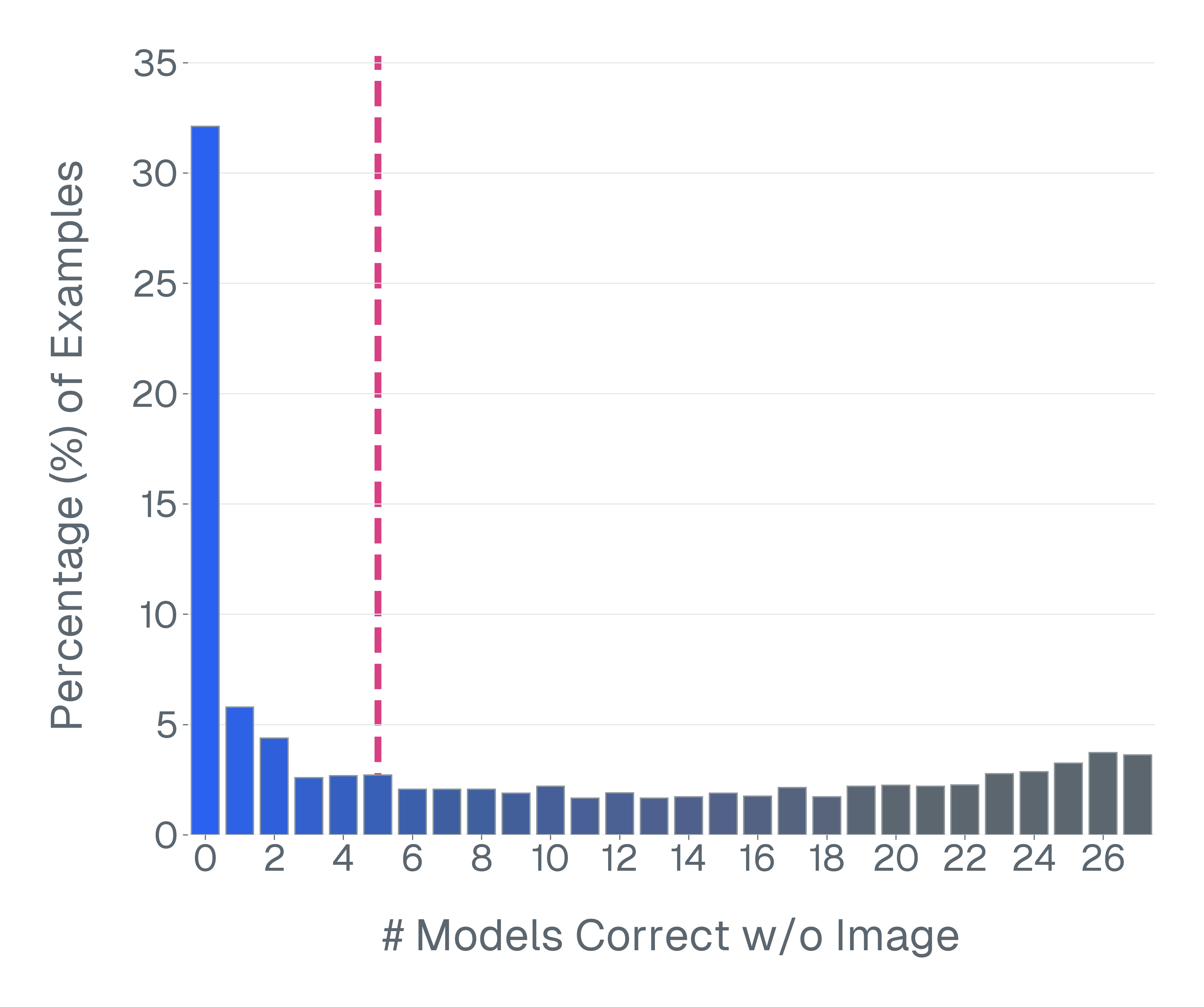}
        \caption{TextVQA}
    \end{subfigure}\hfill
    \begin{subfigure}{0.44\textwidth}
        \centering
        \includegraphics[width=\linewidth]{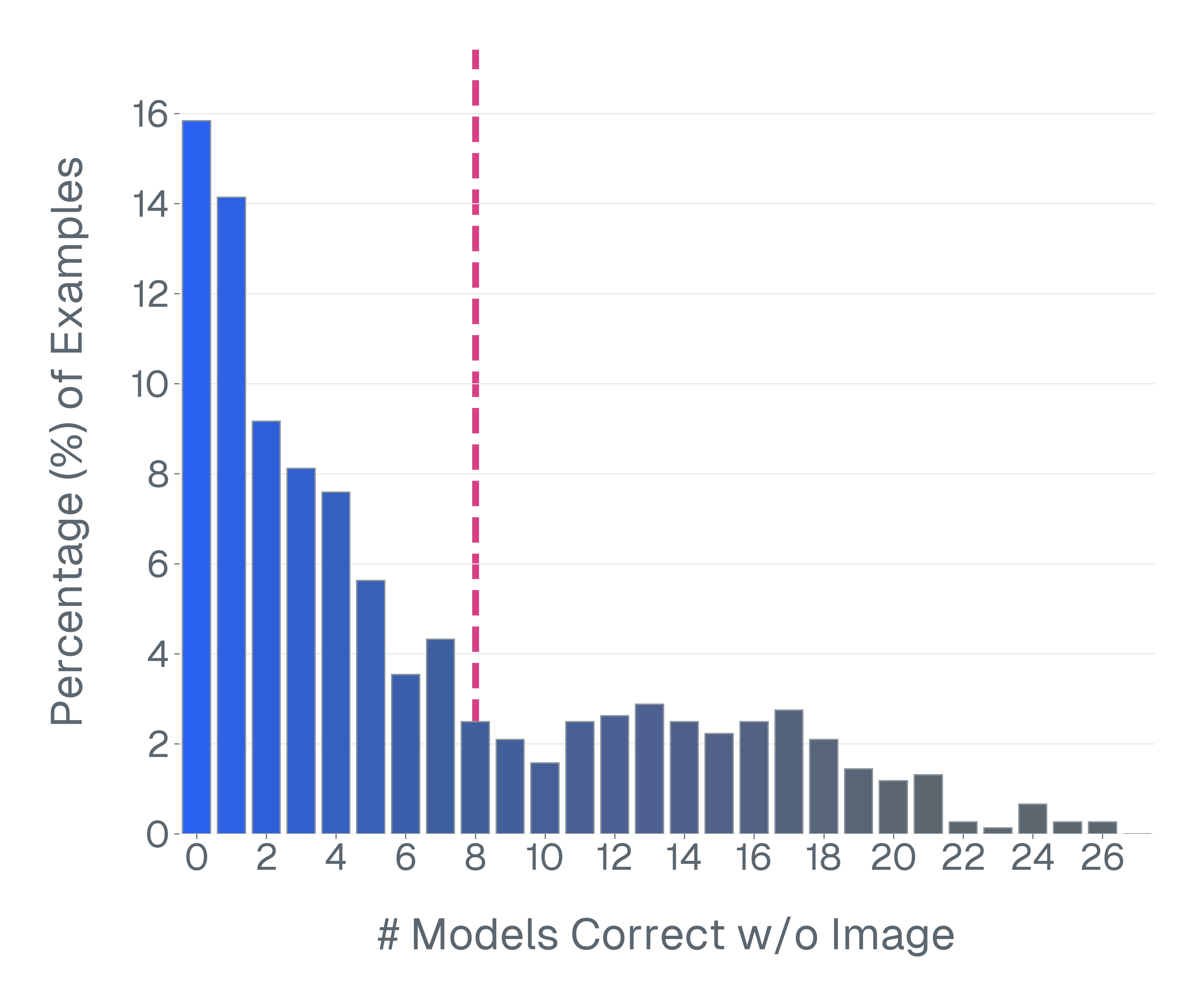}
        \caption{RealworldQA}
    \end{subfigure}
    \caption{Blind solvable thresholds and histograms across datasets (part 1 of 2).}
    \label{fig:blind_threshold_hists_1}
\end{figure}

\begin{figure}[p]\ContinuedFloat
    \centering
    \begin{subfigure}{0.44\textwidth}
        \centering
        \includegraphics[width=\linewidth]{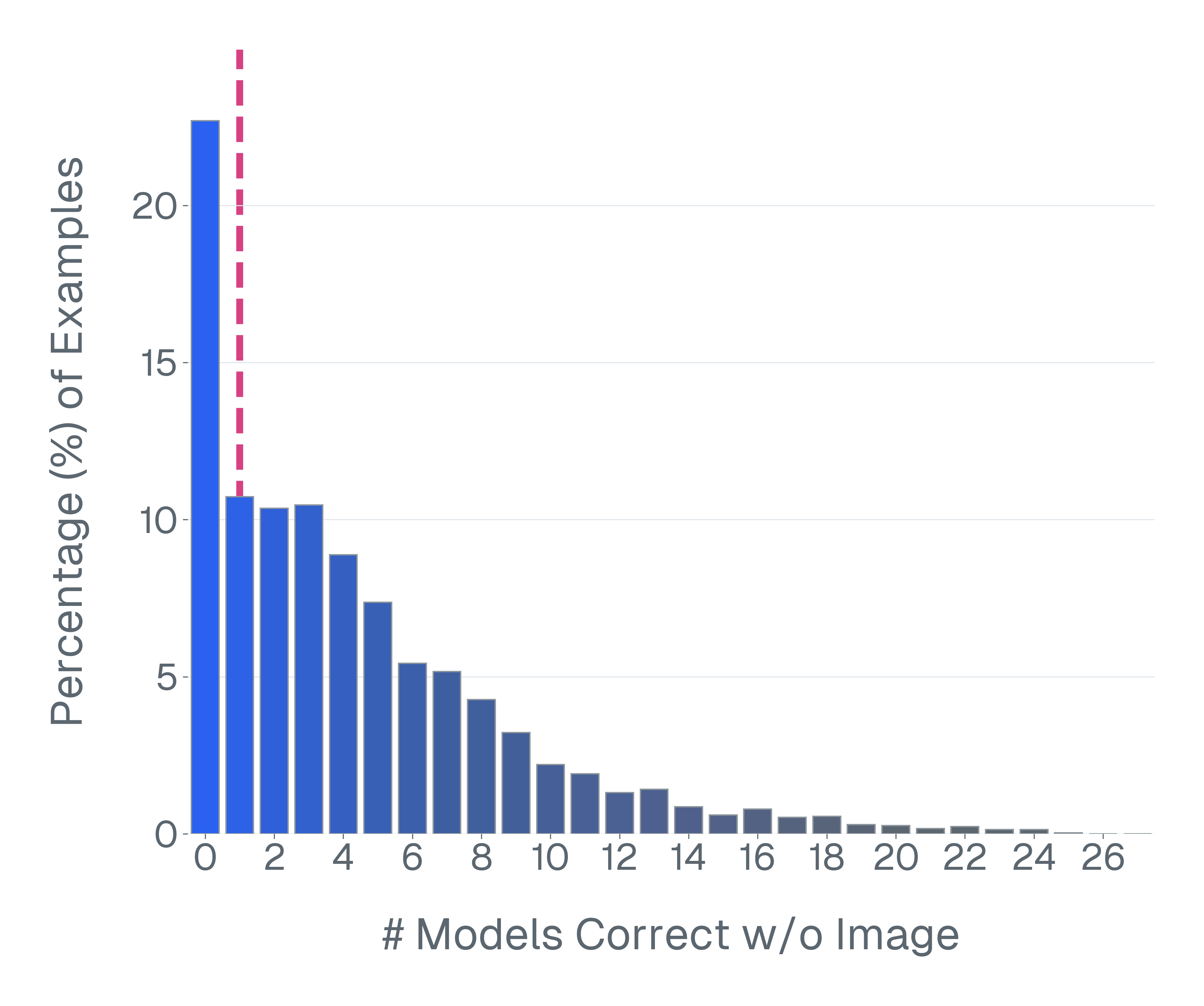}
        \caption{Mathvision}
    \end{subfigure}\hfill
    \begin{subfigure}{0.44\textwidth}
        \centering
        \includegraphics[width=\linewidth]{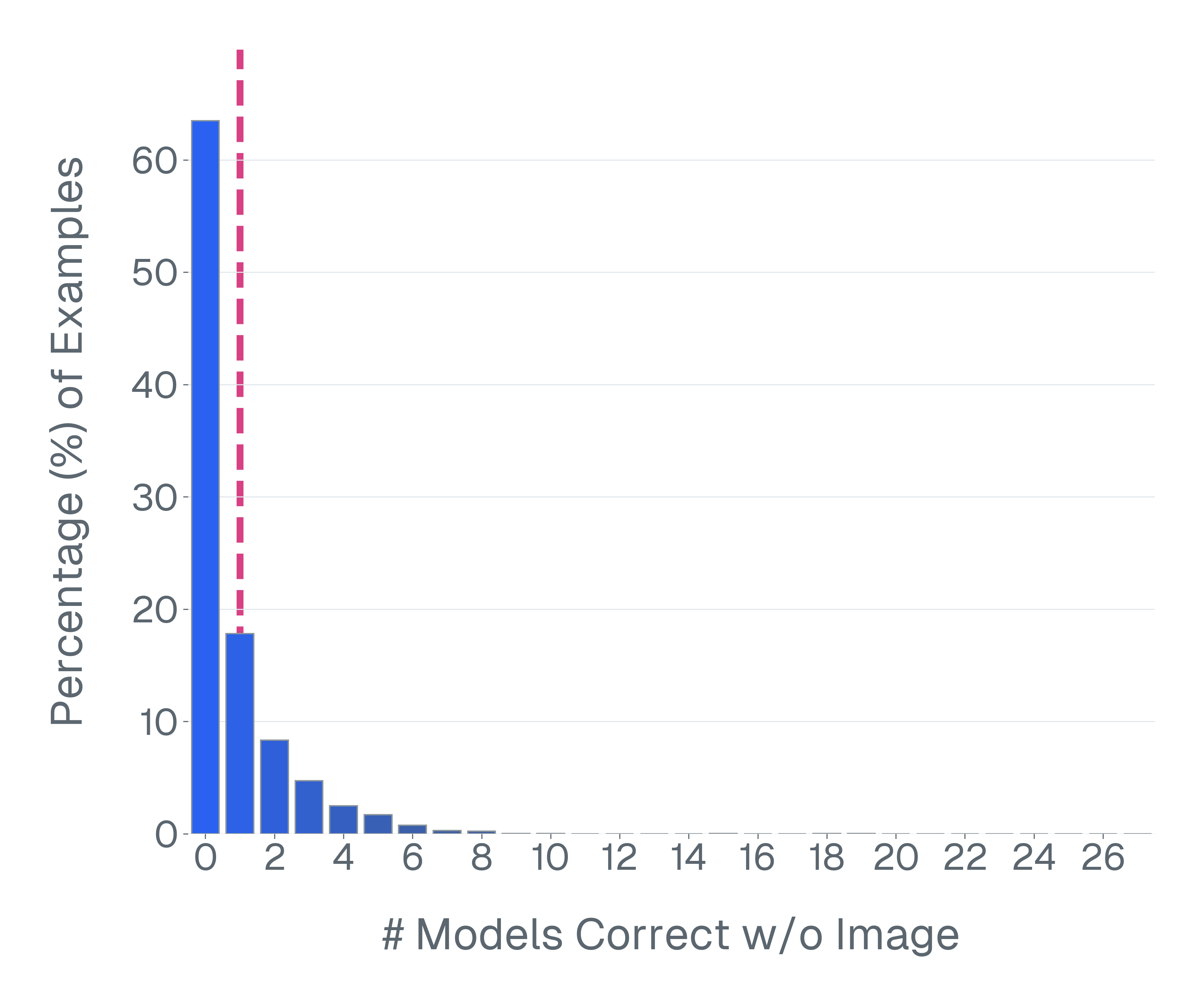}
        \caption{RefCOCO-M}
    \end{subfigure}

    \vspace{0.5em}

    \begin{subfigure}{0.44\textwidth}
        \centering
        \includegraphics[width=\linewidth]{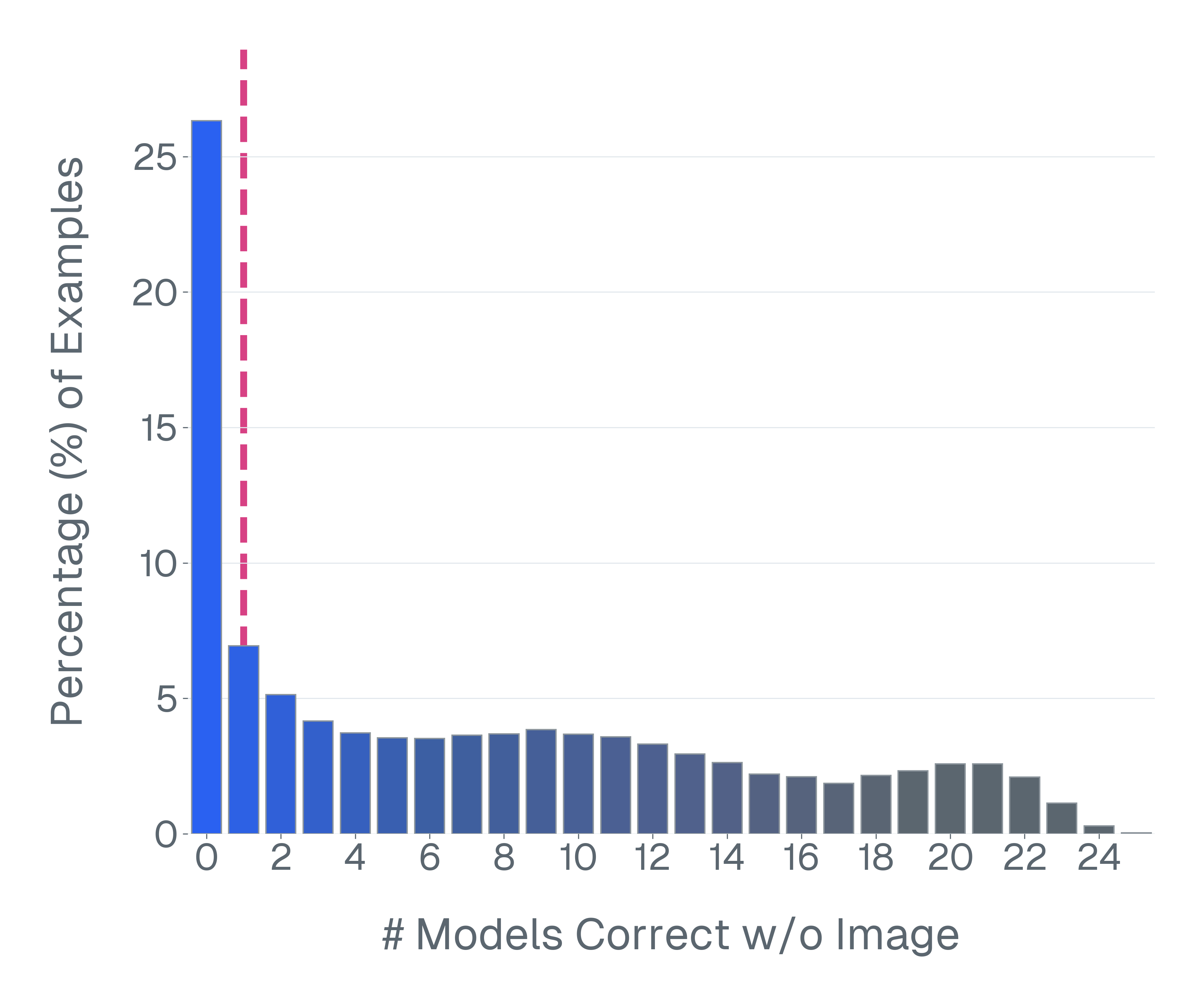}
        \caption{VQAv2}
    \end{subfigure}

    \caption{
    Blind solvable thresholds and histograms across datasets (part 2 of 2).
    }
    \label{fig:blind_threshold_hists_2}
\end{figure}

\end{document}